\documentclass[runningheads]{llncs}

\usepackage[mobile]{eccv}

\usepackage{eccvabbrv}

\usepackage{graphicx}
\usepackage{booktabs}

\usepackage[accsupp]{axessibility}  % Improves PDF readability for those with disabilities.

\usepackage[pagebackref,breaklinks,colorlinks,citecolor=eccvblue]{hyperref}
\usepackage{orcidlink}

\usepackage{wrapfig} 
\usepackage{pifont}% http://ctan.org/pkg/pifont
\usepackage{colortbl}
\usepackage{adjustbox}
\usepackage{makecell}
\usepackage{multirow}
\usepackage{bm}

\definecolor{rowgray}{gray}{0.93}
\newcommand{\xmark}{\ding{55}}
\newcommand{\NA}{---}
\newcommand{\good}[1]{\cellcolor{green!20}#1}
\newcommand{\bad}[1]{\cellcolor{red!20}#1}
\definecolor{best}{HTML}{F4B183}     % darker orange
\definecolor{second}{HTML}{F8CBAD}   % medium peach
\definecolor{third}{HTML}{FCE4D6}    % very light peach
\newcommand{\best}[1]{\cellcolor{best}#1}
\newcommand{\second}[1]{\cellcolor{second}#1}
\newcommand{\third}[1]{\cellcolor{third}#1}
\newcommand{\equalcontrib}{\textsuperscript{*}}
\newcommand{\corresp}{\textsuperscript{\textdagger}}

\usepackage{diagbox}
\definecolor{rankone}{RGB}{255,170,90}
\definecolor{ranktwo}{RGB}{255,185,115}
\definecolor{rankthree}{RGB}{255,200,140}
\definecolor{rankfour}{RGB}{255,215,165}
\definecolor{rankfive}{RGB}{255,230,200}
\definecolor{ranksix}{RGB}{255,242,225}

\begin{document}

\title{DF3DV-1K: A Large-Scale Dataset and Benchmark for Distractor-Free Novel View Synthesis} 

\titlerunning{DF3DV-1K}

\author{
Cheng-You Lu\inst{1}\corresp\orcidlink{0000-0002-7653-897X} \and
Yi-Shan Hung\inst{2}\corresp\orcidlink{0009-0000-0099-5034} \and
Wei-Ling Chi\inst{3}\equalcontrib\orcidlink{0009-0006-2763-8719} \and
Hao-Ping Wang\inst{3}\equalcontrib\orcidlink{0009-0004-5498-2051} \and
Charlie Li-Ting Tsai\inst{1}\equalcontrib\orcidlink{0009-0008-7308-4643} \and
Yu-Cheng Chang\inst{1}\orcidlink{0000-0001-9244-0318} \and
Yu-Lun Liu\inst{3}\orcidlink{0000-0002-7561-6884} \and
Thomas Do\inst{1}\orcidlink{0000-0002-8597-5944} \and
Chin-Teng Lin\inst{1}\orcidlink{0000-0001-8371-8197}
}

\authorrunning{C.-Y. Lu et al.}

\institute{University of Technology Sydney \and
University of Sydney \and
National Yang Ming Chiao Tung University}

\maketitle

\begingroup
\renewcommand{\thefootnote}{\textdagger}
\footnotetext{Equal contribution.}
\renewcommand{\thefootnote}{*}
\footnotetext{Equal contribution.}
\endgroup

\begin{center}
  \includegraphics[width=\textwidth]{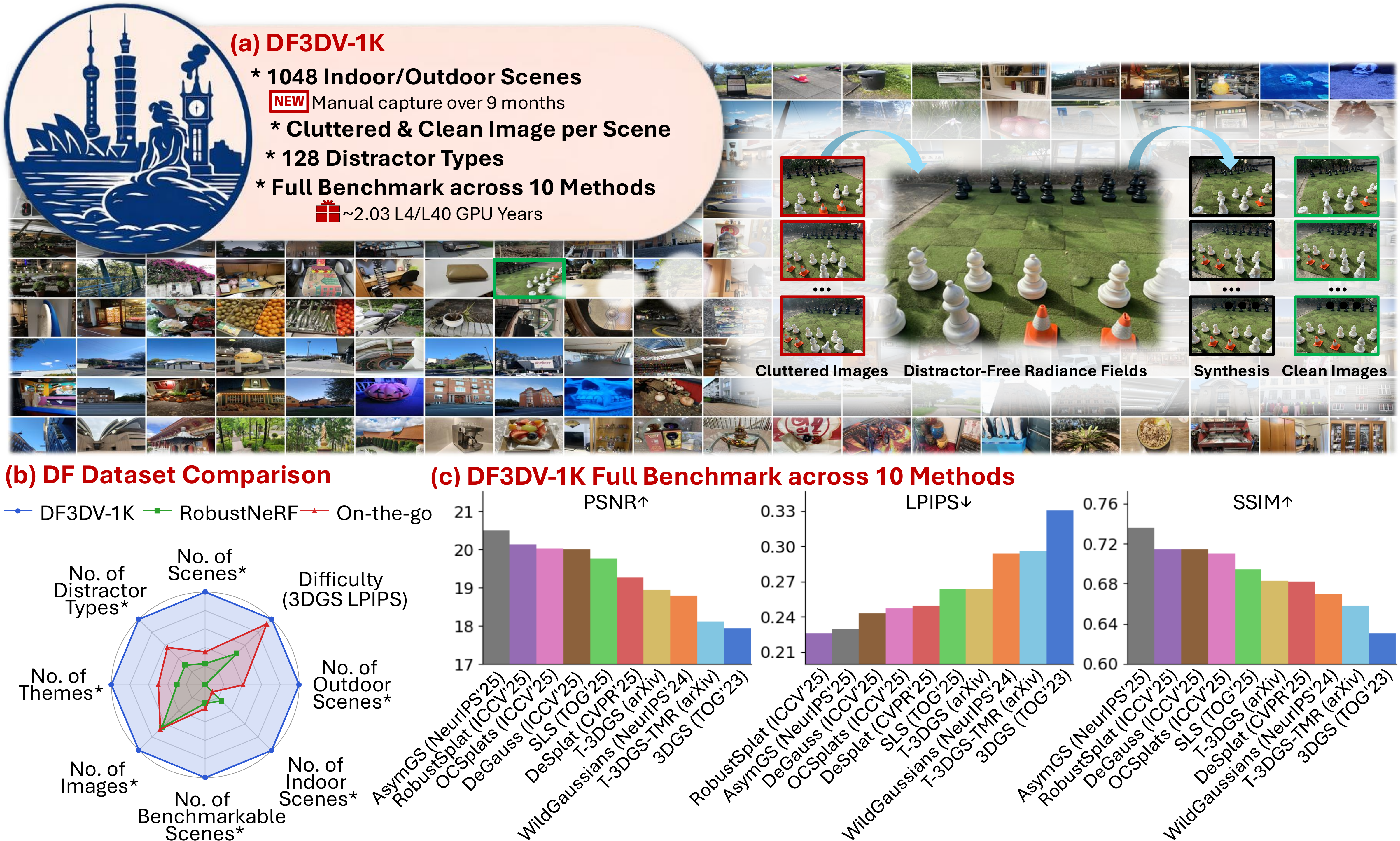}
  \captionof{figure}{\textbf{Features of the DF3DV-1K dataset and benchmark.} \textbf{(a) DF3DV-1K Dataset.} We introduce DF3DV-1K, a large-scale distractor-free dataset comprising 1,048 manually captured indoor and outdoor scenes, with clean and cluttered images per scene spanning 128 distractor types.
\textbf{(b) Distractor-Free Dataset Comparison.} 
* denotes log scale in the normalized radar chart.
We compare DF3DV-1K with public datasets~\cite{sabour2023robustnerf, Ren2024NeRF}, showing a larger scale (e.g., $\sim$100$\times$ more scenes), broader scene diversity (e.g., $\sim$10$\times$ more distractor types), and increased difficulty, reflected by higher 3DGS~\cite{kerbl20233d} LPIPS.
\textbf{(c) DF3DV-1K Benchmark.} A comprehensive benchmark across nine recent distractor-free radiance field methods~\cite{sabour2025spotlesssplats,2025RobustSplat,ling2025ocsplats,markin2024t,kulhanek2024wildgaussians,wang2025desplat,wang2025degauss,li2025asymgs} and 3DGS~\cite{kerbl20233d} reveals varying levels of robustness to distractors. 
%AsymGS~\cite{li2025asymgs} and RobustSplat~\cite{2025RobustSplat} emerge as the most robust methods, while OCSplats~\cite{ling2025ocsplats} and DeGauss~\cite{wang2025degauss} rank second.
}
\label{fig:teaser}
\end{center}

\begin{abstract}
Advances in radiance fields have enabled photorealistic novel view synthesis.
In several domains, large-scale real-world datasets have been developed to support comprehensive benchmarking and to facilitate progress beyond scene-specific reconstruction.
However, for distractor-free radiance fields, a large-scale dataset with clean and cluttered images per scene remains lacking, limiting the development.
To address this gap, we introduce DF3DV-1K, a large-scale real-world dataset comprising 1,048 scenes, each providing clean and cluttered image sets for benchmarking.
In total, the dataset contains 89,924 images captured using consumer cameras to mimic casual capture, spanning 128 distractor types and 161 scene themes across indoor and outdoor environments.
A curated subset of 41 scenes, DF3DV-41, is systematically designed to evaluate the robustness of distractor-free radiance field methods under challenging scenarios.
Using DF3DV-1K, we benchmark nine recent distractor-free radiance field methods and 3D Gaussian Splatting, identifying the most robust methods and the most challenging scenarios.
Beyond benchmarking, we demonstrate an application of DF3DV-1K by fine-tuning a diffusion-based 2D enhancer to improve radiance field methods, achieving average improvements of 0.96 dB PSNR and 0.057 LPIPS on the held-out set (e.g., DF3DV-41) and the On-the-go dataset.
We hope DF3DV-1K facilitates the development of distractor-free vision and promotes progress beyond scene-specific approaches.
The dataset and leaderboard are available at \url{https://johnnylu305.github.io/df3dv1k_web/}.
\end{abstract}
\section{Introduction}
\label{sec:intro}

%Neural Radiance Fields~\cite{mildenhall2021nerf} and 3D Gaussian Splatting (3DGS)~\cite{kerbl20233d} have been popular for novel view synthesis due to their photorealistic rendering abilities.
Radiance fields~\cite{mildenhall2021nerf, kerbl20233d} have shown photorealistic novel view synthesis abilities.
\begin{wrapfigure}[20]{r}{0.48\textwidth} 
  \centering
  \includegraphics[width=0.48\textwidth]{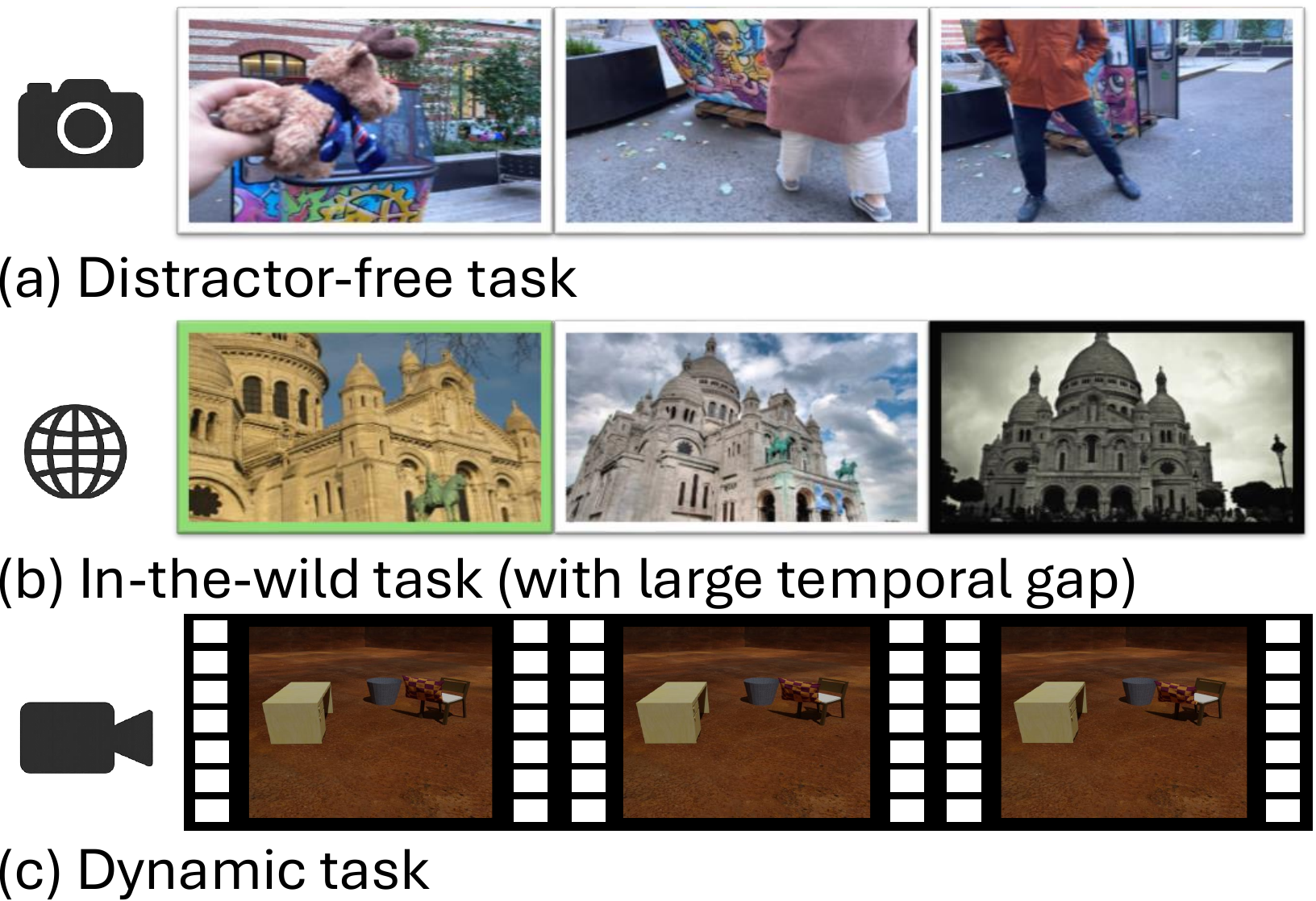}
  \caption{
\textbf{Radiance field variants.}
(a) Distractor-free tasks~\cite{Ren2024NeRF} use casually captured images within a short period.
(b) In-the-wild tasks~\cite{martin2021nerf, jin2021image}, with large temporal gaps, often target images collected across seasons.
(c) Dynamic tasks~\cite{wu2022d} assume densely captured sequential data.
}
\label{fig:qual_three}
\end{wrapfigure}
Their variants such as in-the-wild (with large temporal gap)~\cite{martin2021nerf, dahmani2024swag, xu2024wild, zhang2024gaussian, wang2024we, chen2022hallucinated, chen2024nerfhugs, wang2024distractor, kulhanek2024wildgaussians, li2025asymgs, wang2025desplat, bao2025seg, fu2025robustsplat++, li2025wild3a, wang2025ie, wu2025wildsplatting, mithun2025diffusion} and dynamic radiance fields~\cite{song2023nerfplayer, wang2023mixed, cao2023hexplane, fridovich2023k, wu2022d, wu20244d, li2022streaming, li2023dynibar, xiao2025vdegaussian, sun2025sdd, wang2026uncertainty} have been introduced together with comprehensive benchmarks and large-scale datasets~\cite{reizenstein2021common, tung2024megascenes, zhou2018stereo, liu2021infinite, jensen2014large, yeshwanth2023scannet++, ling2024dl3dv, tung2024megascenes, lu2024diva, lin2021deep, kastingschafer2025seed4d, broxton2020immersive} to identify remaining limitations and support further research.
Despite recent advances, distractor-free radiance field methods, which aim to synthesize clean novel views from images containing visual distractors commonly observed in short-span casual image capture (see~\cref{fig:qual_three}), still lack a large-scale, challenging dataset and benchmark with clean and cluttered images per scene (see~\cref{tab:reg_datasets}).
\begin{table}[tb]
  \caption{\textbf{Comparison of the 3D datasets.} 
We compare datasets based on the number of scenes, image resolution, capture density, scale of visual changes, camera intrinsic consistency within each scene, availability of clean and cluttered images, and indoor/outdoor coverage.
Despite recent progress, distractor-free radiance field research still lacks a large-scale real-world dataset with clean and cluttered images per scene, limiting further development in this area.}
  %\scriptsize
  %\tiny
  \begin{adjustbox}{width=\linewidth}
  \label{tab:reg_datasets}
  \centering
  \begin{tabular}{lcc|ccc|cccc}
    \toprule
    \textbf{Dataset} & \textbf{Scene} & \textbf{Resolution} &
    \textbf{Capture} & \textbf{Change Scale} & \textbf{Intrinsics} & \textbf{Clean} & \textbf{Clutter} &
    \textbf{Indoor} & \textbf{Outdoor} \\
    \midrule
    \rowcolor{rowgray}
    \multicolumn{10}{l}{\textit{Vanilla}} \\
    DTU~\cite{jensen2014large} & 124 & 2MP & \textit{sparse} & \NA & \textit{shared} & \checkmark & \xmark &\checkmark & \xmark \\
    RealEstate10K~\cite{zhou2018stereo} & 10K & 720p HD &
    \textit{dense} & \NA & \textit{shared} & \checkmark & \xmark &
    \checkmark & \xmark \\
    ACID~\cite{liu2021infinite} & 13K & \NA &
    \textit{dense} & \NA & \textit{shared} & \checkmark & \xmark &
    \xmark & \checkmark \\
    %CO3D~\cite{reizenstein2021common} & 19K & \NA &
    %\textit{dense} & \NA & \textit{shared} & \checkmark & \xmark &
    %\checkmark & \checkmark \\
    ScanNet++~\cite{yeshwanth2023scannet++} & 1K & $\sim$8K UHD &
    \textit{dense} & \NA & \textit{shared} & \checkmark & \xmark &
    \checkmark & \xmark \\
    DL3DV-10K~\cite{ling2024dl3dv} & 10K & 4K UHD &
    \textit{dense} & \NA & \textit{shared} & \checkmark & \xmark &
    \checkmark & \checkmark \\
    \midrule
    \rowcolor{rowgray}
    \multicolumn{10}{l}{\textit{In-the-wild (with large temporal gap)}} \\
    Phototourism~\cite{jin2021image} & 25 & \NA &
    \textit{sparse} & \textit{global} & \textit{multiple} &  \checkmark & \checkmark &
    \xmark & \checkmark \\
    MegaScenes~\cite{tung2024megascenes} & 430K & \NA &
    \textit{sparse} & \textit{global} & \textit{multiple} & \xmark & \checkmark &
    \checkmark & \checkmark \\
    \midrule
    \rowcolor{rowgray}
    \multicolumn{10}{l}{\textit{Dynamic}} \\
    D$^2$NeRF~\cite{wu2022d} & 5 & <qHD &
    \textit{dense} & \textit{local} & \textit{shared} & \checkmark & \checkmark &
    \checkmark & \xmark \\
    SEED4D~\cite{kastingschafer2025seed4d} & 10K & \NA &
    \textit{dense} & \textit{local} & \textit{shared} & \xmark & \checkmark &
    \xmark & \checkmark \\
 %   Dynamic RealEstate10K~\cite{chen2026wildrayzer} & 15K & \NA &
 %   \textit{dense} & \textit{local} & \textit{shared} & \xmark & \checkmark &
%    \checkmark & \xmark \\
    \midrule
    \rowcolor{rowgray}
    \multicolumn{10}{l}{\textit{Distractor-free }} \\
    RobustNeRF~\cite{sabour2023robustnerf} & 5 & 12MP &
    \textit{sparse} & \textit{local} & \textit{shared} & \checkmark & \checkmark &
    \checkmark & \xmark \\
    On-the-go~\cite{Ren2024NeRF} & 12 & 12MP &
    \textit{sparse} & \textit{local} & \textit{shared} & \checkmark & \checkmark &
    \checkmark & \checkmark \\
    DF3DV-1K (Ours) & 1K & $\sim$12MP &
    \textit{sparse} & \textit{local} & \textit{shared} & \checkmark & \checkmark &
    \checkmark & \checkmark \\
    \bottomrule
  \end{tabular}
  \end{adjustbox}
\end{table}
This absence leads to two major issues: (1) it becomes difficult to identify the strengths and limitations of recent approaches, and (2) progress from scene-specific methods toward generalizable solutions is hindered.
\begin{wrapfigure}[21]{r}{0.48\textwidth}
  \centering
  \includegraphics[width=0.48\textwidth]{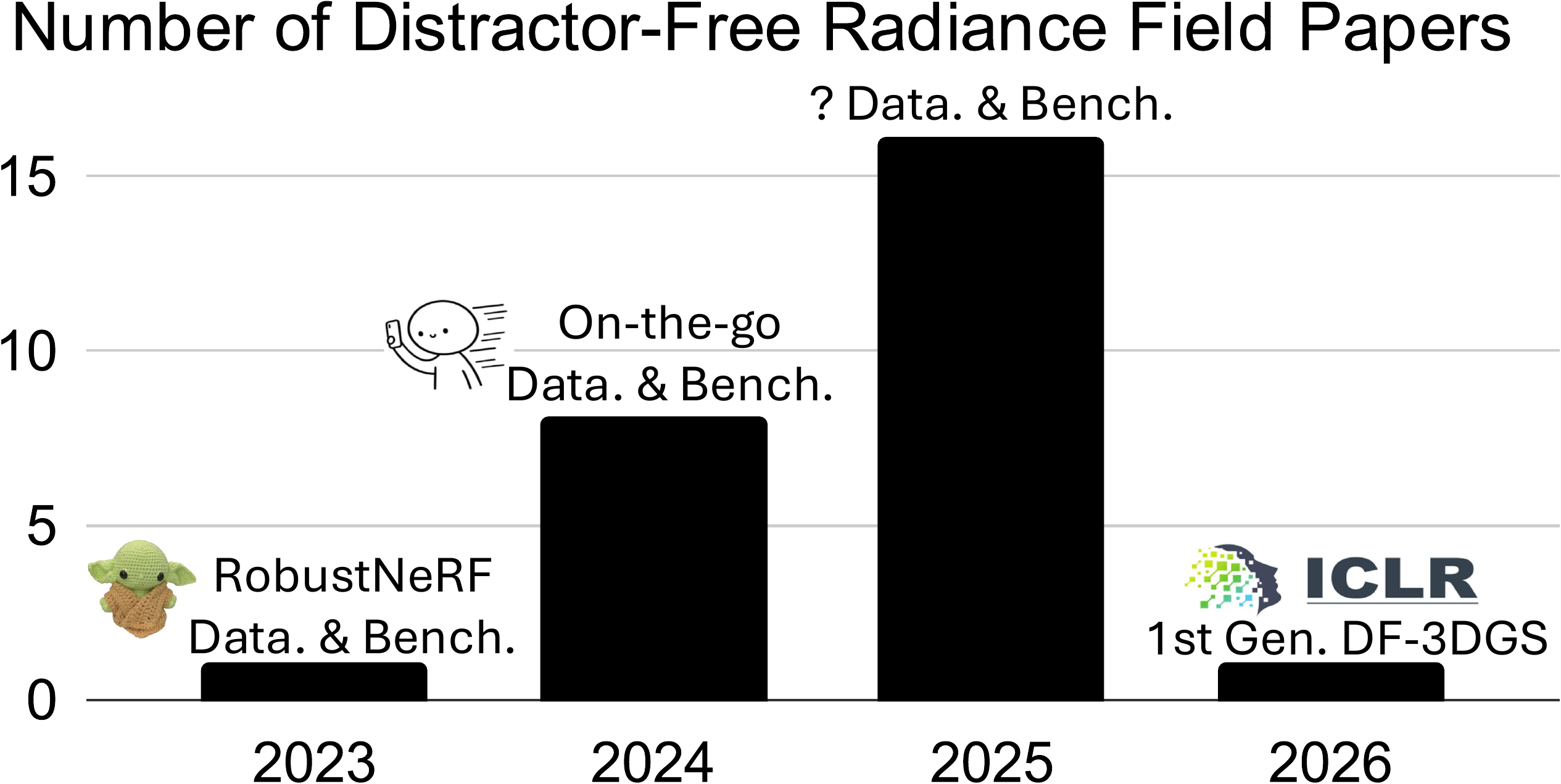}
  \caption{
\textbf{Number of distractor-free radiance field papers.}
The first distractor-free radiance field method~\cite{sabour2023robustnerf}, targeting images captured over short time spans, was introduced in 2023 together with a benchmark.
The research area rapidly gained attention in 2024 with the release of the On-the-go benchmark~\cite{Ren2024NeRF}.
Although the number of works doubled in 2025, a public, large-scale, challenging dataset and benchmark systematically designed for this area are still lacking.
%This highlights the need for a new dataset and benchmark.
}
  \label{fig:papers}
\end{wrapfigure}

\begin{figure}[tb]
  \centering
  \includegraphics[width=0.99\linewidth]{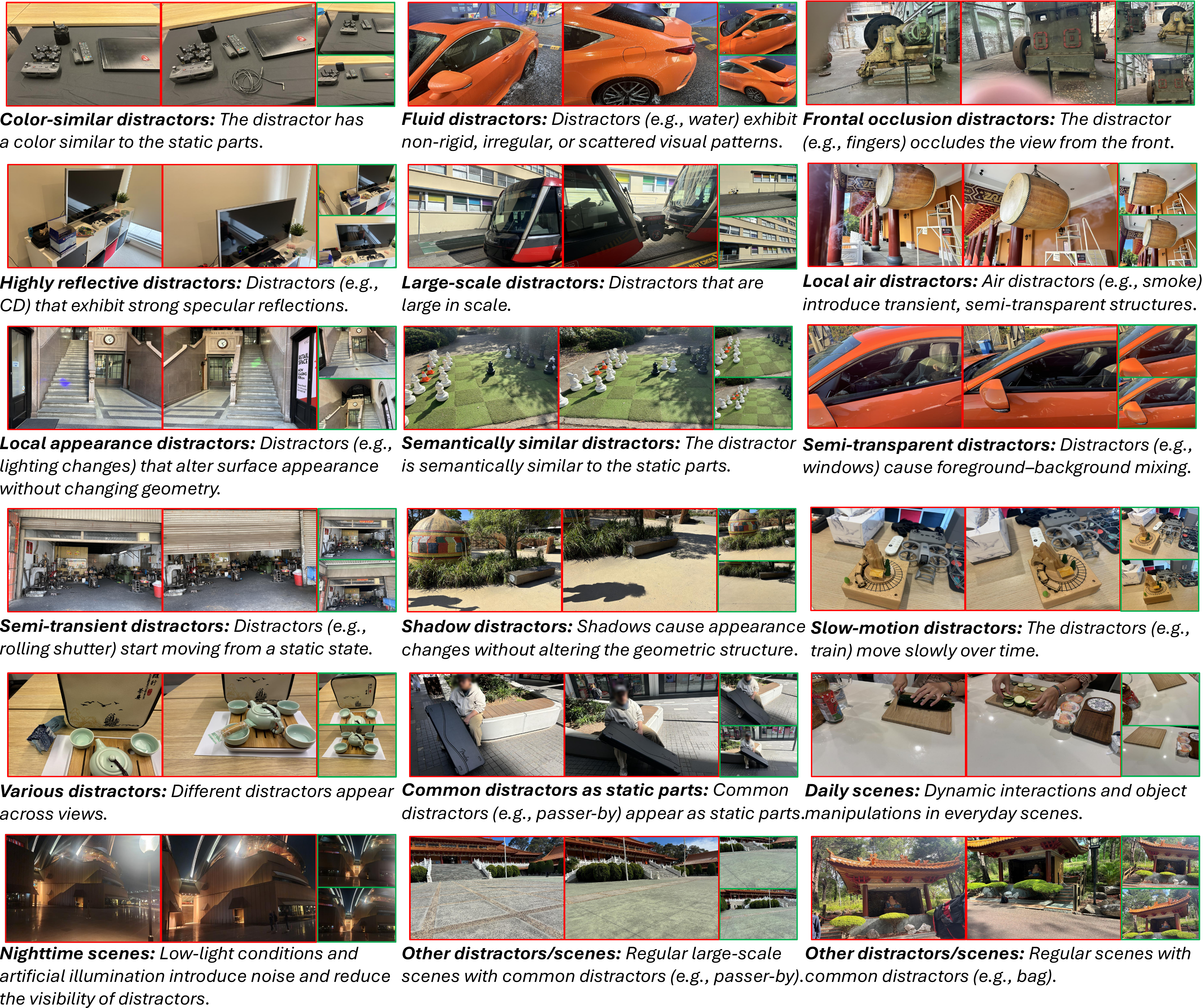}
  \caption{
\textbf{Samples of systematically designed scenarios in DF3DV-41.}
Each example, where green and red boxes denote clean and cluttered images, respectively, illustrates a carefully designed scenario with curated distractor types or scene conditions, highlighting the benchmark’s diversity.
Motivated by the observations of the potential limitations of prior methods, these 17 scenarios systematically evaluate the robustness of distractor-free 3D reconstruction methods under challenging conditions.}
\label{fig:bench41}
\end{figure}

In fact, the first issue can be observed in recent works~\cite{2025RobustSplat, li2025asymgs, ling2025ocsplats, wang2025degauss}.
For example, Fig.~10 in~\cite{ling2025ocsplats}, Fig.~7 in~\cite{2025RobustSplat}, and Fig.~4 in~\cite{li2025asymgs} reveal subtle background differences often visible only under zoomed-in views.
In other cases (see Fig.~4 in~\cite{wang2025degauss}), evaluations rely on cross-domain datasets supporting only qualitative analysis. 
%and lacking clean views for quantitative benchmarking.
%In addition,~\cref{fig:papers} shows the number of distractor-free radiance field papers targeting casually captured images.
The number of distractor-free radiance field papers targeting casually captured images doubled in 2025 (see~\cref{fig:papers}), yet no systematically designed, large-scale public dataset or benchmark has emerged to keep pace with this rapid growth.
\begin{table}[tb]
  \caption{
\textbf{Comparison of real-world distractor-free radiance field datasets.}
We compare datasets based on the number of scenes, indoor and outdoor environments, distractor types, themes, images, and image resolution. 
DF3DV-1K is a large-scale, diverse real-world dataset, providing clean and cluttered images for each scene.
  }
  \scriptsize
  \label{tab:df_datasets}
  \centering
  \begin{tabular}{lccccccc|c}
    \toprule
    \textbf{Dataset} & \textbf{Scene} & \textbf{Indoor} & \textbf{Outdoor} & \textbf{Distractor} & \textbf{Theme} & \textbf{Image} & \textbf{Resolution} & \textbf{Clean} \\
    \midrule
    \rowcolor{gray!15}
    \multicolumn{9}{l}{\textit{Not published yet (2026 June)}} \\
%    RealX3D~\cite{liu2025realx3d} & 55 & 55 & 0 & \NA & \NA & \NA & 33MP & \checkmark \\
    T-3DGS~\cite{markin2024t} & 5 & 5 & 0 & >2 & 4 & \NA & \NA & \checkmark \\
    %DF3DV-1K (Ours) & 1K & 726 & 322 & 128 & 161 & 89.9K & $\sim$12MP & \checkmark \\
    \midrule
    \rowcolor{gray!15}
    \multicolumn{9}{l}{\textit{Not released yet (2026 June)}} \\
    Entity-NeRF~\cite{otonari2024entity} & 3 & 0 & 3 & >3 & 3 & 78 & \NA & \xmark \\
    UniVerse~\cite{cao2025universe} & 6 & \NA & >3 & >3 & >3 & \NA & \NA & \checkmark \\
    Wild3A~\cite{li2025wild3a} & 3 & 2 & 1 & >3 & 3 & \NA & \NA & \checkmark \\
    StaticNeRF~\cite{lee2025freeze} & 8 & 8 & 0 & \NA & 2 & \NA & \NA & \checkmark \\
%    D-RE10K-iPhone~\cite{chen2026wildrayzer} & 50 & 50 & 0 & >2 & 4 & 1.8K & \NA & \checkmark \\
    \midrule
    \rowcolor{gray!15}
    \multicolumn{9}{l}{\textit{Public}} \\
    DeGauss~\cite{wang2025degauss} & 4 & 4 & 0 & >5 & 4 & >11K & \NA & \xmark \\
    RobustNeRF~\cite{sabour2023robustnerf} & 5 & 5 & 0 &  4 & 4 & 1.6K & 12MP & \checkmark \\
    On-the-go~\cite{Ren2024NeRF} & 12 & 2 & 10 & 14 & 10 & 2.2K & 12MP & \checkmark\\
    D-RE10K-iPhone~\cite{chen2026wildrayzer} & 50 & 50 & 0 & >2 & 4 & 1.8K & \NA & \checkmark \\
    RealX3D~\cite{liu2025realx3d} & 55 & 55 & 0 & \NA & \NA & \NA & 33MP & \checkmark \\
    DF3DV-1K (Ours) & 1K & 726 & 322 & 128 & 161 & 89.9K & $\sim$12MP & \checkmark \\
    \bottomrule
  \end{tabular}
\end{table}
These observations do not indicate shortcomings of existing benchmarks but rather reflect the rapid progress of the field, highlighting the need for a new, challenging, and systematic benchmark (see~\cref{fig:bench41}) better aligned with current method capabilities, which our work aims to address.
Similarly, the second issue can be observed by examining the development history of distractor-free radiance fields.
The first distractor-free radiance field method~\cite{sabour2023robustnerf}, explicitly targeting images captured over short time spans, was introduced in 2023 together with a benchmark, followed by several subsequent approaches~\cite{chen2024nerfhugs, otonari2024entity, ungermann2024robust, wang2024distractor, Ren2024NeRF, kulhanek2024wildgaussians, markin2024t}.
Among them,~\cite{Ren2024NeRF} introduced a new paradigm and benchmark that influenced later research.
Subsequently, many methods emerged~\cite{kong2025rogsplat, lin2025hybridgs, li2025modeling, ling2025ocsplats, wang2025degauss, li2025asymgs, 2025RobustSplat, sabour2025spotlesssplats, wang2025desplat, bao2025seg, fu2025robustsplat++, li2025wild3a, cao2025universe}, yet most require per-scene optimization.
Only recently, in 2026, a generalizable radiance field model fine-tuned on synthetic and small-scale real-world datasets~\cite{Ren2024NeRF} has been explored in this setting~\cite{bao2024distractor}.
This delay highlights the need for a large-scale real-world dataset with clean and cluttered images per scene to enable research beyond scene-specific approaches.

To facilitate progress, we present \textbf{DF3DV-1K (\underline{D}istractor-\underline{F}ree \underline{3D} \underline{V}ision \underline{1K})}, a newly collected dataset designed for distractor-free novel view synthesis. 
The dataset comprises clean and cluttered images for each scene. 
In total, DF3DV-1K contains \textbf{1,048} indoor and outdoor scenes spanning \textbf{128} distractor types and \textbf{161} scene themes, with \textbf{89,924} images overall.
To support and challenge distractor-free radiance field research focusing on casually captured images, we mimic real-world capture conditions by systematically collecting data using consumer cameras (see~\cref{fig:bench41,fig:pipeline}).
%smartphones with a small portion of scenes captured using drones (see~\cref{fig:bench41,fig:pipeline}). 
%The data collection process spans nine months to reach over one thousand scenes (see~\cref{fig:distribution1}).
DF3DV-1K represents a \textbf{large-scale, real-world, and diverse dataset with clean and cluttered images per scene} for this research area (see~\cref{tab:df_datasets}).
This extensive effort establishes a large-scale benchmark for distractor-free novel view synthesis.
Using this benchmark, we evaluate nine recent open-source distractor-free radiance field methods~\cite{sabour2025spotlesssplats,2025RobustSplat,ling2025ocsplats,markin2024t,kulhanek2024wildgaussians,wang2025desplat,wang2025degauss,li2025asymgs} together with 3DGS~\cite{kerbl20233d} across all 1,048 scenes. 
The large-scale evaluation identifies AsymGS~\cite{li2025asymgs} and RobustSplat~\cite{2025RobustSplat} as the most robust methods, followed by OCSplats~\cite{ling2025ocsplats} and DeGauss~\cite{wang2025degauss} as the second-best methods (see~\cref{tab:df_avg_benchmark} and~\cref{fig:rank}).
%This ranking differs from prior benchmarks, highlighting the importance of large-scale evaluation for reliably assessing robustness.
Interestingly, the ranking follows the publication timeline, particularly in PSNR and SSIM, reflecting a consistent and reasonable progression in performance as more recent methods advance the field.
These methods are further benchmarked on DF3DV-41, a subset of DF3DV-1K consisting of 41 systematically designed capture scenes covering 17 distractor and scene scenarios (see~\cref{fig:bench41} and~\cref{fig:bench_41_all_1,fig:bench_41_all_2}).
This analysis identifies the most challenging cases (e.g., semantically similar and fluid distractors and nighttime scenes) and shows that AsymGS~\cite{li2025asymgs}, RobustSplat~\cite{2025RobustSplat}, and OCSplats~\cite{ling2025ocsplats} remain comparatively robust, while other methods degrade under certain scenarios (see~\cref{tab:bench41_psnr,tab:bench41_ssim,tab:bench41_lpips}).
\begin{figure}[tb]
  \centering
  \includegraphics[width=0.99\linewidth]{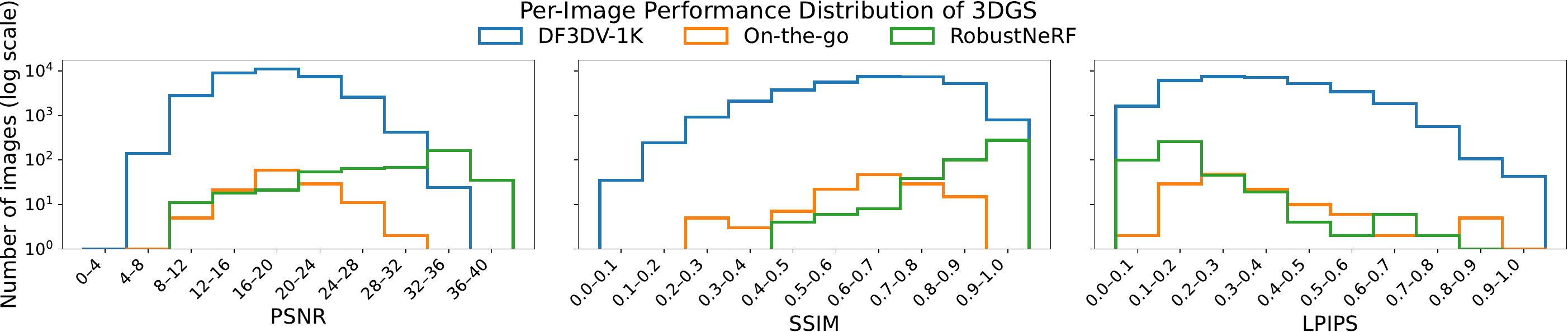}
  \caption{
\textbf{Dataset difficulty analysis via per-image performance distributions.}
%Per-image performance distributions of 3DGS~\cite{kerbl20233d} evaluated on DF3DV-1K, On-the-go~\cite{Ren2024NeRF}, and RobustNeRF~\cite{sabour2023robustnerf}.
DF3DV-1K shows wider LPIPS and SSIM distributions, indicating greater diversity and increased difficulty.
In contrast, RobustNeRF~\cite{sabour2023robustnerf} is relatively clean, where a vanilla 3DGS~\cite{kerbl20233d} frequently achieves PSNR values exceeding 30 dB.}
\label{fig:3dgs_distribution}
\end{figure}
In addition, the results in~\cref{fig:main_vis} and~\cref{fig:vis_41_1,fig:vis_41_2,fig:vis_41_3,fig:vis_41_4} reveal that existing approaches, while demonstrating promising performance on prior datasets, still leave room for improvement in challenging scenarios.
%Finally, to promote progress beyond scene-specific methods, we split DF3DV-1K into DF3DV-1K* as a training set and DF3DV-41 as a challenging test set. 
Finally, to promote progress beyond scene-specific methods, we apply DIFIX~\cite{wu2025difix3d+}, a 2D enhancement framework for sparse-view radiance fields, to distractor-free radiance fields by fine-tuning on DF3DV-1K* (e.g., 1,007 scenes) and evaluating on DF3DV-41 and the On-the-go dataset~\cite{Ren2024NeRF}. 
The resulting model, referred to as DI$^2$FIX (Distractor-Free DIFIX), achieves improvements of 0.96 dB PSNR and 0.057 LPIPS.
The contributions are: 
(1) DF3DV-1K, a large-scale, real-world, and diverse dataset for distractor-free novel view synthesis containing 1,048 indoor and outdoor scenes with clean and cluttered images spanning 128 distractor types and 161 scene themes. 
(2) A comprehensive benchmark evaluating nine recent distractor-free radiance field methods together with 3DGS, enabling systematic robustness analysis across methods and scenarios. 
(3) DI$^2$FIX, an enhancer that leverages large-scale data to improve distractor-free radiance field rendering quality, achieving gains of 0.96 dB PSNR and a 0.057 reduction in LPIPS.

% The contributions of this work are as follows:
% \begin{itemize}
%     \item \textbf{Dataset.} We introduce DF3DV-1K, a large-scale real-world dataset designed for distractor-free novel view synthesis, comprising 1,048 indoor and outdoor scenes with clean and cluttered image sets, covering 128 distractor types and 161 scene themes collected over nine months using consumer cameras.
%     \item \textbf{Benchmark.} We establish a large-scale benchmark for distractor-free radiance fields by evaluating nine recent distractor-free radiance field methods together with vanilla 3D Gaussian Splatting across all scenes, providing a systematic and comprehensive evaluation and identifying the most robust methods and scenarios.
%     \item \textbf{Algorithm.} We demonstrate how large-scale data enables progress beyond scene-specific reconstruction by presenting DI$^2$FIX, a 2D enhancer for improving distractor-free radiance field methods, achieving improvements of 0.9 dB in PSNR and 0.055 in LPIPS.
% \end{itemize}

\section{Related Work}
\label{sec:related}

\subsection{Radiance Fields}

%\textbf{Radiance Field Methods.}
Radiance Fields~\cite{mildenhall2021nerf,kerbl20233d} are widely used for novel view synthesis due to their photorealistic rendering capabilities. 
%To overcome the limitations of the methods~\cite{mildenhall2021nerf,kerbl20233d}, several research directions have been proposed. 
One line extends radiance fields from scene-specific to generalizable models~\cite{houchens2022neuralodf,yu2021pixelnerf, wang2021ibrnet, charatan2024pixelsplat, chen2024mvsplat, dey2025hfgaussian, ziwen2025long, kang2025ilrm, zheng2024gps, xu2024depthsplat, bao2024distractor, zhou2025gps, hosseinzadeh2025g3splat, jiang2025anysplat}.
%, which eliminate test-time optimization.
%by training on large-scale datasets to learn scene priors. 
Another line focuses on dynamic radiance fields~\cite{wu2022d, cao2023hexplane, wang2023mixed, song2023nerfplayer, fridovich2023k, wang2025degauss}, enabling novel view synthesis for 4D scenes.
\begin{figure}[tb]
  \centering
  \includegraphics[width=0.99\linewidth]{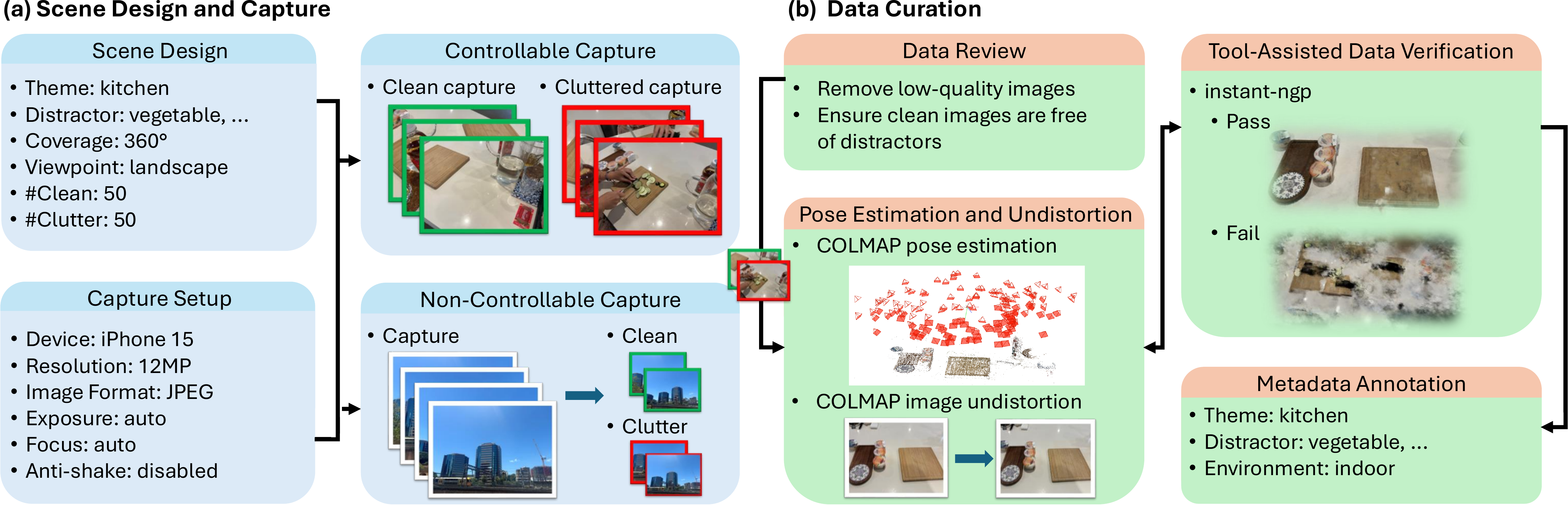}
  \caption{
\textbf{Overview of the data collection and curation pipeline.}
\textbf{(a) Scene design and capture.}
We design scenes with predefined themes, distractor types, coverage (e.g., $180^\circ$-$360^\circ$), viewpoint orientation (e.g., landscape or portrait), and the number of images per scene.
Images are captured using consumer cameras (e.g., iPhone).
Camera settings such as exposure and focus are set to automatic by default and adjusted only when necessary.
Anti-shake settings are disabled to ensure consistent image resolution within each scene.
For controllable setups, clean and cluttered scenes are captured separately.
Otherwise, they are captured simultaneously and later separated by operators.
\textbf{(b) Data curation.}
Low-quality images (e.g., defocused samples) are removed through manual inspection by two experts per scene.
Then, camera poses are jointly estimated using COLMAP~\cite{schoenberger2016mvs, schoenberger2016sfm}, followed by image undistortion.
Next, scenes are reconstructed with instant-ngp~\cite{muller2022instant} for verification.
Failed cases are reprocessed with adjusted COLMAP~\cite{schoenberger2016mvs, schoenberger2016sfm} parameters and discarded if reconstruction remains unsuccessful.
Finally, metadata are manually annotated.}
  \label{fig:pipeline}
\end{figure}
A separate line investigates in-the-wild radiance fields with large temporal gaps~\cite{martin2021nerf, dahmani2024swag, xu2024wild, zhang2024gaussian, wang2024we, chen2022hallucinated, chen2024nerfhugs, wang2024distractor, kulhanek2024wildgaussians, li2025asymgs, wang2025desplat, bao2025seg, fu2025robustsplat++, li2025wild3a, wang2025ie, wu2025wildsplatting, mithun2025diffusion}, which typically aim to render novel views from images captured across seasons.
Among the variants, distractor-free radiance fields~\cite{sabour2023robustnerf, chen2024nerfhugs, otonari2024entity, wang2024distractor, Ren2024NeRF, cao2025universe, 2025RobustSplat, kong2025rogsplat, kulhanek2024wildgaussians, li2025asymgs, li2025modeling, lin2025hybridgs, ling2025ocsplats, sabour2025spotlesssplats, wang2025degauss, wang2025desplat, bao2025seg, fu2025robustsplat++, markin2024t, li2025wild3a, ungermann2024robust, bao2024distractor, zheng2025wildgs, lee2025freeze, park2026forestsplats, tang2024nexussplats} demonstrate the ability to synthesize clean novel views from casually captured images collected over short time spans\footnote{As this survey focuses on general scenarios, multi-modal~\cite{li2025egosplat, rematas2022urban, prabakaran2026semantic}, task-specific~\cite{zheng2025up, trevithick2025simvs, du2025rge, Nerfbusters2023}, and domain-specific~\cite{huang2025skysplat, tang2025dronesplat, zhu2023occlusion, li2024derainnerf, liu2025deraings, bai2025satgs} approaches are excluded.}.
Although the boundaries between these lines~\cite{wang2025degauss, chen2024nerfhugs, wang2024distractor, kulhanek2024wildgaussians, li2025asymgs, wang2025desplat, bao2025seg, fu2025robustsplat++, li2025wild3a} may overlap, we emphasize that each line typically targets a specific data domain (see~\cref{fig:qual_three}). 
%Notably, the boundaries between these research directions~\cite{wang2025degauss, chen2024nerfhugs, wang2024distractor, kulhanek2024wildgaussians, li2025asymgs, wang2025desplat, bao2025seg, fu2025robustsplat++, li2025wild3a} may overlap. 
%For example, DeGauss~\cite{wang2025degauss}, categorized as a distractor-free radiance field method, also models dynamic scenes. 
%Similarly, several distractor-free approaches~\cite{chen2024nerfhugs, wang2024distractor, kulhanek2024wildgaussians, li2025asymgs, wang2025desplat, bao2025seg, fu2025robustsplat++, li2025wild3a} can also handle in-the-wild images with long temporal gaps.
%However, we emphasize that each research direction typically targets a specific data domain (see~\cref{fig:qual_three}). 
%Dynamic radiance fields usually assume densely captured sequential inputs, while in-the-wild radiance fields operate on data collected across long time spans, such as web-crawled images. 
%In contrast, distractor-free radiance fields focus on casually captured data collected within short time spans.
Most existing distractor-free radiance field methods remain scene-specific and require per-scene optimization. 
Only recently, the first generalizable distractor-free 3DGS~\cite{bao2024distractor} was introduced by fine-tuning a pretrained generalizable model~\cite{chen2024mvsplat} on synthetic data and a small-scale real-world dataset~\cite{Ren2024NeRF}. 
This delayed emergence of generalizable methods underscores the need for a large-scale, real-world, distractor-free dataset to enable scalable training.
\begin{figure}[t]
  \centering
    \includegraphics[height=0.115\textheight]{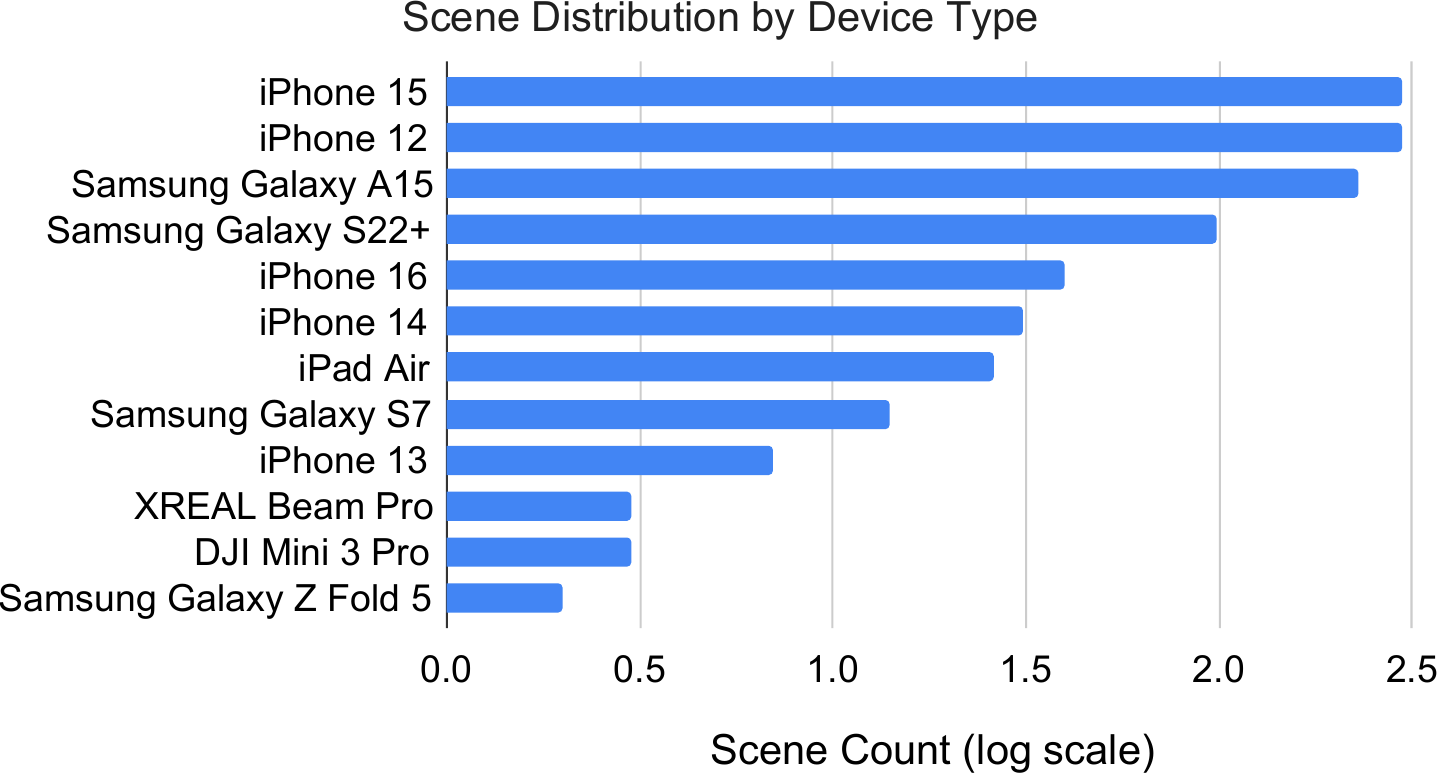}
  \hfill
    \includegraphics[height=0.115\textheight]{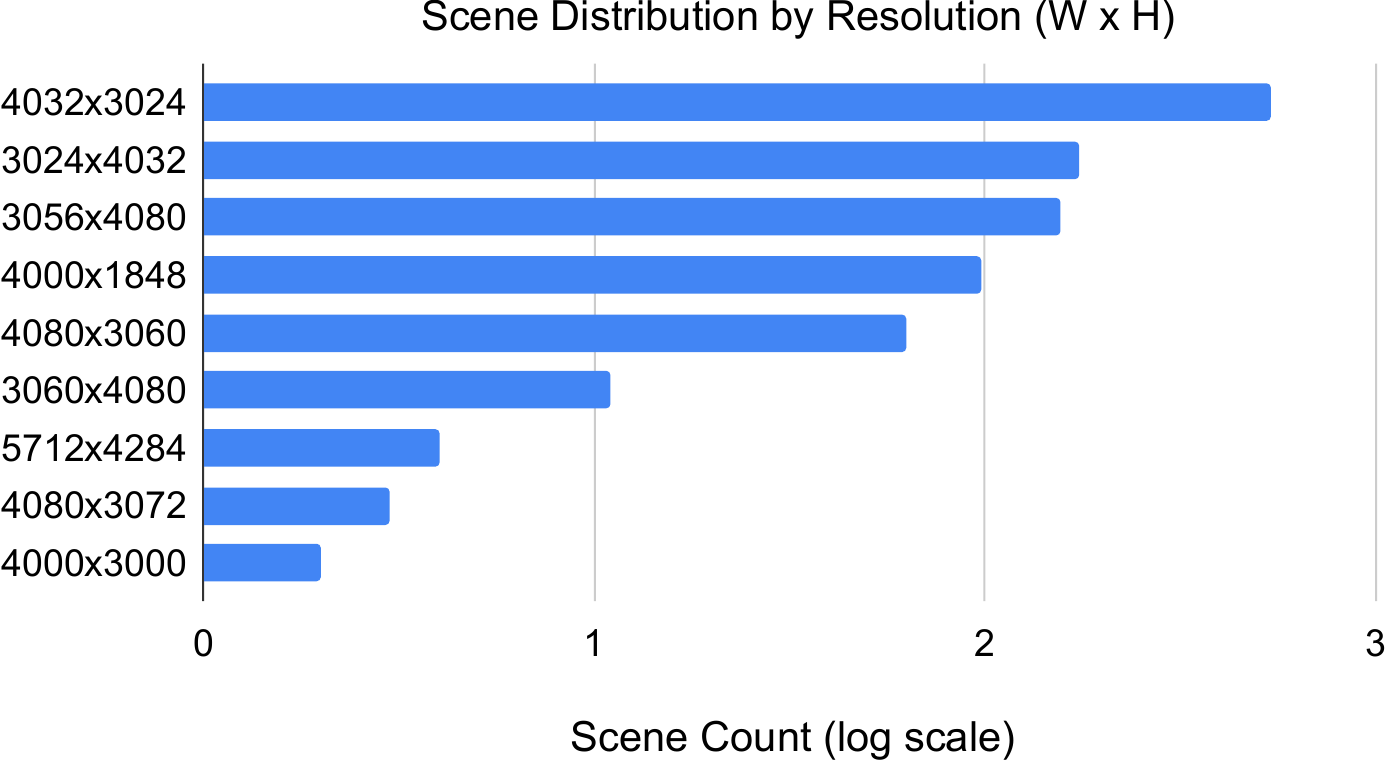}
  \hfill
    \includegraphics[height=0.115\textheight]{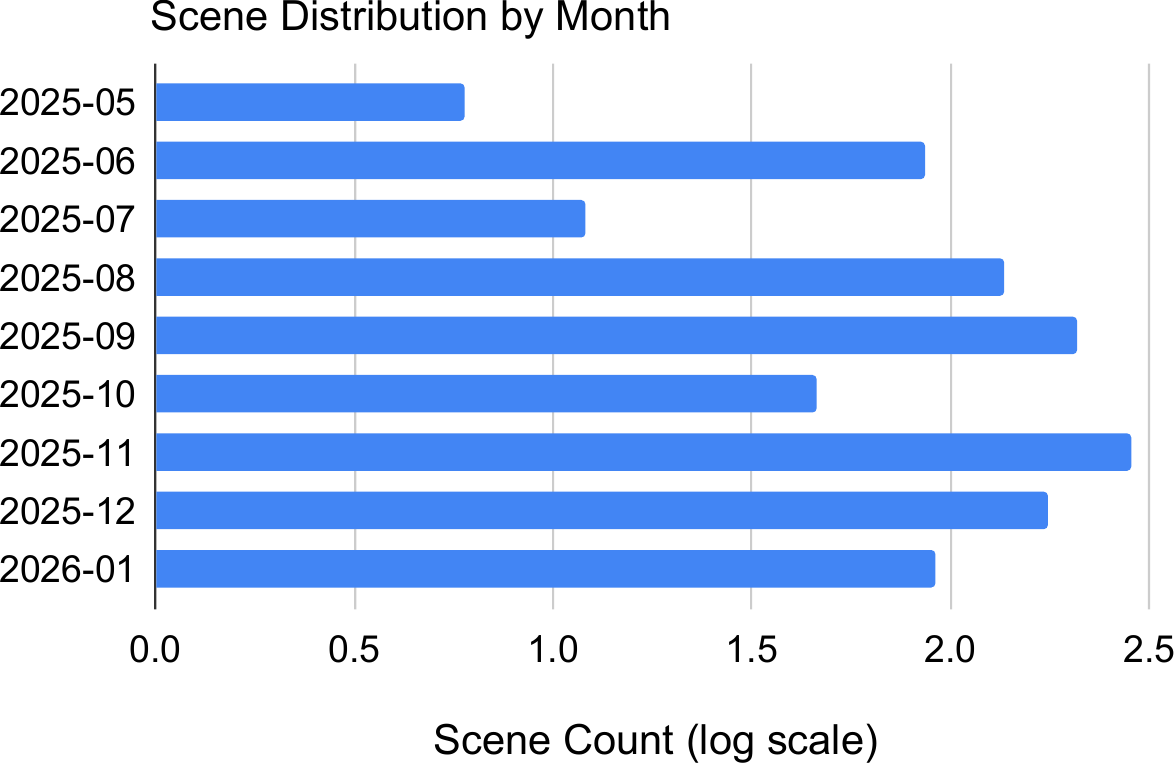}
  \caption{
\textbf{Distribution of DF3DV-1K by capture device, resolution, and month of data collection.}
The distributions highlight the diversity of acquisition settings.
}
  \label{fig:distribution1}
\end{figure}
Meanwhile, the rapid growth of distractor-free radiance field research (see~\cref{fig:papers}) underscores the need for a large-scale, challenging dataset and benchmark for systematic evaluation across methods.
%Furthermore, the delayed emergence of generalizable distractor-free radiance fields underscores the necessity of a large-scale real-world distractor-free dataset to support scalable training beyond scene-specific optimization.

%\subsection{Dataset collection methods}
% synthetic data
% manual capture data
% machine capture data
% web collected data
% data from existing dataset

\subsection{Datasets for Distractor-Free Radiance Fields}

%\textbf{3D Vision Datasets.} 
% Although many large-scale datasets~\cite{jensen2014large, zhou2018stereo, liu2021infinite, reizenstein2021common, yeshwanth2023scannet++, ling2024dl3dv, tung2024megascenes, kastingschafer2025seed4d} have been proposed for 3D vision tasks (see~\cref{tab:reg_datasets}), most are designed for vanilla 3D problems. 
% %such as static scene reconstruction, with some~\cite{kastingschafer2025seed4d, tung2024megascenes} targeting specific application domains. 
% No real-world large-scale dataset with clean and cluttered images per scene has been designed for distractor-free radiance field research (see~\cref{tab:df_datasets}).
% Public distractor-free datasets remain limited in both scale and diversity.
RobustNeRF~\cite{sabour2023robustnerf} introduces an indoor, tabletop-scale dataset consisting of five scenes spanning four themes and four distractor types. 
Each scene provides both clean and cluttered image sets, in which distractor objects are manually repositioned by operators between captures.
%, with images acquired using commercial cameras (e.g., smartphones) as well as a robotic arm equipped with a camera sensor. 
%This dataset has been widely adopted for benchmarking distractor-free radiance field methods. 
Although it served as the first widely adopted dataset and benchmark, it has become less challenging for recent state-of-the-art methods, making it difficult to clearly distinguish among their performance differences. 
%In fact, even vanilla 3DGS~\cite{kerbl20233d} can achieve PSNR values exceeding 30 dB on several views of the dataset (see~\cref{fig:3dgs_distribution}).
Another widely used dataset is the On-the-go dataset~\cite{Ren2024NeRF}, which contains twelve scenes, including two indoor and ten outdoor scenes. 
%captured using commercial cameras such as smartphones and drones. 
The dataset spans ten scene themes and fourteen distractor types. 
Among these, six scenes provide clean images, enabling both qualitative and quantitative benchmarking.
This dataset has been widely adopted because it extends the evaluation setting to outdoor environments and presents more challenging scenes (see~\cref{fig:3dgs_distribution}). 
Despite being more challenging, as discussed in~\cref{sec:intro}, it has also become saturated for recent methods, requiring many approaches to emphasize subtle visual differences in background regions to demonstrate performance gains.
%Some private datasets~\cite{otonari2024entity, cao2025universe, li2025wild3a} or unpublished datasets~\cite{markin2024t, liu2025realx3d, chen2026wildrayzer}, which are not publicly accessible before submission, are proposed. 
Recently, two amazing concurrent works, RealX3D~\cite{liu2025realx3d} and D-RE10K-iPhone~\cite{chen2026wildrayzer}, provide paired clean and cluttered images captured with rail-mounted professional cameras and sparse-view indoor settings, respectively.
As shown in~\cref{fig:teaser,tab:df_datasets}, DF3DV-1K stands out by providing a large-scale collection of data with clean and cluttered images per scene, greater diversity, support for both qualitative and quantitative benchmarking, and more challenging, systematically designed scenarios under casual capture settings.
\section{Data Acquisition and Curation}
\label{sec:method}

\subsection{Data Acquisition}

\textbf{Scene Design.} The goal is to construct a large-scale dataset with clean and cluttered images per scene for distractor-free radiance fields. 
%To mimic real-world casual capture conditions, operators collect data using consumer cameras.
Operators (see~\cref{fig:pipeline}) first predefine each scene by specifying the scene theme (e.g., kitchen), potential distractors (e.g., vegetables), viewpoint coverage ranging from 180$^\circ$ to 360$^\circ$, viewpoint orientation (e.g., landscape or portrait), and the approximate number of clean and cluttered images to be captured. 
The indoor and outdoor capture protocols follow prior works~\cite{sabour2023robustnerf, Ren2024NeRF}. 
For indoor scenes, distractors are manually introduced. 
In contrast, for uncontrolled environments where distractors naturally exist, no or limited additional distractors are introduced. \\
\textbf{Capture Setup.} Operators select a device configured to capture $\sim$12MP JPEG images with automatic exposure and focus. 
Images are taken individually rather than in video mode, and anti-shake is disabled to maintain consistent resolution. 
Operators may adjust exposure or focus when needed to prevent defocusing.\\
\textbf{Capture.} 
%During early data collection, approximately 300 scenes were captured using interval shooting with intervals of 1-5 seconds. 
%However, this approach occasionally produced motion blur, resulting in substantial manual effort during data curation to remove affected images. 
%Consequently, for the remaining data collection, images were captured manually frame by frame, ensuring higher image quality while significantly reducing post-processing and curation time.
For controllable scenes, operators capture clean and cluttered images separately. 
For uncontrollable scenes, both conditions are captured together and later manually separated.
Operators are instructed to pause capture during windy conditions.
This manual capture process closely mimics casual image acquisition and thereby aligns with our dataset design goals.
Furthermore, collecting data ourselves rather than sourcing images from existing data enables the customized design of challenging scenarios (see~\cref{fig:bench41} and~\cref{fig:bench_41_all_1,fig:bench_41_all_2}).

\subsection{Data Curation}

\textbf{Data Review.} Operators manually remove low-quality images (e.g., motion blur in static regions) and ensure the clean image set contains no distractors (see~\cref{fig:pipeline}).
%The reviewed data are then further proofread by a professional with a computer vision background to ensure consistency and quality. 
We allow distractors themselves to exhibit lower visual quality (see~\cref{fig:low_q}), as such artifacts naturally occur in casual capture scenarios. \\
\textbf{Pose Estimation and Undistortion.} Following common practice~\cite{sabour2023robustnerf, Ren2024NeRF}, we use COLMAP~\cite{schoenberger2016mvs, schoenberger2016sfm} to jointly estimate camera poses from clean and cluttered images, leveraging reliable correspondences from clean views to improve robustness and enforce a shared camera coordinate system.
We then use these camera parameters to undistort all images, producing data used in our benchmark. \\
\textbf{Tool-Assisted Data Verification.}
We reconstruct each scene using instant-ngp~\cite{muller2022instant} to verify data quality. 
Failure cases typically include missing geometry, degenerate solutions, or fragmented reconstructions. 
When failures occur, we adjust COLMAP~\cite{schoenberger2016mvs, schoenberger2016sfm} parameters (e.g., minimum inlier threshold, registration trials, and minimum model size) to improve pose registration and avoid degeneracy. 
Scenes that still fail after tuning are discarded. 
In practice, most scenes pass verification directly, with only a few requiring additional tuning. \\
\textbf{Annotation.} We annotate passed scenes with metadata that requires manual labeling, including the scene theme, distractor type, scenario, and environment.

% \begin{figure}[tb]
%   \centering
%   \includegraphics[width=0.99\linewidth]{figure/rank_alex.pdf}
%   \caption{
% \textbf{Top-4 performing methods on each dataset.}
% Results on DF3DV-1K identify AsymGS~\cite{li2025asymgs} and RobustSplat~\cite{2025RobustSplat} as the most robust methods, followed by OCSplats~\cite{ling2025ocsplats} and DeGauss~\cite{wang2025degauss}.
% This ranking differs from those observed on On-the-go~\cite{Ren2024NeRF} and RobustNeRF~\cite{sabour2023robustnerf}, highlighting the importance of large-scale evaluation.
% Interestingly, performance trends on DF3DV-1K largely follow publication chronology, with AsymGS~\cite{li2025asymgs} (NeurIPS 2025) outperforming RobustSplat~\cite{2025RobustSplat}, OCSplats~\cite{ling2025ocsplats}, and DeGauss~\cite{wang2025degauss} (ICCV 2025), while the three ICCV 2025 methods exhibit similar performance in PSNR and SSIM.
% Such trends are less evident on existing datasets.
% }
% \label{fig:rank}
% \end{figure}

\subsection{Data Statistics}
\begin{figure}[tb]
  \centering
  \includegraphics[width=0.99\linewidth]{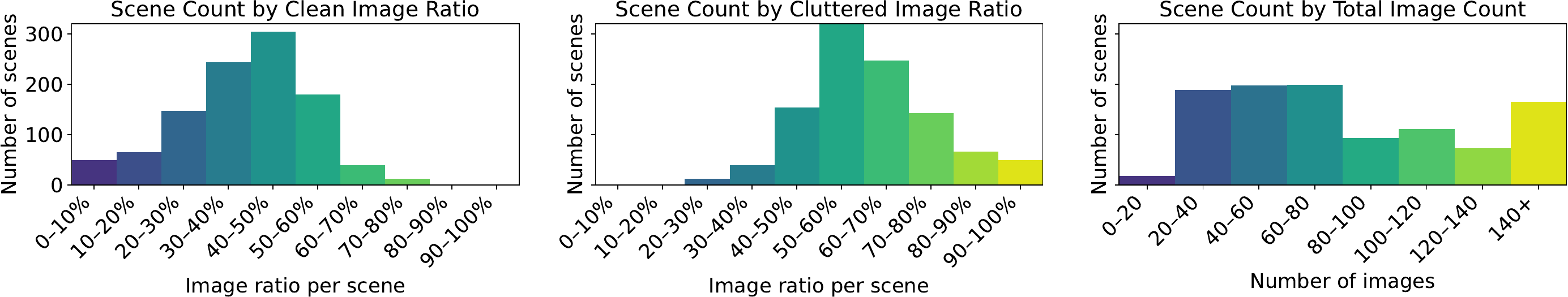}
  \caption{
\textbf{Scene count distribution.}
Clean images per scene skew slightly toward lower bins for efficient novel-view benchmarking, whereas cluttered images extend toward higher bins to capture diverse distractor conditions. Total images per scene remain comparable to typical non-sparse radiance field settings.}
  \label{fig:image_count_distribution}
\end{figure}
\textbf{Scale.}
DF3DV-1K contains 1,048 scenes (726 indoor, 322 outdoor) with clean and cluttered image sets, totaling 89,924 images (see~\cref{tab:df_datasets}).
%Most scenes contain 20-40 clean images (see~\cref{fig:image_count_distribution}), skewed toward lower counts for efficient benchmarking, while cluttered images exhibit a broader distribution due to diverse distractor conditions. 
The clean image ratio (see~\cref{fig:image_count_distribution}) per scene skews slightly toward lower ratios for efficient benchmarking, while the cluttered image ratio shifts toward higher ratios to cover more diverse distractors. Overall, the total number of images per scene is centered around 50 images, comparable to commonly used radiance field datasets. \\
%Overall, the large number of scenes and images enables reliable evaluation and reduces performance variance across different scenes, supporting fair large-scale benchmarking of distractor-free radiance field methods and facilitating development beyond scene-specific methods. \\
\textbf{Semantic Diversity.}
DF3DV-1K covers 128 distractor types.
The types span from common objects (e.g., cars) to less frequent or large-scale distractors (e.g., planes).
In addition, DF3DV-41 (see~\cref{fig:bench41}) contains systematically designed distractor types (e.g., semantically similar distractors challenging feature-based methods) and scene themes.
See~\cref{fig:scene_and_theme,fig:scenario_dis} for detailed distributions. \\
\textbf{Capture Diversity.}
DF3DV-1K is collected over nine months.
The dataset includes data captured using 12 consumer cameras, ranging from iPhone to Samsung smartphones, covering nine different image resolutions (see~\cref{fig:distribution1}).
%This diversity in capture devices and acquisition conditions enables more realistic evaluation under varied real-world scenarios.
\section{Benchmarks \& Experiments}
\label{sec:exp}

\subsection{DF3DV-1K and DF3DV-41 Benchmarks}

\begin{table}[t]
\caption{
\textbf{Benchmark results on RobustNeRF~\cite{sabour2023robustnerf}, On-the-go~\cite{Ren2024NeRF}, DF3DV-1K, and DF3DV-41.}
RobustNeRF~\cite{sabour2023robustnerf} is the least challenging benchmark, where all methods achieve strong performance, with several methods exceeding 29 dB PSNR, comparable to results on clean benchmarks~\cite{kulhanek2025nerfbaselines}.
Compared with RobustNeRF~\cite{sabour2023robustnerf}, On-the-go~\cite{Ren2024NeRF} is more challenging due to the inclusion of outdoor scenes.
DF3DV-1K and DF3DV-41 are the most challenging benchmarks, featuring large-scale data with diverse distractor scenarios.
On DF3DV-1K and DF3DV-41, AsymGS~\cite{li2025asymgs} and RobustSplat~\cite{2025RobustSplat} are the most robust, followed by OCSplats~\cite{ling2025ocsplats} and DeGauss~\cite{wang2025degauss}.
}
\label{tab:df_avg_benchmark}
\centering
%\tiny
\begin{adjustbox}{width=\linewidth}
\begin{tabular}{lc|ccc|ccc|ccc|ccc}
\toprule
\multirow{2}{*}{\textbf{Method}} &
\multirow{2}{*}{\textbf{Venue}}
& \multicolumn{3}{c|}{\textbf{RobustNeRF~\cite{sabour2023robustnerf}}}
& \multicolumn{3}{c|}{\textbf{On-the-go~\cite{Ren2024NeRF}}}
& \multicolumn{3}{c|}{\textbf{DF3DV-1K}}
& \multicolumn{3}{c}{\textbf{DF3DV-41}} \\

& 
& PSNR$\uparrow$ & SSIM$\uparrow$ & LPIPS$\downarrow$
& PSNR$\uparrow$ & SSIM$\uparrow$ & LPIPS$\downarrow$
& PSNR$\uparrow$ & SSIM$\uparrow$ & LPIPS$\downarrow$
& PSNR$\uparrow$ & SSIM$\uparrow$ & LPIPS$\downarrow$ \\
\midrule

3DGS~\cite{kerbl20233d} & TOG'23
& 25.44 & 0.834 & 0.157
& 19.33 & 0.662 & 0.306
& 17.93 & 0.630 & 0.330
& 18.10 & 0.620 & 0.331 \\
\midrule

T-3DGS~\cite{markin2024t} & arXiv
& 28.46 & \third{0.890} & 0.098
& 22.80 & 0.818 & 0.128
& 18.92 & 0.682 & 0.263
& 19.00 & 0.672 & 0.259 \\

T-3DGS-TMR~\cite{markin2024t} & arXiv
& 26.79 & 0.872 & 0.121
& 19.82 & 0.736 & 0.231
& 18.11 & 0.657 & 0.296
& 18.41 & 0.654 & 0.279 \\

WildGaussians~\cite{kulhanek2024wildgaussians} & NeurIPS'24
& 27.25 & 0.872 & 0.130
& 22.60 & 0.793 & 0.155
& 18.78 & 0.669 & 0.293
& 19.26 & 0.670 & 0.285 \\

SLS~\cite{sabour2025spotlesssplats} & TOG'25
& 28.70 & 0.873 & 0.103
& 22.73 & 0.779 & 0.153
& 19.75 & 0.694 & 0.263
& 19.37 & 0.662 & 0.287 \\

DeSplat~\cite{wang2025desplat} & CVPR'25
& 28.57 & 0.880 & \third{0.096}
& 22.74 & 0.805 & 0.121
& 19.26 & 0.681 & 0.249
& 19.31 & 0.662 & 0.248 \\

DeGauss~\cite{wang2025degauss} & ICCV'25
& \second{29.22} & \second{0.892} & \second{0.089}
& \best{23.57} & \best{0.831} & \best{0.107}
& 20.00 & \second{0.714} & \third{0.243}
& \second{19.98} & \third{0.695} & \second{0.236} \\

OCSplats~\cite{ling2025ocsplats} & ICCV'25
& 28.91 & 0.884 & \best{0.088}
& 22.98 & 0.817 & \second{0.108}
& \third{20.02} & \third{0.710} & 0.247
& 19.84 & 0.689 & 0.249 \\

RobustSplat~\cite{2025RobustSplat} & ICCV'25
& \third{29.19} & 0.889 & \second{0.089}
& \third{23.09} & \second{0.825} & \third{0.112}
& \second{20.13} & \second{0.714} & \best{0.226}
& \third{19.95} & \second{0.696} & \best{0.232} \\

AsymGS~\cite{li2025asymgs} & NeurIPS'25
& \best{29.44} & \best{0.896} & \third{0.096}
& \second{23.16} & \third{0.821} & 0.136
& \best{20.49} & \best{0.735} & \second{0.229}
& \best{20.35} & \best{0.712} & \third{0.247} \\

\bottomrule
\end{tabular}
\end{adjustbox}
\end{table}

\begin{figure}[tb]
  \centering
  \includegraphics[width=0.96\linewidth]{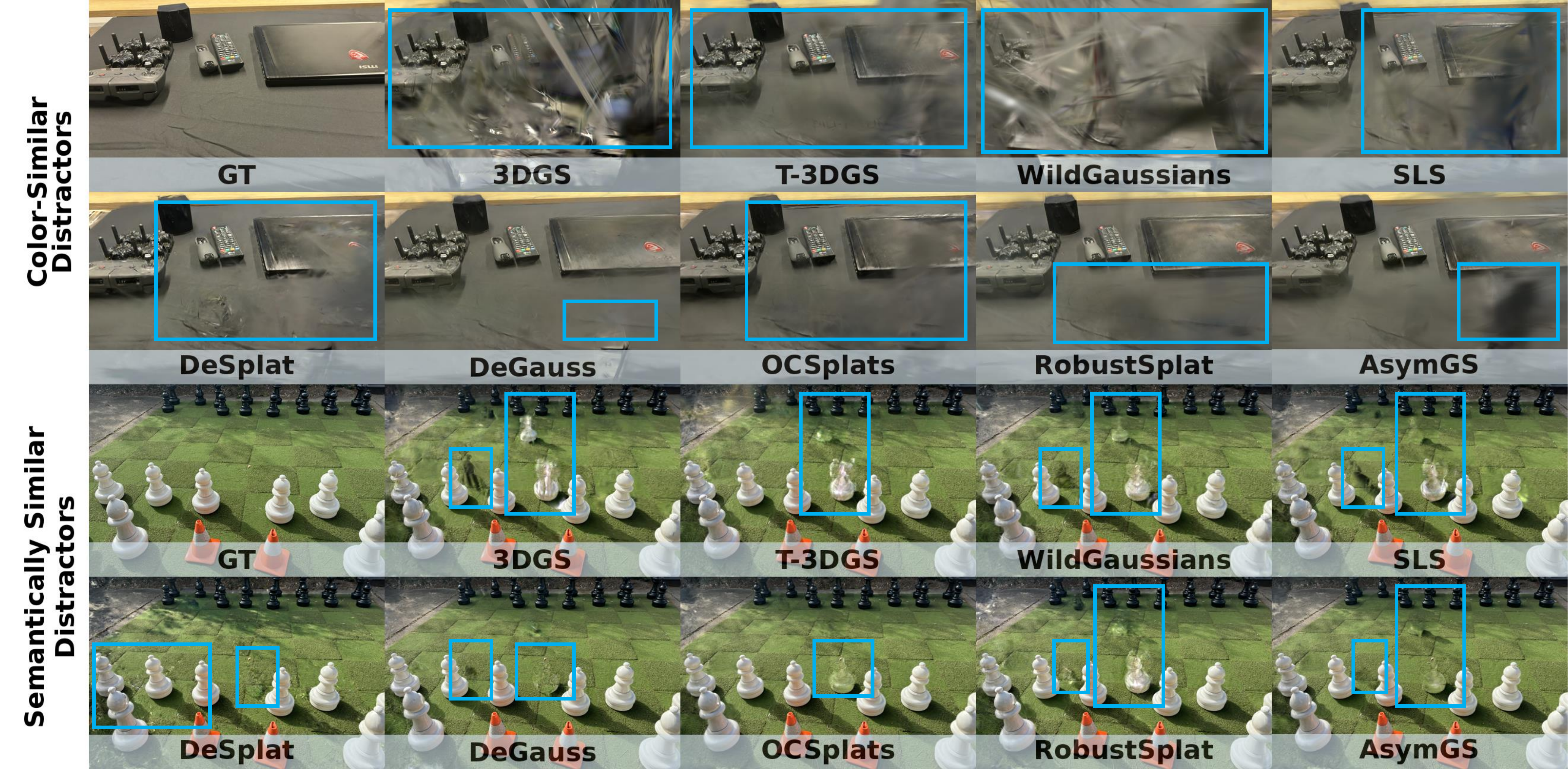}
  \caption{\textbf{Qualitative results of the radiance field methods on DF3DV-41.} 
The benchmark introduces systematically challenging conditions that enable clear visual comparison across methods and support the evaluation of robustness differences. 
For instance, WildGaussian~\cite{kulhanek2024wildgaussians} is less robust to color-similar distractors and tends to blend the black distractors with the background, while DeSplat~\cite{wang2025desplat} produces a noisier background in the chess scene.}
  \label{fig:main_vis}
\end{figure}

\textbf{Benchmark Methods.} We evaluate recent open-source distractor-free radiance field methods, including AsymGS~\cite{li2025asymgs}, RobustSplat~\cite{2025RobustSplat}, OCSplats~\cite{ling2025ocsplats}, DeGauss~\cite{wang2025degauss}, SLS~\cite{sabour2025spotlesssplats}, DeSplat~\cite{wang2025desplat}, WildGaussians~\cite{kulhanek2024wildgaussians}, T-3DGS and T-3DGS-TMR~\cite{markin2024t}, together with 3DGS~\cite{kerbl20233d}.
%on DF3DV-1K and DF3DV-41.
See~\cref{sec:bench} for more details.\\
%In total, we evaluate nine distractor-free radiance field methods alongside one vanilla 3DGS.\\
% \textbf{Experiment Details.} Some methods perform image undistortion on-the-fly using OpenCV~\cite{bradski2000opencv}, adopt different parameter settings for different scenes, or downsample images before undistortion. 
% To ensure a fair comparison, we standardize the preprocessing pipeline and training parameters across all methods and scenes. 
% Specifically, images are undistorted offline using COLMAP~\cite{schoenberger2016mvs, schoenberger2016sfm}, followed by downsampling by a factor of 8 after undistortion. 
% We apply the same parameter configuration to all scenes to prevent scene-specific tuning advantages. 
% We also observe that training for some methods, such as DeGauss~\cite{wang2025degauss}, can be unstable and occasionally crash for a small number of scenes. 
% In such cases, we reduce the learning rate to stabilize training and retrain the model for the affected scenes. \\
\textbf{Most Robust Methods on Large-Scale Benchmarks.} \cref{tab:df_avg_benchmark} shows that AsymGS~\cite{li2025asymgs} and RobustSplat~\cite{2025RobustSplat} are the most robust methods on DF3DV-1K and DF3DV-41, followed by OCSplats~\cite{ling2025ocsplats} and DeGauss~\cite{wang2025degauss} as the second most robust.
This trend aligns with the publication timeline, reflecting steady performance improvements as methods evolve and supporting benchmark reliability.
~\cref{tab:bench41_psnr,tab:bench41_ssim,tab:bench41_lpips} present the relative improvements of each method compared to 3DGS~\cite{kerbl20233d} across scenarios.
Among them, AsymGS~\cite{li2025asymgs}, RobustSplat~\cite{2025RobustSplat}, and OCSplats~\cite{ling2025ocsplats} show stronger robustness, consistently outperforming 3DGS~\cite{kerbl20233d}, while other methods degrade under specific scenarios.
\cref{fig:main_vis} and~\cref{fig:vis_41_1,fig:vis_41_2,fig:vis_41_3,fig:vis_41_4} show similar a trend, with AsymGS~\cite{li2025asymgs}, RobustSplat~\cite{2025RobustSplat}, OCSplats~\cite{ling2025ocsplats}, and DeGauss~\cite{wang2025degauss} performing relatively well across several scenarios. \\
\textbf{DF3DV-1K and DF3DV-41 vs. Prior Benchmarks.}
Compared with the public benchmarks~\cite{sabour2023robustnerf,Ren2024NeRF}, DF3DV-1K and DF3DV-41 are larger in scale and more diverse in distractor types and scene themes, discouraging per-scene tuning, and posing greater challenges.
\cref{tab:df_avg_benchmark} further confirms this trend. 
The RobustNeRF~\cite{sabour2023robustnerf} benchmark appears less challenging, as improvements over 3DGS~\cite{kerbl20233d} are modest while many methods still achieve PSNR above 29 dB, comparable to 3DGS performance on clean-scene benchmarks~\cite{kulhanek2025nerfbaselines}. 
In contrast, DF3DV-1K and DF3DV-41 are more challenging than On-the-go~\cite{Ren2024NeRF}, exhibiting lower PSNR and SSIM, higher LPIPS, and smaller gains over 3DGS~\cite{kerbl20233d} across methods.
~\cref{fig:rank} shows the top-4 most robust methods on RobustNeRF~\cite{sabour2023robustnerf}, On-the-go~\cite{Ren2024NeRF}, and DF3DV-1K. 
Rankings differ across datasets, underscoring the importance of large-scale and diverse benchmarking. 
Notably, rankings on DF3DV-1K roughly follow publication timelines, a trend less evident in prior benchmarks. 
Recent methods often rely on subtle background differences in qualitative comparisons on earlier benchmarks to highlight strengths or limitations (see~\cref{sec:intro}). 
This is not a shortcoming of existing benchmarks, but rather reflects rapid methodological progress and the need for more challenging evaluations. 
Our benchmark addresses this gap by enabling \textit{“zoom-in no more”} qualitative comparisons, where performance differences are directly observable, and providing significance analyses (see~\cref{sec:bench}).
%As shown in~\cref{fig:vis_41_1,fig:vis_41_2,fig:vis_41_3,fig:vis_41_4}, these challenging scenarios expose remaining limitations and clearly distinguish methods.
\\
\textbf{Limitations of Current Methods.}
~\cref{tab:bench41_psnr,tab:bench41_ssim,tab:bench41_lpips} show the relative improvement of each method relative to 3DGS~\cite{kerbl20233d} across scenarios.
The quantitative results show that semantically similar and fluid distractors are the most challenging distractor types, while nighttime scenes represent the most difficult scene type, as indicated by the lower 3DGS performance and smaller achievable improvements.
This difficulty arises because many methods rely on semantic features~\cite{oquab2024dinov2, ravisam, kirillov2023segment, tang2023emergent} to identify distractors.
When distractors share similar semantic meanings with static scene objects, distinguishing distractors becomes difficult (see~\cref{fig:main_vis} and~\cref{fig:vis_41_2}).
%The chess scene in \cref{fig:vis_41_2} further confirms this observation, where chess pieces that should be considered distractors partially remain in the reconstructed scene across many methods.
For fluid distractors, many methods learn masks online to remove dynamic content. 
However, fluid phenomena (e.g., splashes) are often spatially scattered and semi-transparent, which causes masking strategies to inadvertently remove large valid scene or fail to fully suppress the distractors (see blending artifacts in~\cref{fig:vis_41_1}).
%The washing car scene in \cref{fig:vis_41_1} further confirms this observation, where blending artifacts caused by fluid distractors spread across the scene.
For nighttime scenes, many approaches rely on fixed thresholds to filter distractors. 
Because these thresholds are typically tuned for daytime conditions, they become unreliable under low illumination, where distractors and static components are harder to distinguish (see~\cref{fig:vis_41_4}).
In addition, distractor removal may occasionally affect view-dependent effects on static objects, although~\cref{sec:bench} shows that the benefits generally outweigh this drawback.

\subsection{Beyond Scene-Specific Methods}

\begin{table}[t]
\begin{minipage}[t]{.32\textwidth}
\centering
  \caption{
\textbf{Effect of enhancers.}
Vanilla is the rendering result of the methods~\cite{kerbl20233d, li2025asymgs, wang2025degauss, wang2025desplat, ling2025ocsplats, 2025RobustSplat, sabour2025spotlesssplats, markin2024t, kulhanek2024wildgaussians}.
DIFIX+RobustNeRF and DI$^2$FIX are DIFIX~\cite{wu2025difix3d+} fine-tuned on RobustNeRF~\cite{sabour2023robustnerf} and DF3DV-1K*, respectively.
Results report mean performance relative to Vanilla on the On-the-go~\cite{Ren2024NeRF} and DF3DV-41.
}
  \label{tab:difix_overall_bench}
  \centering
  \resizebox{\columnwidth}{!}{%
  \begin{tabular}{l|ccc}
    \toprule
    \textbf{Method} & \textbf{PSNR$\uparrow$} & \textbf{SSIM$\uparrow$} & \textbf{LPIPS$\downarrow$} \\
    \midrule
    Vanilla & $\second{20.82}$ & $\second{0.731}$ & $\third{0.211}$ \\
    \midrule 
    DIFIX~\cite{wu2025difix3d+} & $20.16$ & $0.663$ & $0.223$ \\ 
    DIFIX+RobustNeRF & $\third{20.54}$ & $\third{0.695}$ & $\second{0.203}$ \\
    DI$^2$FIX (Ours) & $\best{21.78}$ & $\best{0.737}$ & $\best{0.154}$ \\
    %DIFIX~\cite{wu2025difix3d+} & $\bad{-0.66}$ & $\bad{-0.067}$ & $\bad{+0.012}$ \\
    %DIFIX+RobustNeRF & $\bad{-0.27}$ & $\bad{-0.036}$ & $\good{-0.008}$ \\
    %DI$^2$FIX (Ours) & $\good{+0.96}$ & $\good{+0.006}$ & $\good{-0.056}$ \\
    \bottomrule
  \end{tabular}
  }
\end{minipage}
% \hspace{1mm}
\begin{minipage}[t]{.32\textwidth}
\centering
  \caption{
\textbf{Effect of training data scale and diversity on DI$^2$FIX.}
We vary the number of DF3DV-1K* training scenes from 250, 500, 750, to 1,007 and report the mean performance on the On-the-go~\cite{Ren2024NeRF} and DF3DV-41.
Performance consistently improves as the training set size increases, and begins to plateau at larger scales.
}
  \label{tab:df3dv_size_abl}
  \centering
  \resizebox{\columnwidth}{!}{%
  \begin{tabular}{lc|ccc}
    \toprule
    \textbf{Method} & \textbf{Scenes} & \textbf{PSNR$\uparrow$} & \textbf{SSIM$\uparrow$} & \textbf{LPIPS$\downarrow$} \\
    \midrule
    \multirow{4}{*}{DI$^2$FIX} 
      & 250 & $21.42$ & $0.723$ & $0.169$\\
      & 500 & $\third{21.57}$ & $\third{0.731}$ & $\third{0.163}$ \\
      & 750 & $\second{21.69}$ & $\second{0.736}$ & $\second{0.159}$ \\
      & 1K & $\best{21.78}$ & $\best{0.737}$ & $\best{0.154}$ \\
    \bottomrule
  \end{tabular}
  }
\end{minipage}
% \hspace{1mm}
\begin{minipage}[t]{.32\textwidth}
\centering
  \caption{
\textbf{Effect of training data degradation level on DI$^2$FIX.}
We vary the LPIPS threshold used to select training pairs and report the mean performance on the On-the-go~\cite{Ren2024NeRF} and DF3DV-41.
Moderate thresholds yield the best performance, while overly strict or overly loose selection slightly reduces performance.
}
  \label{tab:df3dv_lpips_abl}
  \centering
  \resizebox{\columnwidth}{!}{%
  \begin{tabular}{lc|ccc}
    \toprule
    \textbf{Method} & \textbf{$\bm{\leq}$LPIPS} & \textbf{PSNR$\uparrow$} & \textbf{SSIM$\uparrow$} & \textbf{LPIPS$\downarrow$} \\
    \midrule
    \multirow{5}{*}{DI$^2$FIX} 
      & 0.1 & $21.55$ & $\second{0.734}$ & $0.166$\\
      & 0.3 & $21.64$ & $\best{0.737}$ & $\second{0.160}$ \\
      & 0.5 & $\best{21.78}$ & $\best{0.737}$ & $\best{0.154}$\\
      & 0.7 & $\third{21.70}$ & $\second{0.734}$ & $\third{0.161}$\\
      & 0.9 & $\second{21.73}$ & $\third{0.731}$ & $0.165$\\
    \bottomrule
  \end{tabular}
  }
\end{minipage}
\end{table}

Distractor-free radiance field methods show promising results but still exhibit limitations (see~\cref{fig:main_vis} and~\cref{fig:vis_41_1,fig:vis_41_2,fig:vis_41_3,fig:vis_41_4}), motivating the question of whether performance can be improved by enhancing existing methods without model modifications or scene-specific tuning. 
We therefore conduct a pilot experiment demonstrating how DF3DV-1K facilitates generalizable solutions. \\
\textbf{Experimental Details.} 
We introduce \textbf{DI$\bm{^2}$FIX} (Distractor-Free DIFIX), a plug-and-play 2D enhancer built on DIFIX~\cite{wu2025difix3d+} to improve radiance field renderings. 
%DIFIX~\cite{wu2025difix3d+} is an image-to-image diffusion model based on~\cite{sauer2024adversarial} and fine-tuned using the training strategy of~\cite{parmar2024one}. 
DIFIX~\cite{wu2025difix3d+} is a diffusion-based model~\cite{sauer2024adversarial,parmar2024one} that enhances radiance field renderings under sparse input using a clean reference view and degraded target view.
To adapt DIFIX~\cite{wu2025difix3d+} to distractor-free tasks, we replace the clean reference view with a radiance field rendering (see~\cref{sec:d2fix} for details).
We construct paired training data by reconstructing scenes from cluttered images in DF3DV-1K* (excluding DF3DV-41) using radiance field methods~\cite{kerbl20233d,2025RobustSplat,ling2025ocsplats,markin2024t,kulhanek2024wildgaussians,wang2025desplat,wang2025degauss,li2025asymgs} and rendering novel views at clean-image views. 
This yields 316,890 candidate pairs, from which those with LPIPS$\leq \gamma$ are retained, where $\gamma=0.5$ in the final setting.
We then fine-tune DIFIX~\cite{wu2025difix3d+} into DI$\bm{^2}$FIX.
\begin{figure}[t]
\centering
\includegraphics[width=0.99\linewidth]{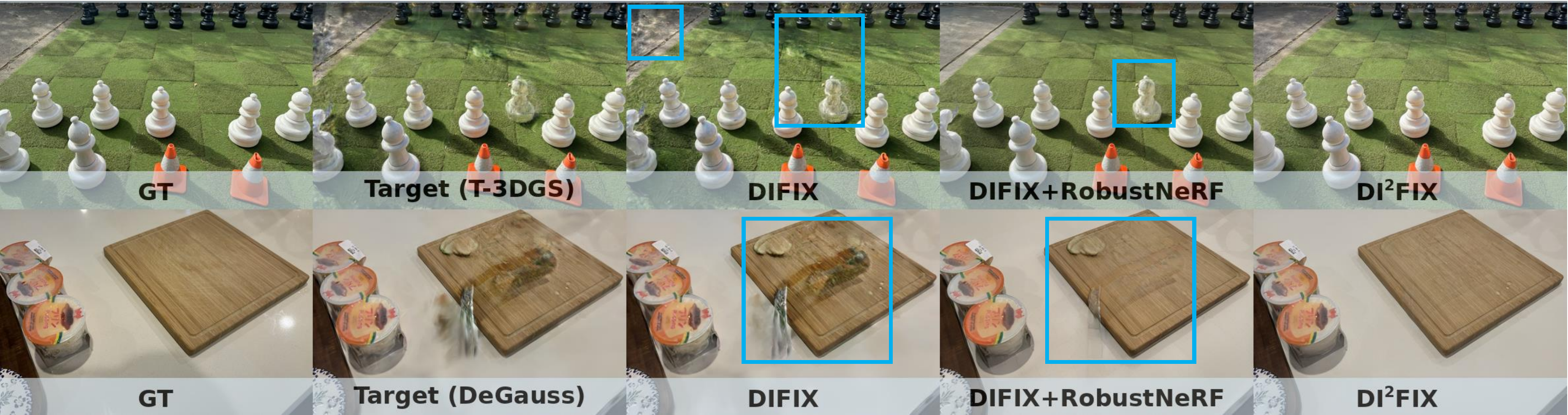}
\caption{
\textbf{Qualitative results of enhancers.}
Leveraging DF3DV-1K*, a large-scale and diverse dataset, DI$^2$FIX effectively removes distractor artifacts (e.g., dynamic chess pieces and vegetable artifacts) while inpainting occluded regions in target views.
}
\label{fig:main_fix_vis}
\end{figure}
\begin{figure}[t]
\centering
\includegraphics[width=0.99\linewidth]{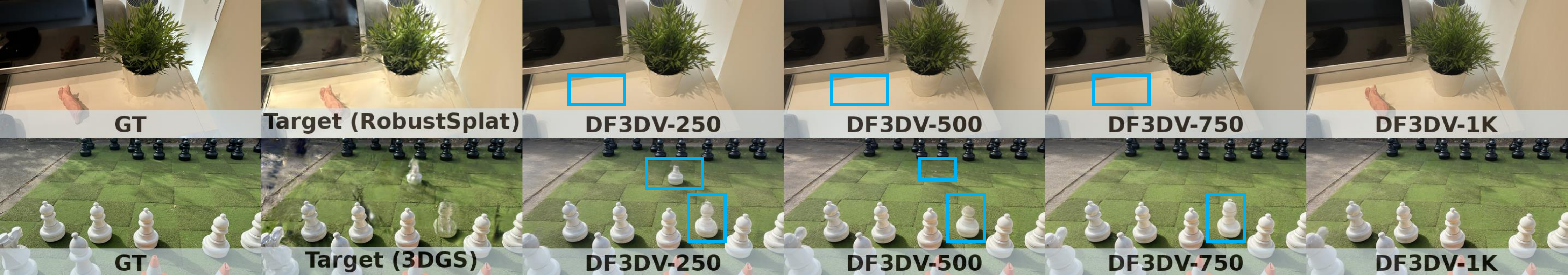}
\caption{
\textbf{Qualitative results of DI$^2$FIX trained with different data scales.} 
Increasing the amount and diversity of training data improves robustness. 
In particular, DI$^2$FIX progressively removes distractor artifacts (e.g., the chess pieces) while avoiding incorrect modifications to static scene content (e.g., the animal toy).
}
\label{fig:abl_size}
\end{figure}
\begin{figure}[t]
\centering
\includegraphics[width=0.99\linewidth]{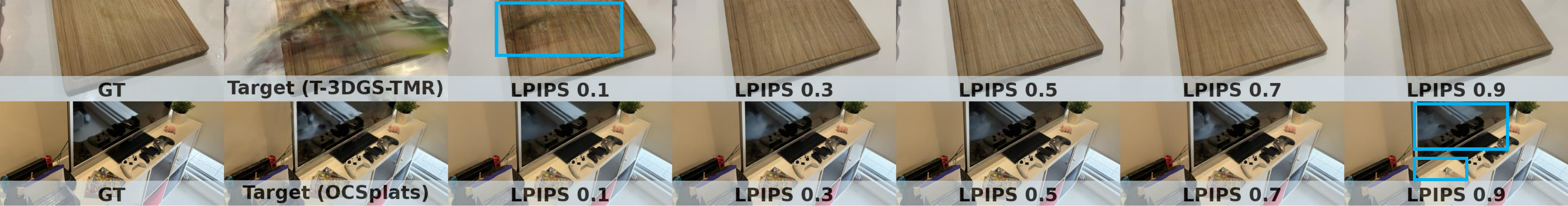}
\caption{
\textbf{Qualitative results of DI$^2$FIX trained using different LPIPS filtering thresholds.} 
Overly strict thresholds (e.g., 0.1) exclude many challenging image pairs, making artifacts difficult to remove (e.g., floaters around the cutting board). 
Overly loose thresholds (e.g., 0.9) introduce excessive noisy training samples, which may lead to undesired modifications of scene content (e.g., disappearing game cards). 
A moderate threshold provides a better balance between data quality and diversity.
%resulting in more stable and visually consistent reconstruction.
}
\label{fig:abl_lpips}
\end{figure} \\
\textbf{DI$\bm{^2}$FIX Results.}
%~\cref{tab:difix_overall_bench} reports the average performance change of DIFIX~\cite{wu2025difix3d+} and its variants fine-tuned on RobustNeRF~\cite{sabour2023robustnerf} and DF3DV-1K* (DI$^2$FIX), relative to the corresponding vanilla baseline. 
\cref{tab:difix_overall_bench} reports the performance of 2D enhancers, where DIFIX+Robust and DI$^2$FIX denote DIFIX~\cite{wu2025difix3d+} fine-tuned on RobustNeRF~\cite{sabour2023robustnerf} data and DF3DV-1K*, respectively.
DI$^2$FIX improves the average rendering quality of 3DGS~\cite{kerbl20233d} and distractor-free radiance-field methods, particularly in PSNR and LPIPS, achieving a 0.96 dB PSNR gain and a 0.057 reduction in LPIPS. 
In contrast, DIFIX~\cite{wu2025difix3d+} without domain-specific fine-tuning degrades performance on average, while DIFIX+RobustNeRF still results in notable performance drops.
%The results emphasize the role of large-scale, domain-specific data.
~\cref{tab:difix_all_metrics} shows that DI$^2$FIX consistently improves PSNR and reduces LPIPS for every method. 
The SSIM changes are relatively limited, which is unsurprising since DIFIX~\cite{wu2025difix3d+} does not use an SSIM loss, and the perceptual trade-off with respect to SSIM has been reported in prior studies~\cite{blau2018perception, blau2019rethinking, li2025one, lin2025harnessing}.
%\cref{fig:main_fix_vis,fig:fix_41_1,fig:fix_41_2,fig:fix_41_3,fig:fix_41_4,fig:fix_41_5,fig:fix_41_6} 
%show substantial perceptual quality improvements achieved by DI$^2$FIX, highlighting the importance of DF3DV-1K. 
%Specifically, we present qualitative comparisons of DIFIX, DIFIX fine-tuned on RobustNeRF~\cite{sabour2023robustnerf}, and DF3DV-1K* (DI$^2$FIX), applied to enhance the outputs of 3DGS~\cite{kerbl20233d}, SLS~\cite{sabour2025spotlesssplats}, and AsymGS~\cite{li2025asymgs}. 
Benefiting from large-scale and diverse supervision, DI$^2$FIX suppresses distractor artifacts and restores visual structures across several challenging scenarios, whereas DIFIX and DIFIX+RobustNeRF often retain distractors or mistakenly modify static components (see~\cref{fig:main_fix_vis} and~\cref{fig:fix_41_1,fig:fix_41_2,fig:fix_41_3,fig:fix_41_4,fig:fix_41_5,fig:fix_41_6}). 
%Surprisingly, even in large-scale distractor scenarios, DI$^2$FIX can plausibly recover background content occluded by severe floaters. 
%While a small number of failure cases remain, such as semi-transient distractor scenarios where observations are heavily degraded, DI$^2$FIX still produces more stable and visually faithful reconstructions overall.
\\
\textbf{Importance of Dataset Scale and Diversity.}
We randomly split DF3DV-1K into training sets of 250, 500, and 750 scenes and fine-tune DIFIX~\cite{wu2025difix3d+} to obtain DI$^2$FIX. 
Performance improves with larger training sets (see~\cref{tab:df3dv_size_abl}), and qualitative results (see~\cref{fig:abl_size}) show that additional scenes help DI$^2$FIX better identify distractors and avoid editing static regions.
\\
\textbf{Effect of Training Data Degradation Level.} To analyze the effect, we select pairs with LPIPS $\leq \gamma$, and fine-tune DIFIX~\cite{wu2025difix3d+} to obtain DI$^2$FIX.
\cref{tab:df3dv_lpips_abl,fig:abl_lpips} show that a moderate threshold promotes better performance. Stricter thresholds make the training set less diverse and challenging, weakening the model’s ability to handle distractors, while looser thresholds encourage unnecessary edits to non-distractor regions. \\
\textbf{Out-of-Distribution Test.} To evaluate the generalizability of DI$^2$FIX to unseen radiance-field methods, we conduct a leave-one-method-out training and evaluation. 
%Specifically, for each of the ten distractor-free methods, we hold that method out and construct training pairs using renderings from the remaining nine methods, then fine-tune DIFIX to obtain one DI$^2$FIX model. 
%This yields ten DI$^2$FIX models, each trained without seeing renderings from one specific method. 
%We then evaluate each model and report the performance change relative to the DI$^2$FIX model trained using data from all ten methods. 
~\cref{tab:difix_cross_method_abl} shows stable out-of-distribution performance, with worst-case changes limited to 0.1 dB PSNR, 0.005 SSIM decrease, and a 0.009 LPIPS increase, indicating that DI$^2$FIX generalizes well to unseen methods. 

\section{Conclusion}
\label{sec:conclusion}

We introduced DF3DV-1K, a large-scale real-world dataset and benchmark for distractor-free novel view synthesis. 
It contains 1,048 indoor and outdoor scenes with clean and cluttered images, spanning 161 scene themes and 128 distractor types, and includes DF3DV-41 as a systematically designed challenging subset for scenario-wise evaluation. 
We establish a comprehensive benchmark across 10 representative methods, enabling reliable large-scale comparisons and revealing remaining failure cases. 
Beyond benchmarking, we show that DF3DV-1K enables progress beyond scene-specific vision by fine-tuning DIFIX on DF3DV-1K* to obtain DI$^2$FIX, a plug-and-play 2D enhancer for distractor-free radiance fields that delivers consistent improvements. 
We hope DF3DV-1K accelerates the development of more robust and generalizable distractor-free vision methods. \\
\textbf{Limitations.} %Although DF3DV-1K is a large-scale, diverse, real-world dataset with clean and cluttered images, it 
DF3DV-1K is smaller than general datasets~\cite{ling2024dl3dv} due to higher capture costs and careful scenario design.
Slight cloud movement may still occur in clean images, but it is minimal due to the short capture duration.
DI$^2$FIX may fail when input views are severely corrupted or exhibit confirmation bias (see~\cref{fig:fix_fail}). 
A multi-reference framework is a good future direction.
%, as our focus is on DF3DV-1K.

\section{Acknowledgements}
\label{sec:ack}

This work was supported in part by the Australian Research Council (ARC) under discovery grant DP250103612 and DP260101395, ARC Research Hub for Human-Robot Teaming for Sustainable and Resilient Construction (ITRH) grant IH240100016, and Australian National Health and Medical Research Council (NHMRC) Ideas Grant APP2021183.
We thank the NYCU Computational Photography Lab for insightful discussions related to this work, and Fu-Jung Liu, Li-Chen Liu, Su-Lan Liu, and Yu-Jun Huang for their assistance with data collection.
%\input{sec/temp}

\bibliographystyle{splncs04}
\bibliography{main}

\clearpage
\appendix
\setcounter{section}{0}
\setcounter{figure}{0}
\setcounter{table}{0}
\renewcommand{\thefigure}{S.\arabic{figure}}
\renewcommand{\thetable}{S.\arabic{table}}
\renewcommand{\thesection}{S.\arabic{section}}

\section{Author Statement}

In this supplementary material, we provide detailed information on the experimental setup of the DF3DV-1K and DF3DV-41 benchmarks, a significance analysis of the top four methods, scenario-wise performance analyses including quantitative and qualitative results for each benchmarked radiance field method, a trade-off analysis of view-dependent static objects (see~\cref{sec:bench}), experimental details of DI$^2$FIX, the performance of DI$^2$FIX across different radiance field methods, out-of-distribution evaluations of DI$^2$FIX, and failure cases of DI$^2$FIX (see~\cref{sec:d2fix}). 
We also present all scenes in DF3DV-41, together with detailed descriptions of themes and distractor types (see~\cref{sec:df3dv}).
Due to the page limitations of the main paper, we aim to closely connect the main paper and the supplementary material to facilitate a smoother reading experience.
\\
\\
\textbf{Dataset Organization and File Structure.}
%We ensure the work follows the ECCV dataset and benchmark track policy.
%Upon acceptance, we will release and maintain the dataset.
%We also plan to maintain the benchmark as a leaderboard.
The dataset and leaderboard are available at \url{https://johnnylu305.github.io/df3dv1k_web/}.
All data were collected in accordance with applicable local laws and regulations.
The dataset is distributed under the Creative Commons Attribution-NonCommercial 4.0 International (CC BY-NC 4.0) license.
%The current dataset is hosted on a private Google Drive. 
The 1,007 scenes are divided into 25 folders, named from "0000" to "0024", where each folder contains up to 49 scene zip files.
Each scene zip file follows the naming convention "series\_number-scene\_name" 
(e.g., "221225-Livingroom"). 
Every scene contains three subfolders: "series\_number-scene\_name-All", "series\_number-scene\_name-Clean", and "series\_number-scene\_name-Clutter".
The "Clean" and "Clutter" folders each contain an "images" directory storing all clean and cluttered JPG images, respectively.
The"series\_number-scene\_name-All" folder contains curated data prepared for distractor-free radiance field research. 
Specifically, it includes the following components:
\begin{itemize}
    \item \textbf{images} --- stores the valid images selected from the "Clean" and "Clutter" folders. 
    Cluttered images are named "clutter\_xxx.JPG", while clean images are named "extra\_xxx.JPG".
    \item \textbf{sparse} --- contains the COLMAP~\cite{schoenberger2016mvs,schoenberger2016sfm} sparse reconstruction results, including estimated camera poses, intrinsic parameters, and reconstructed sparse 3D points.
    \item \textbf{undistortion\_images} --- stores undistorted images generated from the "images" folder after COLMAP~\cite{schoenberger2016mvs, schoenberger2016sfm} image undistortion.
    \item \textbf{undistortion\_sparse} --- contains the corresponding COLMAP~\cite{schoenberger2016mvs, schoenberger2016sfm} sparse reconstruction with undistorted camera parameters, where distortion coefficients have been removed.
    \item \textbf{instant-ngp~\cite{muller2022instant} JSON files} --- camera parameters are additionally parsed into the official instant-ngp~\cite{muller2022instant} JSON format, including "transforms.json", "transforms\_clutter.json", and "transforms\_extra.json", which contain data information for all images, cluttered images only, and clean images only, respectively.
\end{itemize}
The naming convention and directory organization of DF3DV-1K follow commonly adopted structures in distractor-free radiance field research, with minor renaming for clarity (e.g., "sparse" $\rightarrow$ "undistortion\_sparse" and "images" $\rightarrow$ "undistortion\_images"). 
DF3DV-41 follows the same organization but is hosted in a separate folder.
%Since each folder contains no more than 49 zip files, users can download the data using gdown without exceeding Google Drive download limits, or alternatively download it directly from Google Drive.
The annotation data are provided as JSON files for each scene.
%following the same directory organization without compression, allowing direct inspection.
The total dataset size is approximately 2~TB.
Overall, this design provides a user-friendly and standardized data layout that enables straightforward integration with existing radiance field training and evaluation pipelines.
%The Google Drive repository or alternative hosting platforms will be made publicly accessible upon acceptance, with the download link provided in the camera-ready paper.

\section{DF3DV-1K and DF3DV-41 Benchmarks}
\label{sec:bench}

In this section, we provide additional experimental details, significance analysis, scenario-wise qualitative and quantitative results, and a trade-off analysis of view-dependent static objects.
We evaluate nine recent open-source distractor-free radiance field methods, including AsymGS~\cite{li2025asymgs}, RobustSplat~\cite{2025RobustSplat}, OCSplats~\cite{ling2025ocsplats}, DeGauss~\cite{wang2025degauss}, SLS~\cite{sabour2025spotlesssplats}, DeSplat~\cite{wang2025desplat}, WildGaussians~\cite{kulhanek2024wildgaussians}, T-3DGS and T-3DGS-TMR~\cite{markin2024t}, together with 3DGS~\cite{kerbl20233d}, on the DF3DV-1K and DF3DV-41 benchmarks.
The goal is to identify the most robust methods under challenging scenarios and the most challenging scenarios in the benchmarks.
\\
\\
\textbf{Experimental Details.}
Our evaluation is fully based on the official implementations of distractor-free radiance field methods~\cite{li2025asymgs, wang2025degauss, wang2025desplat, ling2025ocsplats, 2025RobustSplat, sabour2025spotlesssplats, markin2024t, kulhanek2024wildgaussians}, with the following minimal modifications to ensure a fair comparison.
Some methods perform image undistortion on-the-fly using OpenCV~\cite{bradski2000opencv}, adopt scene-dependent parameter settings, or downsample images prior to undistortion.
To ensure a fair comparison, we standardize the data preprocessing pipeline across methods and the training configuration across all scenes.
Specifically, images are first undistorted offline using COLMAP~\cite{schoenberger2016mvs, schoenberger2016sfm}, followed by $8\times$ downsampling after undistortion.
The same parameter configuration (defaulting to the On-the-go setting in the official implementation, when applicable) is used for all scenes to avoid advantages from scene-specific tuning.
We further observe that some methods, such as DeGauss~\cite{wang2025degauss}, may occasionally become unstable and crash on a small subset of scenes.
In such cases, we slightly reduce the learning rate to stabilize optimization and retrain the affected scenes.
\\
\\
\textbf{Significance analysis.}
We report pairwise significance comparisons among the top four methods for PSNR, SSIM, and LPIPS in~\cref{tab:sig_test}.
For each metric block, the upper triangular entries show the paired $t$-test $p$-values between pairs of methods across the evaluated scenes, while the lower triangular entries show the corresponding average performance margins.
Smaller $p$-values indicate stronger statistical evidence of a performance difference between two methods.
Colors encode the ranking of pairwise differences within each metric.
The consistent patterns between the upper triangular $p$-values and the lower triangular performance margins suggest that the observed method rankings are statistically meaningful, highlighting the effectiveness of DF3DV-1K as a benchmark
\begin{table}[b]
\centering

\caption{\textbf{DF3DV-1K significance analysis.} 
We compare top-performing methods using paired $t$-tests across evaluated scenes for PSNR, SSIM, and LPIPS. 
In each metric block, upper triangular entries report $p$-values, while lower triangular entries report the corresponding average performance margins. 
Smaller $p$-values indicate stronger statistical significance, and colors encode the ranking.
}
\label{tab:sig_test}

% ================= PSNR =================
\begin{tabular}{l|cccc}
& \multicolumn{4}{c}{\textbf{PSNR}} \\
\diagbox{margin}{$p$-value}
& AsymGS~\cite{li2025asymgs} & RobustSplat~\cite{2025RobustSplat} & OCSplats~\cite{ling2025ocsplats} & DeGauss~\cite{wang2025degauss} \\ \hline
AsymGS~\cite{li2025asymgs}
& --
& \cellcolor{rankthree}$9.3{\times}10^{-14}$
& \cellcolor{rankone}$7.4{\times}10^{-36}$
& \cellcolor{ranktwo}$4.7{\times}10^{-19}$ \\
RobustSplat~\cite{2025RobustSplat}
& \cellcolor{rankthree}$0.3572$
& --
& \cellcolor{rankfour}$1.9{\times}10^{-2}$
& \cellcolor{rankfive}$2.2{\times}10^{-2}$ \\
OCSplats~\cite{ling2025ocsplats}
& \cellcolor{ranktwo}$0.4708$
& \cellcolor{rankfive}$0.1135$
& --
& \cellcolor{ranksix}$6.8{\times}10^{-1}$ \\
DeGauss~\cite{wang2025degauss}
& \cellcolor{rankone}$0.4908$
& \cellcolor{rankfour}$0.1336$
& \cellcolor{ranksix}$0.0200$
& -- \
\end{tabular}

% ================= SSIM =================
\begin{tabular}{l|cccc}
& \multicolumn{4}{c}{\textbf{SSIM}} \\
\diagbox{margin}{$p$-value}
& AsymGS~\cite{li2025asymgs} & RobustSplat~\cite{2025RobustSplat} & DeGauss~\cite{wang2025degauss} & OCSplats~\cite{ling2025ocsplats} \\ \hline
AsymGS~\cite{li2025asymgs}
& --
& \cellcolor{rankthree}$5.9{\times}10^{-33}$
& \cellcolor{ranktwo}$4.5{\times}10^{-57}$
& \cellcolor{rankone}$8.6{\times}10^{-139}$ \\
RobustSplat~\cite{2025RobustSplat}
& \cellcolor{rankthree}$0.0210$
& --
& \cellcolor{ranksix}$8.2{\times}10^{-1}$
& \cellcolor{rankfive}$2.4{\times}10^{-2}$ \\
DeGauss~\cite{wang2025degauss}
& \cellcolor{ranktwo}$0.0214$
& \cellcolor{ranksix}$0.0004$
& --
& \cellcolor{rankfour}$5.6{\times}10^{-3}$ \\
OCSplats~\cite{ling2025ocsplats}
& \cellcolor{rankone}$0.0249$
& \cellcolor{rankfour}$0.0039$
& \cellcolor{rankfive}$0.0035$
& -- \\
\end{tabular}

% ================= LPIPS =================
\begin{tabular}{l|cccc}
& \multicolumn{4}{c}{\textbf{LPIPS}} \\
\diagbox{margin}{$p$-value}
& RobustSplat~\cite{2025RobustSplat} & AsymGS~\cite{li2025asymgs} & DeGauss~\cite{wang2025degauss} & OCSplats~\cite{ling2025ocsplats} \\ \hline
RobustSplat~\cite{2025RobustSplat}
& --
& \cellcolor{ranksix}$5.4{\times}10^{-2}$
& \cellcolor{rankthree}$2.5{\times}10^{-14}$
& \cellcolor{ranktwo}$4.5{\times}10^{-32}$ \\
AsymGS~\cite{li2025asymgs}
& \cellcolor{ranksix}$0.0035$
& --
& \cellcolor{rankfour}$2.2{\times}10^{-9}$
& \cellcolor{rankone}$4.7{\times}10^{-37}$ \\
DeGauss~\cite{wang2025degauss}
& \cellcolor{rankthree}$0.0168$
& \cellcolor{rankfour}$0.0133$
& --
& \cellcolor{rankfive}$2.8{\times}10^{-2}$ \\
OCSplats~\cite{ling2025ocsplats}
& \cellcolor{rankone}$0.0213$
& \cellcolor{ranktwo}$0.0178$
& \cellcolor{rankfive}$0.0045$
& -- \\
\end{tabular}

\end{table}
\\
\\
\textbf{Scenario-wise Performance.}
DF3DV-41 includes 17 systematically designed challenging scenarios.
Details of these 41 scenes are provided in~\cref{fig:bench_41_all_1,fig:bench_41_all_2}.
In addition, scenario-wise improvements in PSNR, SSIM, and LPIPS of each benchmark method relative to 3D Gaussian Splatting (3DGS)~\cite{kerbl20233d} are shown in~\cref{tab:bench41_psnr,tab:bench41_lpips,tab:bench41_ssim}.
From the quantitative results per scenario, we observe that AsymGS~\cite{li2025asymgs}, RobustSplat~\cite{2025RobustSplat}, and OCSplats~\cite{ling2025ocsplats} perform robustly across all scenarios, achieving consistent improvements across the benchmark.
In contrast, other methods may fail under specific scenarios, occasionally underperforming compared to 3DGS~\cite{kerbl20233d}.

\begin{figure}[tbh!]
  \centering
  \includegraphics[width=0.96\linewidth]{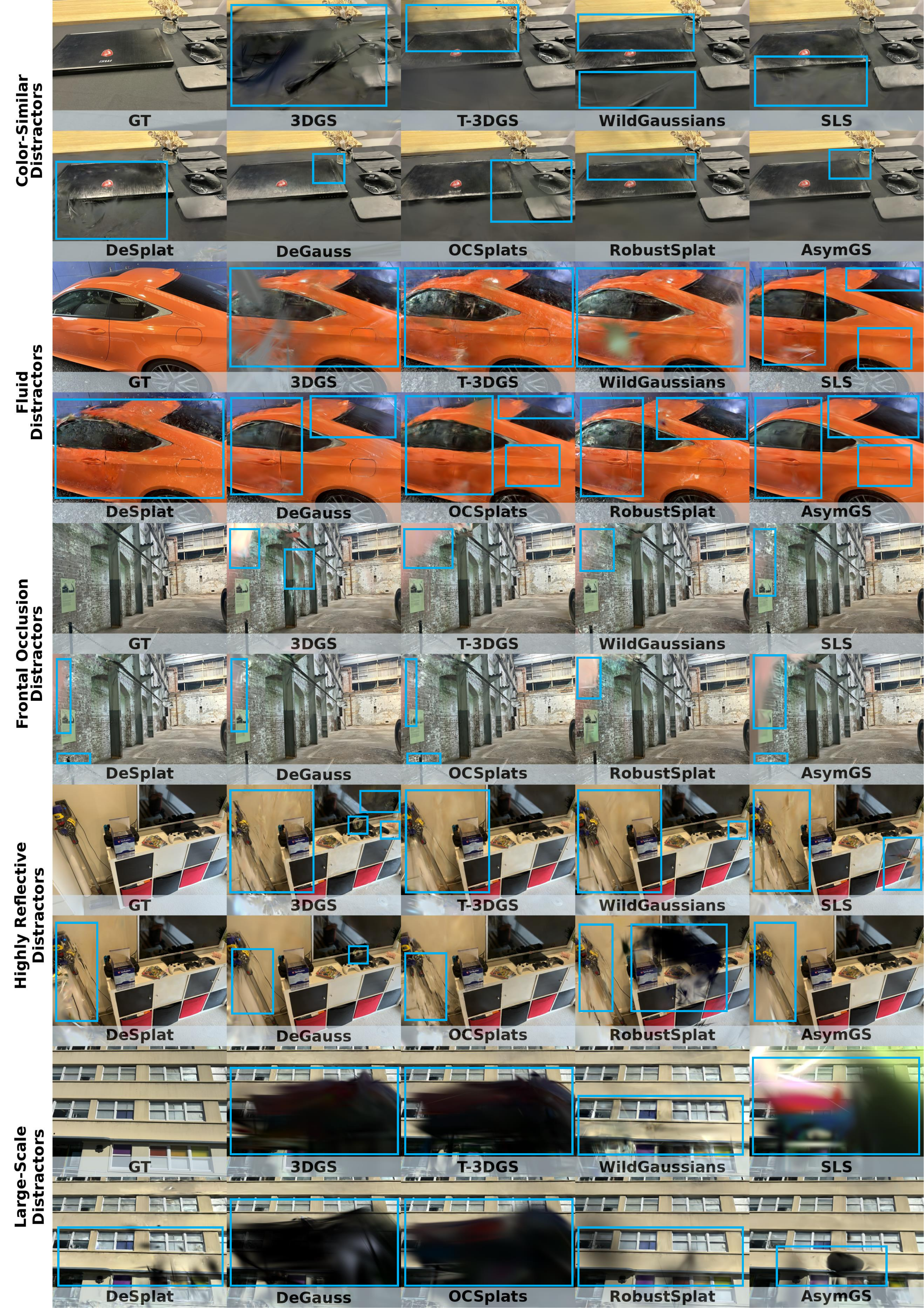}
  \caption{
\textbf{Qualitative comparison of radiance field methods on DF3DV-41 across color-similar, fluid, frontal occlusion, highly reflective, and large-scale distractors.}
The benchmark introduces systematically challenging conditions that enable clear visual comparison across methods and support the evaluation of robustness differences.
}
  \label{fig:vis_41_1}
\end{figure}

\begin{figure}[tbh!]
  \centering
  \includegraphics[width=0.96\linewidth]{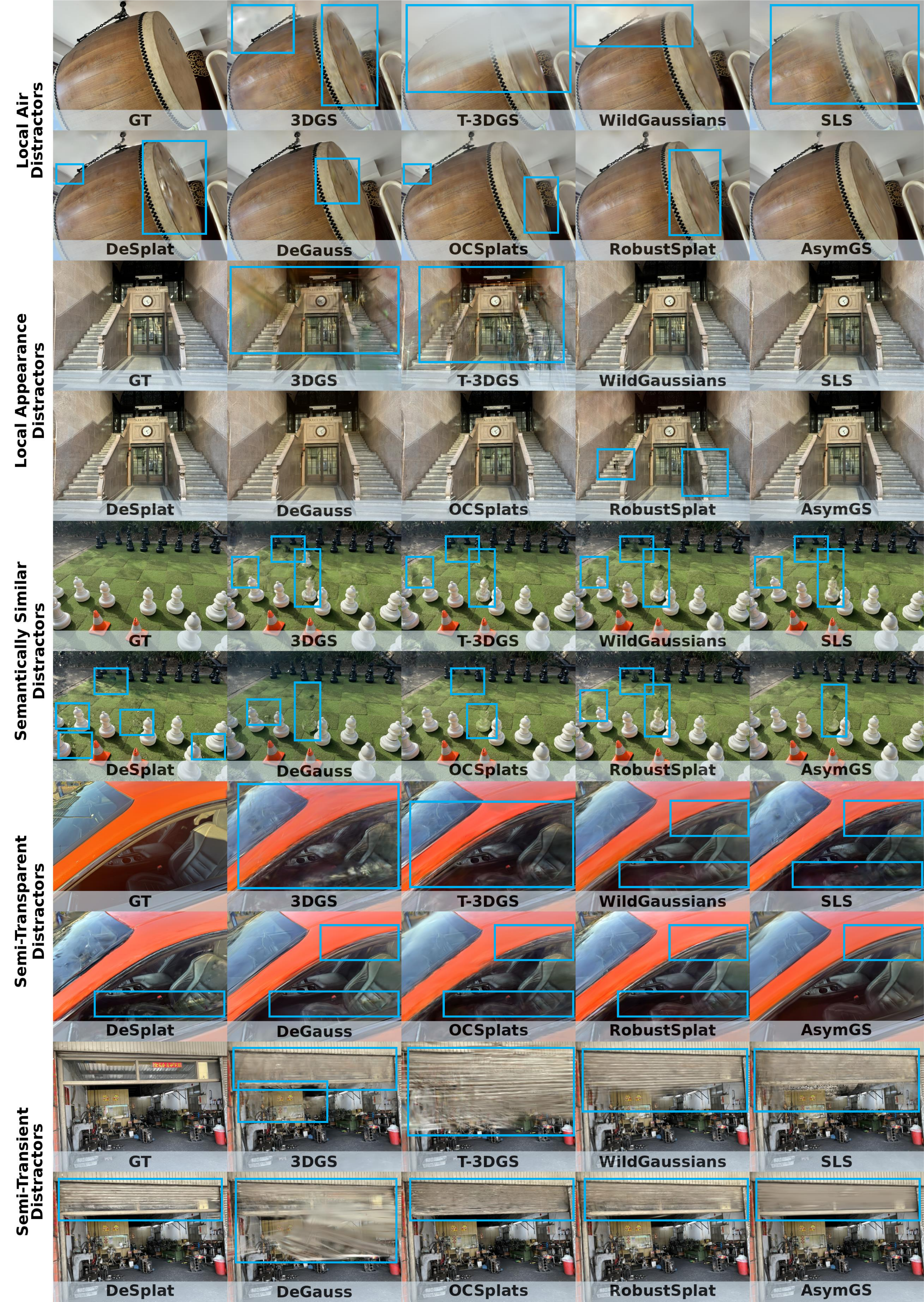}
  \caption{
  \textbf{Qualitative comparison of radiance field methods on DF3DV-41 across local air, local appearance, semantically similar, semi-transparent, and semi-transient distractors.}
  The benchmark introduces systematically challenging conditions that allow clear visual comparison across methods and support the evaluation of robustness differences.}
  \label{fig:vis_41_2}
\end{figure}

\begin{figure}[tbh!]
  \centering
  \includegraphics[width=0.96\linewidth]{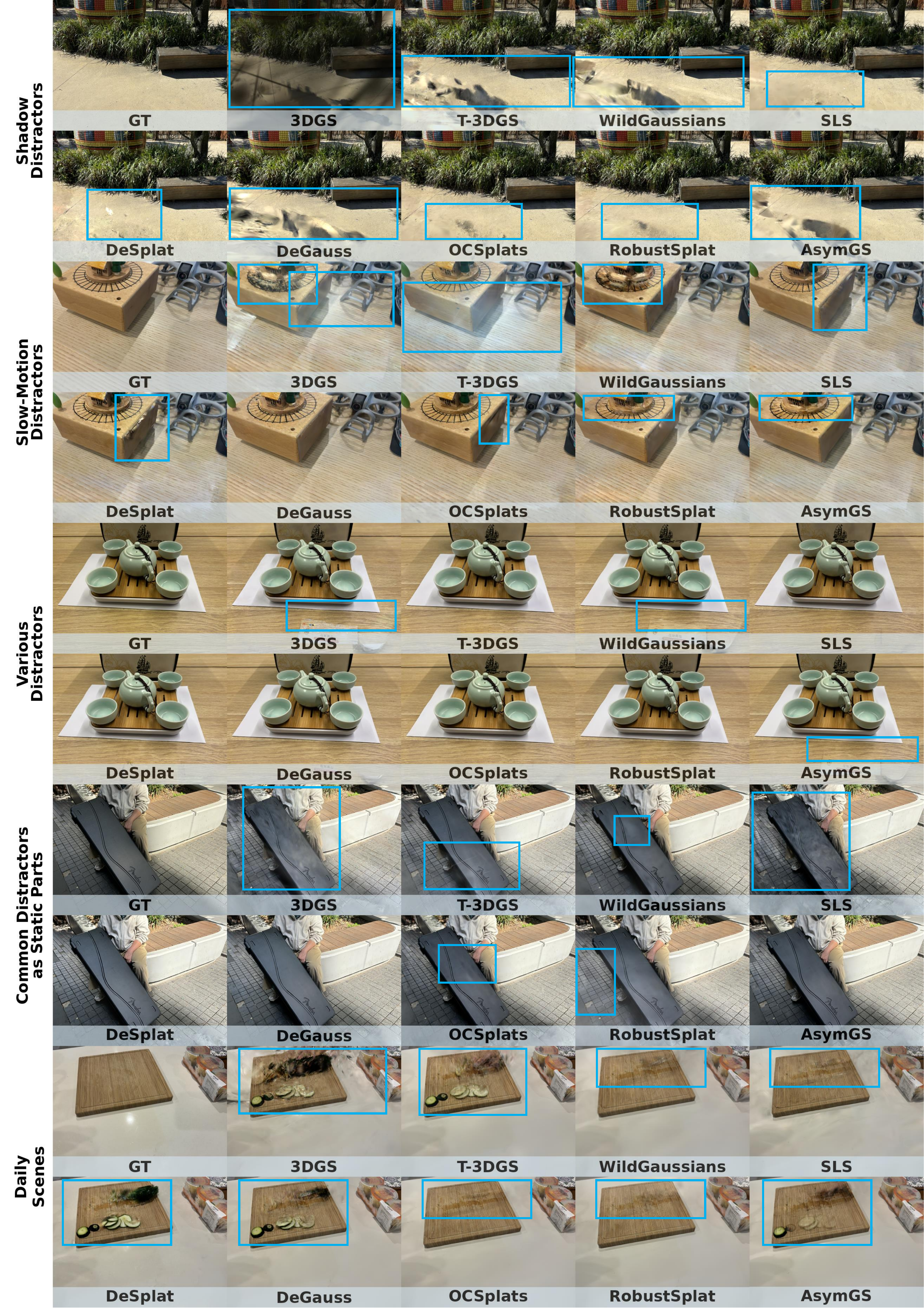}
  \caption{
   \textbf{Qualitative comparison of radiance field methods on DF3DV-41 across shadow, slow-motion, and various distractors, as well as common distractors treated as static parts and daily scenes.}
  The benchmark introduces systematically challenging conditions that allow clear visual comparison across methods and support the evaluation of robustness differences.}
  \label{fig:vis_41_3}
\end{figure}

\begin{figure}[tbh!]
  \centering
  \includegraphics[width=0.96\linewidth]{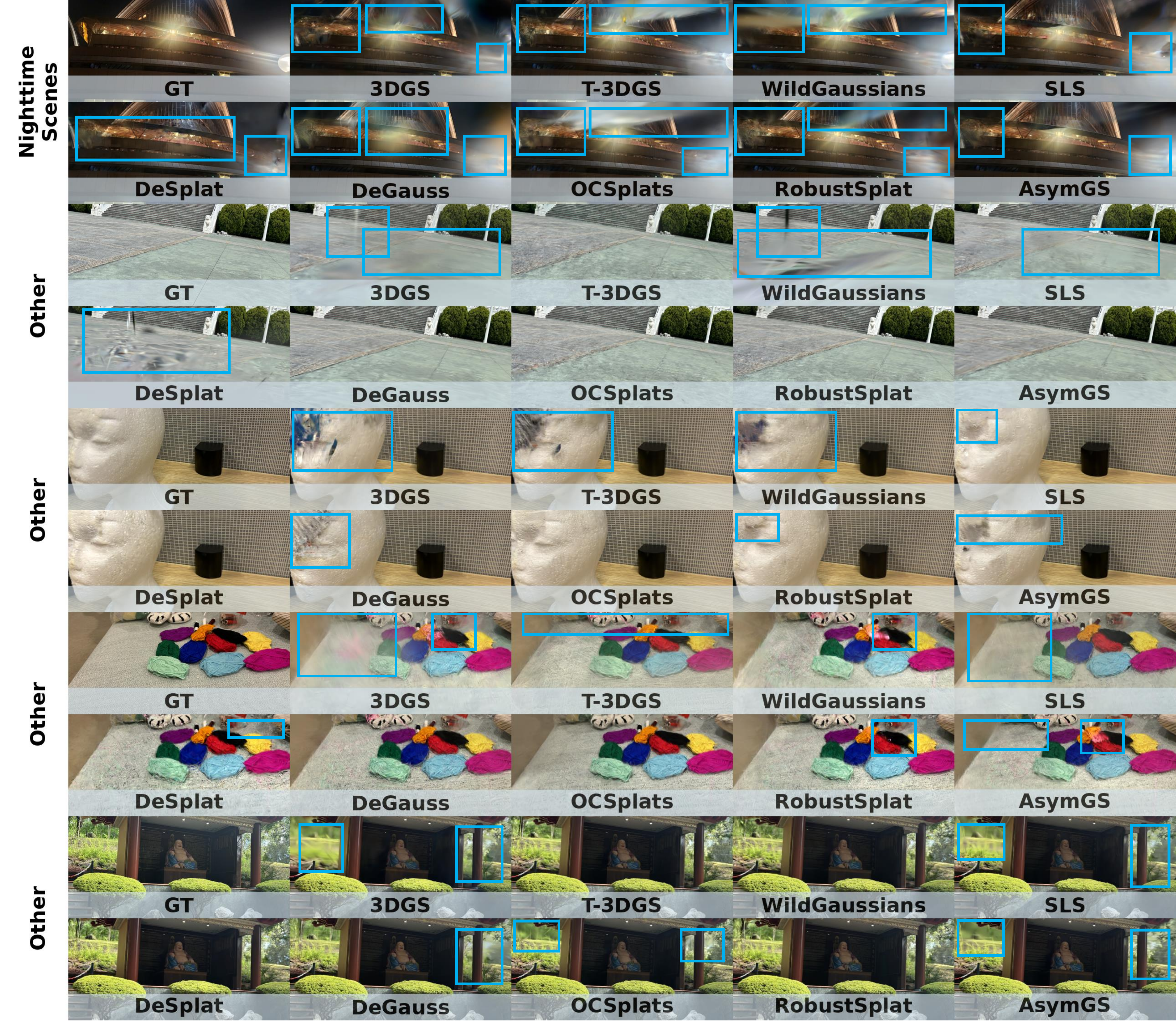}
  \caption{
 \textbf{Qualitative comparison of radiance field methods on DF3DV-41 across nighttime scenes and other distractors and scenes.}
  The benchmark introduces systematically challenging conditions that allow clear visual comparison across methods and support the evaluation of robustness differences.}
  \label{fig:vis_41_4}
\end{figure}

\begin{figure}[tbh!]
\centering
\includegraphics[width=0.95\linewidth]{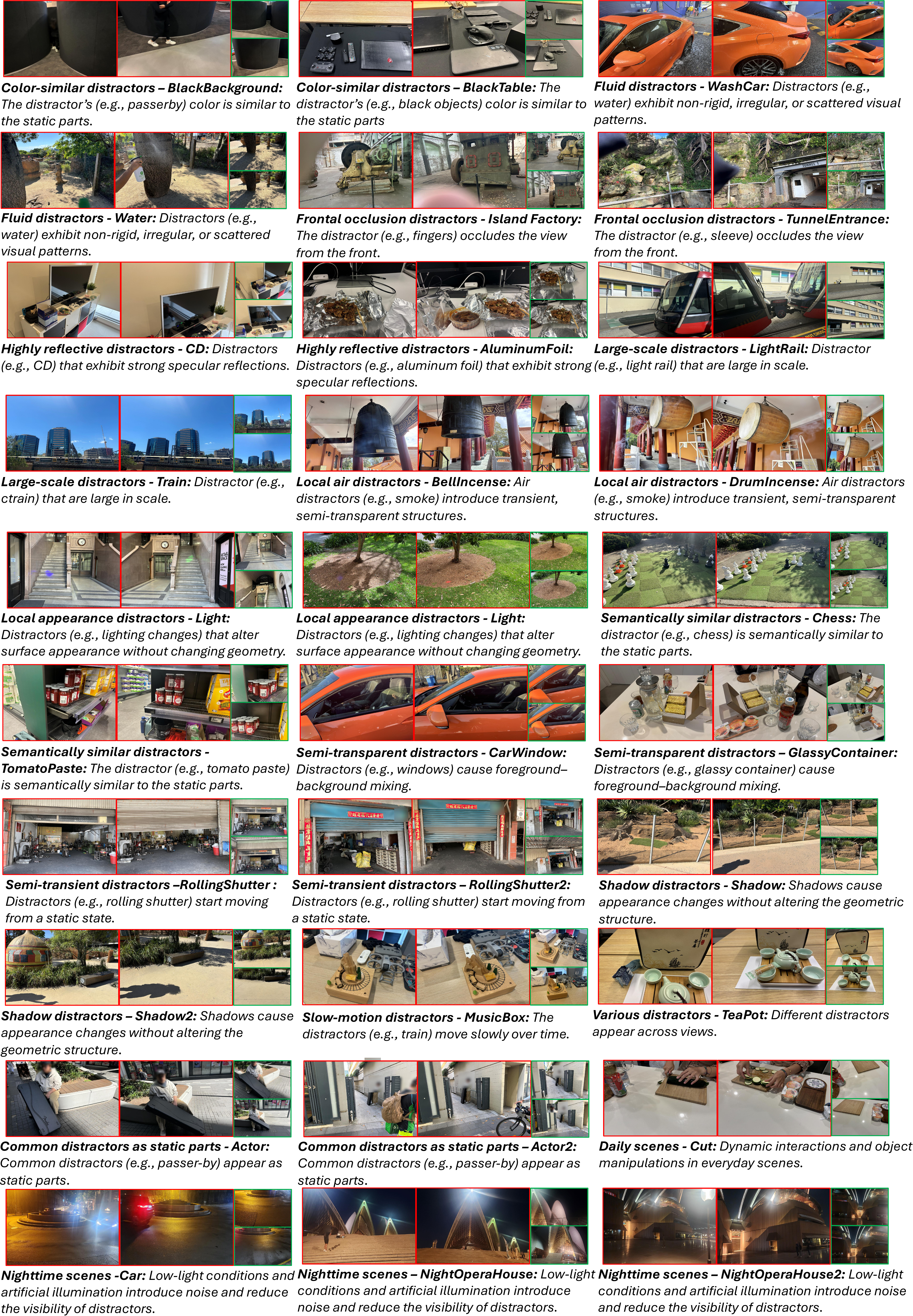}
\caption{
\textbf{Sample views of scenes in DF3DV-41.}
DF3DV-41 covers a wide range of specifically designed distractor types and scene scenarios.
The systematic design enables evaluation and clear comparison of method robustness under diverse challenging conditions.
}
\label{fig:bench_41_all_1}
\end{figure}

\begin{figure}[tbh!]
\centering
\includegraphics[width=0.99\linewidth]{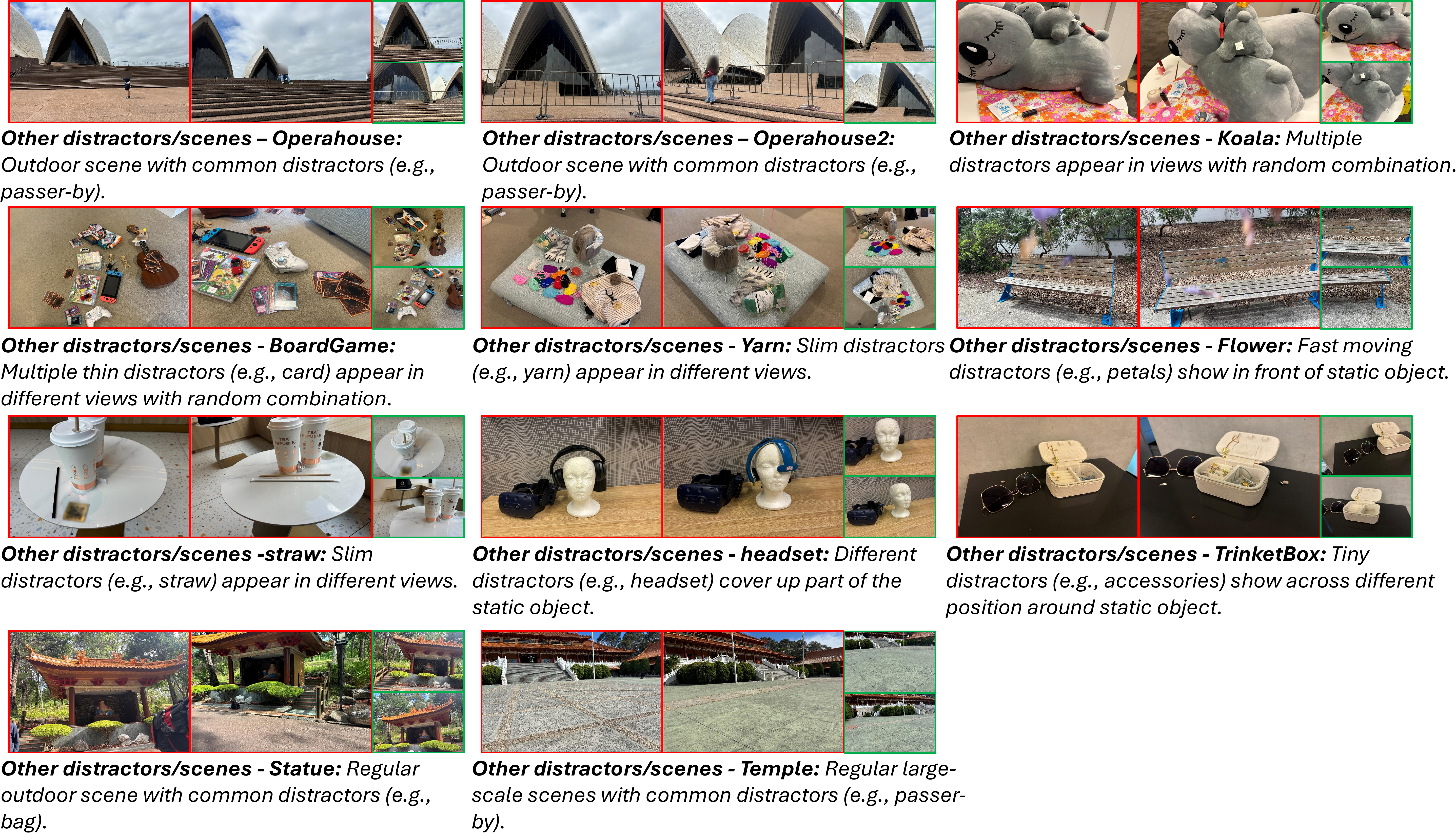}
\caption{
\textbf{Sample views of scenes in DF3DV-41.}
Other distractor and scene scenarios include diverse common distractors that vary in shape and characteristics, such as slim, small, or fast-moving objects.
}
\label{fig:bench_41_all_2}
\end{figure}

\cref{fig:vis_41_1,fig:vis_41_2,fig:vis_41_3,fig:vis_41_4} show that the challenging scenarios in DF3DV-41 enable clearly visible distinctions between benchmark methods.
We hope that this challenging dataset and benchmark facilitate \textit{“zoom-in no more”} qualitative comparisons, allowing recent state-of-the-art methods to be compared directly without relying on subtle background zoom-ins to highlight differences.

We identify semantically similar distractors, fluid distractors, and nighttime scenes as the most challenging scenarios in DF3DV-41, as reflected by their low 3DGS performance and the relatively limited improvements achieved by distractor-free radiance field methods compared to other scenarios.
These challenges are largely related to the reliance of many existing approaches on semantic features~\cite{oquab2024dinov2, ravisam, kirillov2023segment, tang2023emergent} for distractor identification.
When distractors share similar semantic meanings with static scene objects, distinguishing true distractors from valid scene content becomes inherently ambiguous. 
For instance, in the chess scene shown in~\cref{fig:vis_41_2}, chess pieces that should ideally be removed as distractors remain partially preserved in the reconstructed results across multiple methods, indicating that semantic similarity can mislead distractor detection.
For fluid distractors, many approaches learn masks online to suppress dynamic content during training.
However, fluid phenomena (e.g., splashes or water sprays) are typically spatially dispersed and semi-transparent, which violates the implicit assumption of coherent object boundaries.
As a result, masking strategies may either remove valid scene regions or fail to completely suppress the distractors, producing noticeable blending artifacts.
This behavior can be clearly observed in the washing-car scene in~\cref{fig:vis_41_1}, where artifacts introduced by fluid distractors spread across large portions of the reconstructed scene.
For nighttime scenes, many approaches rely on fixed thresholds or appearance-based heuristics to filter distractors.
Because these thresholds are commonly tuned under daytime illumination, their reliability degrades under low-light conditions, where distractors and static scene components exhibit reduced contrast and weaker semantic cues, making reliable separation significantly more difficult (see~\cref{fig:vis_41_4}).
\begin{table}[tb]
\caption{
\textbf{Per-scenario PSNR of each method.} 
The first column shows the PSNR achieved by 3DGS~\cite{kerbl20233d}, while all other entries indicate the performance difference relative to 3DGS~\cite{kerbl20233d}. 
AsymGS~\cite{li2025asymgs}, RobustSplat~\cite{2025RobustSplat}, and OCSplats~\cite{ling2025ocsplats} show consistent robustness across scenarios, while other methods remain sensitive to specific conditions, which can result in performance degradation.
In addition, we identify semantically similar distractors as the most challenging distractor scenario, as the PSNR achieved by 3DGS~\cite{kerbl20233d} is low and all methods obtain improvements of less than or equal to 1 dB compared to other distractor scenarios.
}
  \label{tab:bench41_psnr}
  \centering
  \begin{adjustbox}{width=\linewidth}
  \begin{tabular}{l|c|ccccccccc}
    \toprule
     \textbf{Scenario} &
    \makecell{\textbf{3DGS}\\\cite{kerbl20233d}} &
    \makecell{\textbf{T-3DGS}\\\cite{markin2024t}} &
    \makecell{\textbf{T-3DGS-TMR}\\\cite{markin2024t}} &
    \makecell{\textbf{WildGaussian}\\\cite{kulhanek2024wildgaussians}} &
    \makecell{\textbf{SLS}\\\cite{sabour2025spotlesssplats}} &
    \makecell{\textbf{DeSplat}\\\cite{wang2025desplat}} &
    \makecell{\textbf{DeGauss}\\\cite{wang2025degauss}} &
    \makecell{\textbf{OCSplats}\\\cite{ling2025ocsplats}} &
    \makecell{\textbf{RobustSplat}\\\cite{2025RobustSplat}} &
    \makecell{\textbf{AsymGS}\\\cite{li2025asymgs}} \\
    \midrule

\makecell[l]{Color-similar\\distractors} & $17.47$ & \good{$+2.02$} & \good{$+0.81$} & \good{$+1.37$} & \good{$+1.60$} & \good{$+2.15$} & \good{$\bm{+3.53}$} & \good{$+2.26$} & \good{$+3.18$} & \good{$+3.01$} \\
\midrule
\makecell[l]{Fluid\\distractors} & $14.54$ & \good{$+0.02$} & \bad{$-0.01$} & \good{$+0.52$} & \good{$+0.96$} & \good{$+0.82$} & \good{$+1.40$} & \good{$+1.30$} & \good{$+0.46$} & \good{$\bm{+1.71}$} \\
\midrule
\makecell[l]{Frontal occlusion\\distractors} & $18.95$ & \good{$+1.71$} & \good{$+0.15$} & \good{$+1.73$} & \good{$+2.07$} & \good{$+1.67$} & \good{$+2.09$} & \good{$+2.08$} & \good{$\bm{+2.28}$} & \good{$+2.08$} \\
\midrule
\makecell[l]{Highly reflective\\distractors} & $17.42$ & \good{$+1.27$} & \good{$+0.12$} & \good{$+1.59$} & \good{$+1.19$} & \good{$+1.41$} & \good{$\bm{+2.39}$} & \good{$+1.85$} & \good{$+1.57$} & \good{$+2.19$} \\
\midrule
\makecell[l]{Large-scale\\distractors} & $18.00$ & \good{$+0.97$} & \good{$+0.16$} & \good{$+1.68$} & \good{$+1.44$} & \good{$+2.72$} & \good{$+2.03$} & \good{$+2.81$} & \good{$\bm{+3.57}$} & \good{$+2.49$} \\
\midrule
\makecell[l]{Local air\\distractors} & $18.70$ & \bad{$-0.38$} & \bad{$-0.33$} & \good{$+0.33$} & \good{$+0.97$} & \good{$+1.24$} & \good{$+2.24$} & \good{$+0.83$} & \good{$+1.57$} & \good{$\bm{+2.34}$} \\
\midrule
\makecell[l]{Local appearance\\distractors} & $15.60$ & \good{$+0.05$} & \bad{$-0.01$} & \good{$+1.70$} & \good{$+2.19$} & \good{$+2.54$} & \good{$+2.95$} & \good{$+2.43$} & \good{$+1.17$} & \good{$\bm{+3.29}$} \\
\midrule
\makecell[l]{Semantically similar\\distractors} & $16.24$ & \good{$+0.68$} & \good{$+0.32$} & \good{$+0.66$} & \good{$+0.24$} & \good{$+0.02$} & \good{$+0.80$} & \good{$+0.41$} & \good{$+0.63$} & \good{$\bm{+1.00}$} \\
\midrule
\makecell[l]{Semi-transparent\\distractors} & $19.10$ & \good{$+1.03$} & \good{$+0.29$} & \good{$+0.99$} & \good{$+0.78$} & \good{$+0.11$} & \good{$+0.90$} & \good{$+0.57$} & \good{$+1.10$} & \good{$\bm{+1.36}$} \\
\midrule
\makecell[l]{Semi-transient\\distractors} & $15.85$ & \good{$+0.00$} & \bad{$-0.58$} & \good{$+1.31$} & \bad{$-0.25$} & \good{$+0.86$} & \bad{$-0.59$} & \good{$\bm{+2.50}$} & \good{$+2.18$} & \good{$+1.72$} \\
\midrule
\makecell[l]{Shadow\\distractors} & $14.38$ & \good{$+0.55$} & \good{$+0.27$} & \good{$+0.91$} & \good{\bm{$+2.65$}} & \good{$+2.13$} & \good{$+1.70$} & \good{$+2.52$} & \good{$+2.23$} & \good{$+2.61$} \\
\midrule
\makecell[l]{Slow-motion\\distractors} & $19.98$ & \bad{$-0.09$} & \bad{$-0.58$} & \bad{$-0.14$} & \good{$+1.37$} & \good{$+1.32$} & \good{$+1.95$} & \good{$\bm{+2.55}$} & \good{$+1.07$} & \good{$+2.09$} \\
\midrule
\makecell[l]{Various\\distractors} & $23.06$ & \good{$+2.31$} & \good{$+0.53$} & \good{$+1.06$} & \good{$+2.86$} & \good{$+2.64$} & \good{$\bm{+3.11}$} & \good{$+3.08$} & \good{$\bm{+3.11}$} & \good{$+2.78$} \\
\midrule
\makecell[l]{Common distractors\\as static parts} & $17.79$ & \good{$+1.60$} & \good{$+0.66$} & \good{$+2.13$} & \good{$+2.06$} & \good{$+2.13$} & \good{$+2.00$} & \good{$+2.37$} & \good{$+2.13$} & \good{$\bm{+2.95}$} \\
\midrule
\makecell[l]{Daily\\scenes} & $17.73$ & \good{$+2.39$} & \bad{$-0.77$} & \good{$+2.98$} & \good{$+2.68$} & \good{$+2.14$} & \good{$+3.44$} & \good{$+3.04$} & \good{$\bm{+3.88}$} & \good{$+3.67$} \\
\midrule
\makecell[l]{Nighttime\\scenes} & $20.25$ & \good{$+0.60$} & \good{$+0.09$} & \good{$+0.04$} & \good{$+0.47$} & \bad{$-0.63$} & \good{$+0.98$} & \good{$+0.45$} & \good{$\bm{+1.46}$} & \good{$+1.37$} \\
\midrule
\makecell[l]{Other\\distractors/scenes} & $19.32$ & \good{$+1.01$} & \good{$+0.87$} & \good{$+1.23$} & \good{$+1.08$} & \good{$+0.87$} & \good{$+2.04$} & \good{$+1.57$} & \good{$+1.72$} & \good{$\bm{+2.34}$} \\
\bottomrule
\end{tabular}
\end{adjustbox}
\end{table}
\begin{table}[tb]
  \caption{
\textbf{Per-scenario SSIM of each method.} 
The first column shows the SSIM achieved by 3DGS~\cite{kerbl20233d}, while all other entries indicate the performance difference relative to 3DGS~\cite{kerbl20233d}. 
All methods except DeSplat~\cite{wang2025desplat} demonstrate consistent robustness across scenarios.
In addition, we identify semantically similar and fluid distractors as the most challenging distractor scenarios, since 3DGS~\cite{kerbl20233d} achieves relatively low SSIM values and all methods show improvements of less than 0.08 compared to other distractor scenarios.}
  \label{tab:bench41_ssim}
  \centering
  \begin{adjustbox}{width=\linewidth}
  \begin{tabular}{l|c|ccccccccc}
    \toprule
     \textbf{Scenario} &
    \makecell{\textbf{3DGS}\\\cite{kerbl20233d}} &
    \makecell{\textbf{T-3DGS}\\\cite{markin2024t}} &
    \makecell{\textbf{T-3DGS-TMR}\\\cite{markin2024t}} &
    \makecell{\textbf{WildGaussian}\\\cite{kulhanek2024wildgaussians}} &
    \makecell{\textbf{SLS}\\\cite{sabour2025spotlesssplats}} &
    \makecell{\textbf{DeSplat}\\\cite{wang2025desplat}} &
    \makecell{\textbf{DeGauss}\\\cite{wang2025degauss}} &
    \makecell{\textbf{OCSplats}\\\cite{ling2025ocsplats}} &
    \makecell{\textbf{RobustSplat}\\\cite{2025RobustSplat}} &
    \makecell{\textbf{AsymGS}\\\cite{li2025asymgs}} \\
    \midrule

\makecell[l]{Color-similar\\distractors} & $0.657$ & \good{$+0.089$} & \good{$+0.042$} & \good{$+0.067$} & \good{$+0.078$} & \good{$+0.067$} & \good{$\bm{+0.150}$} & \good{$+0.102$} & \good{$+0.139$} & \good{$+0.128$} \\
\midrule
\makecell[l]{Fluid\\distractors} & $0.410$ & \good{$+0.004$} & \good{$+0.003$} & \good{$+0.029$} & \good{$+0.039$} & \good{$+0.006$} & \good{$+0.052$} & \good{$+0.054$} & \good{$+0.021$} & \good{$\bm{+0.079}$} \\
\midrule
\makecell[l]{Frontal occlusion\\distractors} & $0.678$ & \good{$+0.073$} & \good{$+0.045$} & \good{$+0.066$} & \good{$+0.049$} & \good{$+0.063$} & \good{$+0.069$} & \good{$+0.064$} & \good{$+0.078$} & \good{$\bm{+0.080}$} \\
\midrule
\makecell[l]{Highly reflective\\distractors} & $0.589$ & \good{$+0.101$} & \good{$+0.044$} & \good{$+0.108$} & \good{$+0.076$} & \good{$+0.110$} & \good{$\bm{+0.148}$} & \good{$+0.128$} & \good{$+0.121$} & \good{$+0.146$} \\
\midrule
\makecell[l]{Large-scale\\distractors} & $0.703$ & \good{$+0.037$} & \good{$+0.024$} & \good{$+0.046$} & \good{$+0.042$} & \good{$+0.058$} & \good{$+0.060$} & \good{$+0.073$} & \good{$\bm{+0.104}$} & \good{$+0.069$} \\
\midrule
\makecell[l]{Local air\\distractors} & $0.720$ & \good{$+0.017$} & \good{$+0.015$} & \good{$+0.032$} & \good{$+0.041$} & \good{$+0.067$} & \good{$+0.092$} & \good{$+0.053$} & \good{$+0.073$} & \good{$\bm{+0.097}$} \\
\midrule
\makecell[l]{Local appearance\\distractors} & $0.518$ & \good{$+0.012$} & \good{$+0.011$} & \good{$+0.040$} & \good{$+0.042$} & \good{$+0.045$} & \good{$+0.076$} & \good{$+0.055$} & \good{$+0.043$} & \good{$\bm{+0.092}$} \\
\midrule
\makecell[l]{Semantically similar\\distractors} & $0.457$ & \good{$+0.058$} & \good{$+0.043$} & \good{$+0.052$} & \good{$+0.024$} & \bad{$-0.009$} & \good{$+0.053$} & \good{$+0.050$} & \good{$+0.061$} & \good{$\bm{+0.079}$} \\
\midrule
\makecell[l]{Semi-transparent\\distractors} & $0.689$ & \good{$+0.067$} & \good{$+0.043$} & \good{$+0.064$} & \good{$+0.037$} & \good{$+0.027$} & \good{$+0.065$} & \good{$+0.057$} & \good{$+0.070$} & \good{$\bm{+0.089}$} \\
\midrule
\makecell[l]{Semi-transient\\distractors} & $0.515$ & \good{$+0.033$} & \good{$+0.013$} & \good{$+0.080$} & \good{$+0.019$} & \good{$+0.077$} & \good{$+0.012$} & \good{$+0.112$} & \good{$\bm{+0.114}$} & \good{$\bm{+0.114}$} \\
\midrule
\makecell[l]{Shadow\\distractors} & $0.457$ & \good{$+0.046$} & \good{$+0.038$} & \good{$+0.035$} & \good{$+0.073$} & \good{$+0.052$} & \good{$+0.069$} & \good{$+0.078$} & \good{$\bm{+0.094}$} & \good{$\bm{+0.094}$} \\
\midrule
\makecell[l]{Slow-motion\\distractors} & $0.778$ & \good{$+0.038$} & \good{$+0.014$} & \good{$+0.024$} & \good{$+0.035$} & \good{$+0.053$} & \good{$+0.066$} & \good{$\bm{+0.074}$} & \good{$+0.034$} & \good{$+0.070$} \\
\midrule
\makecell[l]{Various\\distractors} & $0.811$ & \good{$+0.067$} & \good{$+0.043$} & \good{$+0.043$} & \good{$+0.058$} & \good{$+0.061$} & \good{$\bm{+0.074}$} & \good{$+0.070$} & \good{$+0.071$} & \good{$+0.073$} \\
\midrule
\makecell[l]{Common distractors\\as static parts} & $0.621$ & \good{$+0.076$} & \good{$+0.045$} & \good{$+0.079$} & \good{$+0.053$} & \good{$+0.083$} & \good{$+0.087$} & \good{$+0.092$} & \good{$+0.093$} & \good{$\bm{+0.122}$} \\
\midrule
\makecell[l]{Daily\\scenes} & $0.701$ & \good{$+0.097$} & \good{$+0.015$} & \good{$+0.105$} & \good{$+0.090$} & \good{$+0.074$} & \good{$+0.118$} & \good{$+0.112$} & \good{$+0.131$} & \good{$\bm{+0.137}$} \\
\midrule
\makecell[l]{Nighttime\\scenes} & $0.664$ & \good{$+0.030$} & \good{$+0.019$} & \good{$+0.007$} & \good{$+0.011$} & \bad{$-0.016$} & \good{$+0.049$} & \good{$+0.027$} & \good{$+0.047$} & \good{$\bm{+0.055}$} \\
\midrule
\makecell[l]{Other\\distractors/scenes} & $0.645$ & \good{$+0.055$} & \good{$+0.049$} & \good{$+0.042$} & \good{$+0.034$} & \good{$+0.026$} & \good{$+0.074$} & \good{$+0.057$} & \good{$+0.063$} & \good{$\bm{+0.085}$} \\
\bottomrule
\end{tabular}
\end{adjustbox}
\end{table}
\begin{table}[tb]
  \caption{
\textbf{Per-scenario LPIPS of each method.}
The first column shows the LPIPS achieved by 3DGS~\cite{kerbl20233d}, while all other entries indicate the performance difference relative to 3DGS~\cite{kerbl20233d}. 
Lower LPIPS values indicate better perceptual similarity. 
All methods except T-3DGS-TMR~\cite{markin2024t} and WildGaussian~\cite{kulhanek2024wildgaussians} demonstrate consistent robustness across scenarios.
In addition, we identify fluid distractors as the most challenging scenario, since 3DGS~\cite{kerbl20233d} achieves relatively high LPIPS values and all methods show improvements of less than 0.08 compared to other distractor scenarios.}
  \label{tab:bench41_lpips}
  \centering
  \begin{adjustbox}{width=\linewidth}
  \begin{tabular}{l|c|ccccccccc}
    \toprule
     \textbf{Scenario} &
    \makecell{\textbf{3DGS}\\\cite{kerbl20233d}} &
    \makecell{\textbf{T-3DGS}\\\cite{markin2024t}} &
    \makecell{\textbf{T-3DGS-TMR}\\\cite{markin2024t}} &
    \makecell{\textbf{WildGaussian}\\\cite{kulhanek2024wildgaussians}} &
    \makecell{\textbf{SLS}\\\cite{sabour2025spotlesssplats}} &
    \makecell{\textbf{DeSplat}\\\cite{wang2025desplat}} &
    \makecell{\textbf{DeGauss}\\\cite{wang2025degauss}} &
    \makecell{\textbf{OCSplats}\\\cite{ling2025ocsplats}} &
    \makecell{\textbf{RobustSplat}\\\cite{2025RobustSplat}} &
    \makecell{\textbf{AsymGS}\\\cite{li2025asymgs}} \\
    \midrule

\makecell[l]{Color-similar\\distractors}
& $0.477$
& \good{$-0.109$}
& \good{$-0.051$}
& \good{$-0.076$}
& \good{$-0.068$}
& \good{$-0.073$}
& \good{$\bm{-0.192}$}
& \good{$-0.114$}
& \good{$-0.176$}
& \good{$-0.176$} \\
\midrule

\makecell[l]{Fluid\\distractors}
& $0.444$
& \good{$-0.042$}
& \good{$-0.041$}
& \good{$-0.031$}
& \good{$-0.017$}
& \good{$-0.052$}
& \good{$\bm{-0.076}$}
& \good{$-0.031$}
& \good{$-0.056$}
& \good{$-0.038$} \\
\midrule

\makecell[l]{Frontal occlusion\\distractors}
& $0.210$
& \good{$-0.071$}
& \good{$-0.036$}
& \good{$-0.058$}
& \good{$-0.053$}
& \good{$\bm{-0.082}$}
& \good{$-0.062$}
& \good{$-0.070$}
& \good{$-0.081$}
& \good{$-0.043$} \\
\midrule

\makecell[l]{Highly reflective\\distractors}
& $0.333$
& \good{$-0.097$}
& \good{$-0.036$}
& \good{$-0.094$}
& \good{$-0.068$}
& \good{$-0.114$}
& \good{$\bm{-0.141}$}
& \good{$-0.119$}
& \good{$-0.118$}
& \good{$-0.130$} \\
\midrule

\makecell[l]{Large-scale\\distractors}
& $0.283$
& \good{$-0.059$}
& \good{$-0.040$}
& \good{$-0.016$}
& \good{$-0.048$}
& \good{$-0.091$}
& \good{$-0.090$}
& \good{$-0.097$}
& \good{$\bm{-0.133}$}
& \good{$-0.063$} \\
\midrule

\makecell[l]{Local air\\distractors}
& $0.244$
& \good{$-0.020$}
& \good{$-0.021$}
& \good{$-0.038$}
& \good{$-0.035$}
& \good{$-0.078$}
& \good{$\bm{-0.106}$}
& \good{$-0.039$}
& \good{$-0.085$}
& \good{$-0.098$} \\
\midrule

\makecell[l]{Local appearance\\distractors}
& $0.358$
& \good{$-0.011$}
& \good{$-0.008$}
& \good{$-0.056$}
& \good{$-0.036$}
& \good{$\bm{-0.078}$}
& \good{$\bm{-0.078}$}
& \good{$-0.054$}
& \good{$-0.061$}
& \good{$-0.072$} \\
\midrule

\makecell[l]{Semantically similar\\distractors}
& $0.407$
& \good{$-0.093$}
& \good{$-0.065$}
& \good{$-0.030$}
& \good{$-0.038$}
& \good{$-0.061$}
& \good{$\bm{-0.099}$}
& \good{$-0.068$}
& \good{$-0.087$}
& \good{$-0.078$} \\
\midrule

\makecell[l]{Semi-transparent\\distractors}
& $0.289$
& \good{$-0.085$}
& \good{$-0.053$}
& \good{$-0.058$}
& \good{$-0.048$}
& \good{$-0.067$}
& \good{$-0.091$}
& \good{$-0.070$}
& \good{$\bm{-0.093}$}
& \good{$-0.091$} \\
\midrule

\makecell[l]{Semi-transient\\distractors}
& $0.341$
& \good{$-0.035$}
& \good{$-0.014$}
& \good{$-0.071$}
& \good{$-0.017$}
& \good{$-0.095$}
& \good{$-0.029$}
& \good{$-0.124$}
& \good{$\bm{-0.132}$}
& \good{$-0.088$} \\
\midrule

\makecell[l]{Shadow\\distractors}
& $0.398$
& \good{$-0.071$}
& \good{$-0.063$}
& \good{$-0.042$}
& \good{$-0.054$}
& \good{$\bm{-0.111}$}
& \good{$-0.068$}
& \good{$-0.084$}
& \good{$-0.097$}
& \good{$-0.032$} \\
\midrule

\makecell[l]{Slow-motion\\distractors}
& $0.177$
& \good{$-0.024$}
& \bad{$+0.005$}
& \good{$-0.021$}
& \good{$-0.024$}
& \good{$-0.058$}
& \good{$-0.063$}
& \good{$\bm{-0.075}$}
& \good{$-0.024$}
& \good{$-0.057$} \\
\midrule

\makecell[l]{Various\\distractors}
& $0.131$
& \good{$-0.060$}
& \good{$-0.031$}
& \good{$-0.038$}
& \good{$-0.057$}
& \good{$-0.058$}
& \good{$\bm{-0.068}$}
& \good{$-0.066$}
& \good{$\bm{-0.068}$}
& \good{$-0.062$} \\
\midrule

\makecell[l]{Common distractors\\as static parts}
& $0.268$
& \good{$-0.093$}
& \good{$-0.058$}
& \good{$-0.072$}
& \good{$-0.063$}
& \good{$-0.114$}
& \good{$-0.112$}
& \good{$-0.097$}
& \good{$-0.113$}
& \good{$\bm{-0.120}$} \\
\midrule

\makecell[l]{Daily\\scenes}
& $0.409$
& \good{$-0.127$}
& \good{$-0.018$}
& \good{$-0.123$}
& \good{$-0.106$}
& \good{$-0.115$}
& \good{$-0.166$}
& \good{$-0.138$}
& \good{$\bm{-0.177}$}
& \good{$-0.166$} \\
\midrule

\makecell[l]{Nighttime\\scenes}
& $0.428$
& \good{$-0.063$}
& \good{$\bm{-0.081}$}
& \bad{$+0.010$}
& \good{$-0.005$}
& \good{$-0.078$}
& \good{$-0.066$}
& \good{$-0.016$}
& \good{$-0.051$}
& \good{$-0.036$} \\
\midrule

\makecell[l]{Other\\distractors/scenes}
& $0.317$
& \good{$-0.090$}
& \good{$-0.081$}
& \good{$-0.042$}
& \good{$-0.045$}
& \good{$-0.083$}
& \good{$-0.103$}
& \good{$-0.099$}
& \good{$\bm{-0.105}$}
& \good{$-0.090$} \\
\bottomrule
\end{tabular}
\end{adjustbox}
\end{table}
\\
\\
\textbf{View-dependent Static Objects Tradeoff}
DF3DV-1K contains 388 scenes with strong view-dependent static objects, enabling an investigation of the trade-off between preserving view-dependent static objects and removing distractors.
We further select 18 DF3DV-41 scenes with strong view-dependent static objects and evaluate them using three images per scene with human-annotated masks focused on view-dependent regions.
View-dependent static objects may be misclassified as distractors because their appearance changes lie between those of static and transient objects. 
Nevertheless, the benefits of removing distractors still outweigh the occasional misclassification, as shown in~\cref{tab:view_dep,fig:view_dep}.

\begin{table}[b]
\centering
\caption{\textbf{View-dependent static object benchmark.}
Removing distractors may accidentally affect the rendering quality of view-dependent effects on static objects. 
However, the benefits still outweigh this drawback, as evidenced by the better performance of distractor-free radiance field methods compared with 3DGS~\cite{kerbl20233d}.}
\label{tab:view_dep}
\resizebox{\columnwidth}{!}{
\begin{tabular}{l|cccccccccc}
\hline
 & 3DGS~\cite{kerbl20233d} & T-3DGS~\cite{markin2024t} & T-3DGS-TMR~\cite{markin2024t} & WildGaussian~\cite{kulhanek2024wildgaussians} & SLS~\cite{sabour2025spotlesssplats} & DeSplat~\cite{wang2025desplat} & DeGauss~\cite{wang2025degauss} & OCSplats~\cite{ling2025ocsplats} & RobustSplat~\cite{2025RobustSplat} & AsymGS~\cite{li2025asymgs} \\
\hline
PSNR  & 21.35 & 21.83 & 21.56 & 22.18 & 21.90 & 21.38 & \cellcolor{rankfive}{22.28} & 22.13 & \cellcolor{rankthree}{22.49} & \cellcolor{rankone}{23.19} \\
SSIM  & 0.769 & 0.796 & 0.788 & 0.797 & 0.790 & 0.781 & \cellcolor{rankfive}{0.806} & 0.802 & \cellcolor{rankthree}{0.807} & \cellcolor{rankone}{0.826} \\
LPIPS & 0.196 & 0.165 & 0.176 & 0.182 & 0.173 & 0.168 & \cellcolor{rankone}{0.147} & 0.161 & \cellcolor{rankfive}{0.151} & \cellcolor{rankthree}{0.149} \\
\hline
\end{tabular}
}
\end{table}
\begin{figure}[tb]
   \centering
   \includegraphics[width=\linewidth]{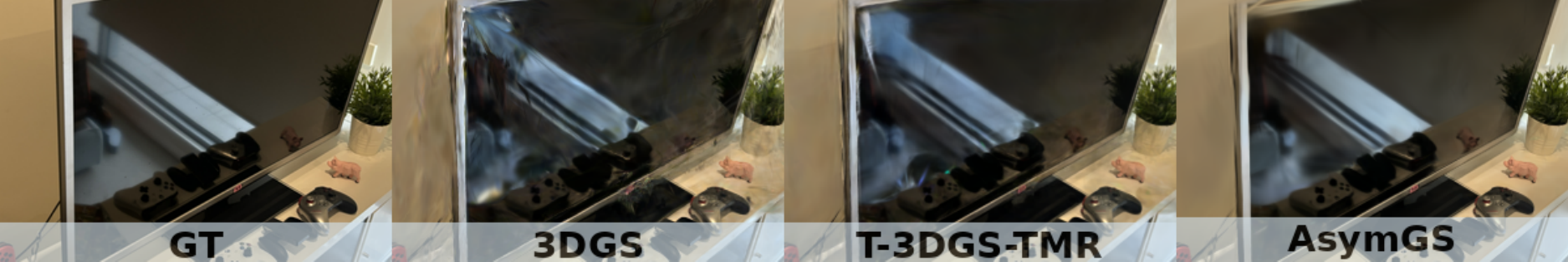}
    \caption{\textbf{Novel view synthesis of view-dependent static objects.} 
    Distractor-free radiance field methods~\cite{markin2024t, li2025asymgs} produce higher-quality novel views around the television screen, showing that the benefits of removing distractors outweigh the potential negative impact on view-dependent effects. 
    In addition, AsymGS~\cite{li2025asymgs} successfully renders the console reflected on the screen, demonstrating its ability to distinguish view-dependent static objects from distractors.}
    \label{fig:view_dep}
\end{figure}

\section{Beyond Scene-Specific Methods}
\label{sec:d2fix}

In this section, we provide additional experimental details, as well as per-method quantitative results, additional qualitative results, and failure cases of DI$^2$FIX.
We demonstrate how DF3DV-1K, a large-scale dataset, facilitates the development of approaches beyond scene-specific methods by introducing DI$^2$FIX, an enhancement module that improves the rendering quality of distractor-free radiance field methods~\cite{kerbl20233d, li2025asymgs, wang2025degauss, wang2025desplat, ling2025ocsplats, 2025RobustSplat, sabour2025spotlesssplats, markin2024t, kulhanek2024wildgaussians} and 3DGS~\cite{kerbl20233d} under distractor-free settings.
\\
\\
\textbf{Experimental Details.}
We introduce \textbf{DI$\bm{^2}$FIX} (Distractor-Free DIFIX), a plug-and-play 2D enhancement module designed to improve the visual quality of radiance field renderings.
The method is built upon DIFIX~\cite{wu2025difix3d+}, a diffusion-based image-to-image model~\cite{sauer2024adversarial,parmar2024one} originally developed to enhance degraded renderings under sparse-view settings using a clean reference image together with a degraded target view.
In distractor-free reconstruction scenarios, clean reference images are not available.
To adapt DIFIX~\cite{wu2025difix3d+} to this setting, we replace the clean reference input with renderings produced by radiance field models themselves.
This modification enables the model to learn corrections directly from reconstruction artifacts arising in cluttered real-world captures while preserving the original DIFIX~\cite{wu2025difix3d+} architecture.
Training data are constructed from DF3DV-1K* (DF3DV-1K excluding DF3DV-41).
Specifically, scenes are reconstructed from cluttered image collections using multiple radiance field methods~\cite{kerbl20233d,2025RobustSplat,ling2025ocsplats,markin2024t,kulhanek2024wildgaussians,wang2025desplat,wang2025degauss,li2025asymgs}, and novel views are rendered at camera poses corresponding to clean-image viewpoints.
Each rendered image is paired with its corresponding clean validation image, resulting in 316,890 candidate training pairs.
To improve training stability and remove severely corrupted samples, we retain only pairs whose perceptual similarity satisfies LPIPS $\leq \gamma$, where $\gamma = 0.5$ in our final configuration.
We then fine-tune DIFIX~\cite{wu2025difix3d+} to obtain DI$\bm{^2}$FIX using two NVIDIA L40 GPUs with a batch size of 1, requiring approximately 3-4 days to converge.
During adaptation, only minimal modifications are introduced, including replacing the clean reference with a radiance field rendering and disabling the Gram-matrix loss~\cite{gatys2015neural}, as we empirically observe negligible benefits from this term in distractor-free reconstruction scenarios.
All remaining training settings follow the original DIFIX~\cite{wu2025difix3d+} configuration.
\\
\\
\textbf{Per-method Improvement.}
\cref{tab:difix_all_metrics} reports the performance improvements of DIFIX~\cite{wu2025difix3d+}, DIFIX~\cite{wu2025difix3d+} fine-tuned on the RobustNeRF dataset~\cite{sabour2023robustnerf} (DIFIX+RobustNeRF), and DIFIX fine-tuned on DF3DV-1K* (DI$^2$FIX), measured relative to each radiance field method being enhanced.
DI$^2$FIX consistently improves all methods in terms of PSNR and LPIPS, achieving non-marginal performance gains.
Changes in SSIM are relatively limited, which is expected since DIFIX~\cite{wu2025difix3d+} does not employ an SSIM loss, and similar perceptual trade-offs with respect to SSIM have been reported in prior studies~\cite{blau2018perception, blau2019rethinking, li2025one, lin2025harnessing}.
In contrast, the original DIFIX~\cite{wu2025difix3d+} provides only marginal improvements across methods, while DIFIX+RobustNeRF shows slightly better performance than DIFIX~\cite{wu2025difix3d+} but still fails to produce consistent gains.
These results highlight the importance and value of domain-specific large-scale datasets.

\begin{figure}[tbh!]
  \centering
  \includegraphics[width=0.99\linewidth]{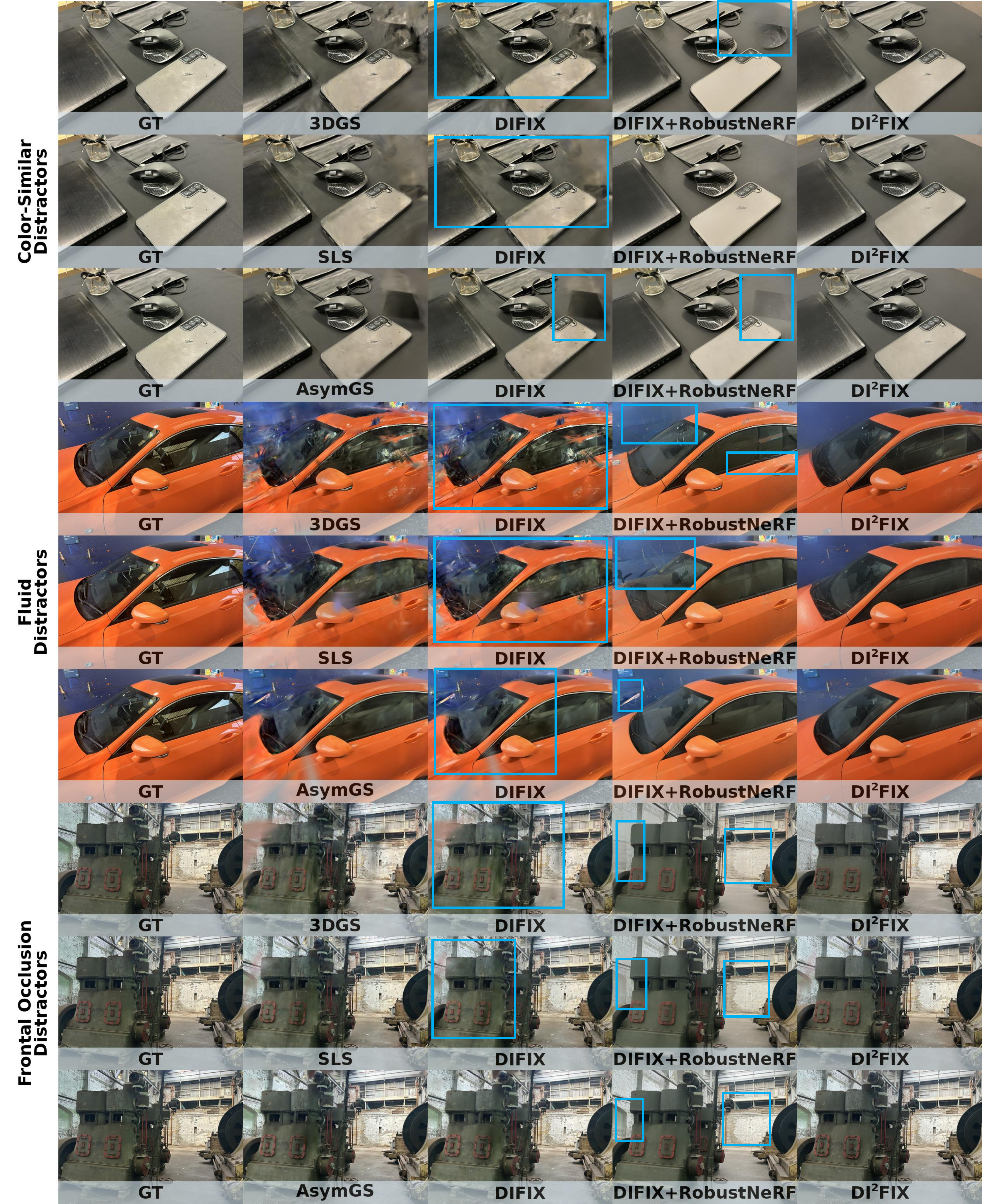}
  \caption{
\textbf{Qualitative comparison of enhancers on radiance-field outputs under color-similar, fluid, and frontal-occlusion distractor scenarios.}
Although DI$^2$FIX cannot always restore severely degraded regions, leveraging DF3DV-1K*, it shows promising results in mitigating distractor artifacts and improving visual quality across different methods.
}
  \label{fig:fix_41_1}
\end{figure}

\begin{figure}[tbh!]
  \centering
  \includegraphics[width=0.99\linewidth]{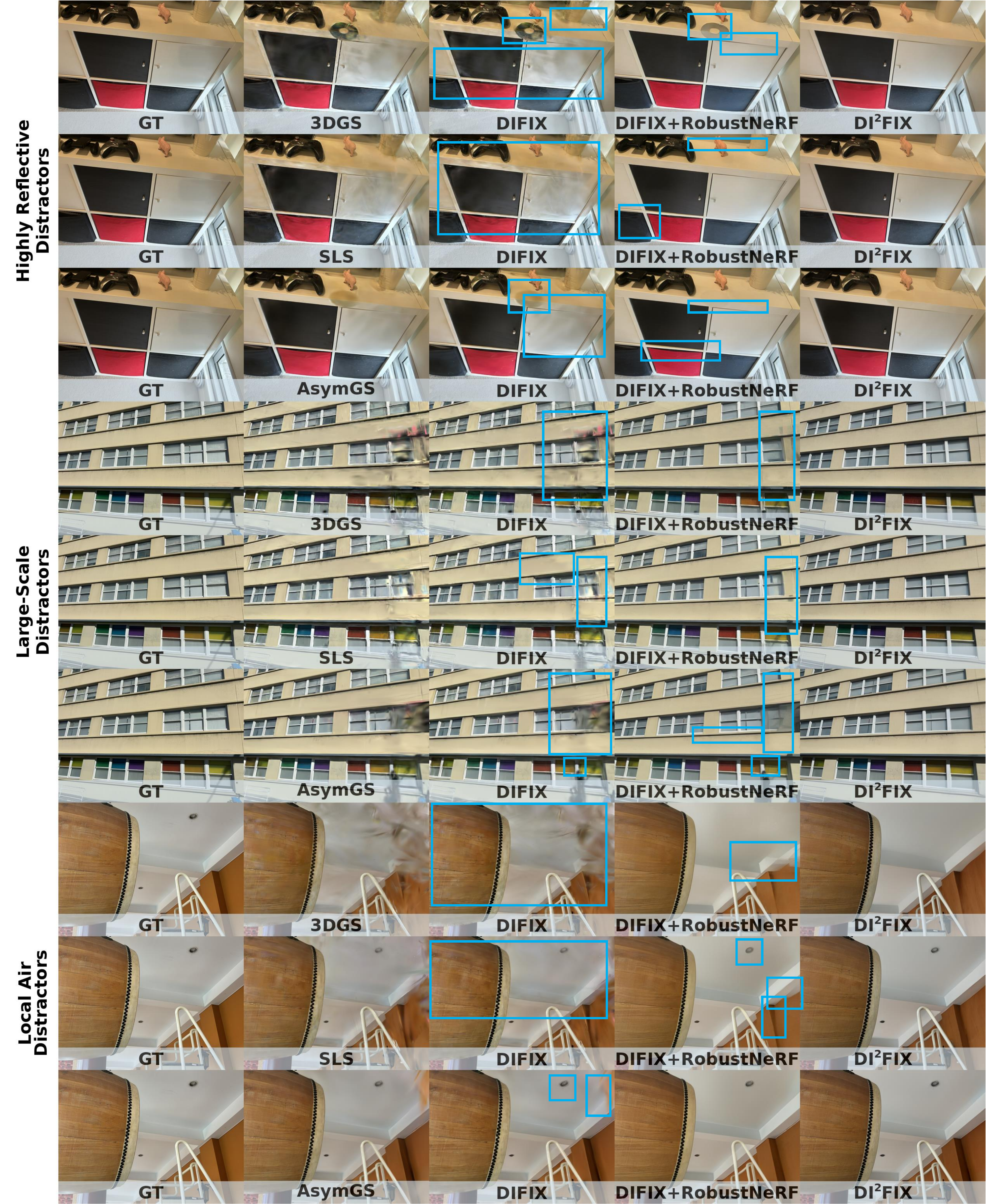}
  \caption{
\textbf{Qualitative comparison of enhancers on radiance-field outputs under highly reflective, large-scale, and local air distractor scenarios.}
Although DI$^2$FIX cannot always restore severely degraded regions, it shows promising results in mitigating distractor artifacts and improving visual quality across different methods.}
  \label{fig:fix_41_2}
\end{figure}

\begin{figure}[tbh!]
  \centering
  \includegraphics[width=0.99\linewidth]{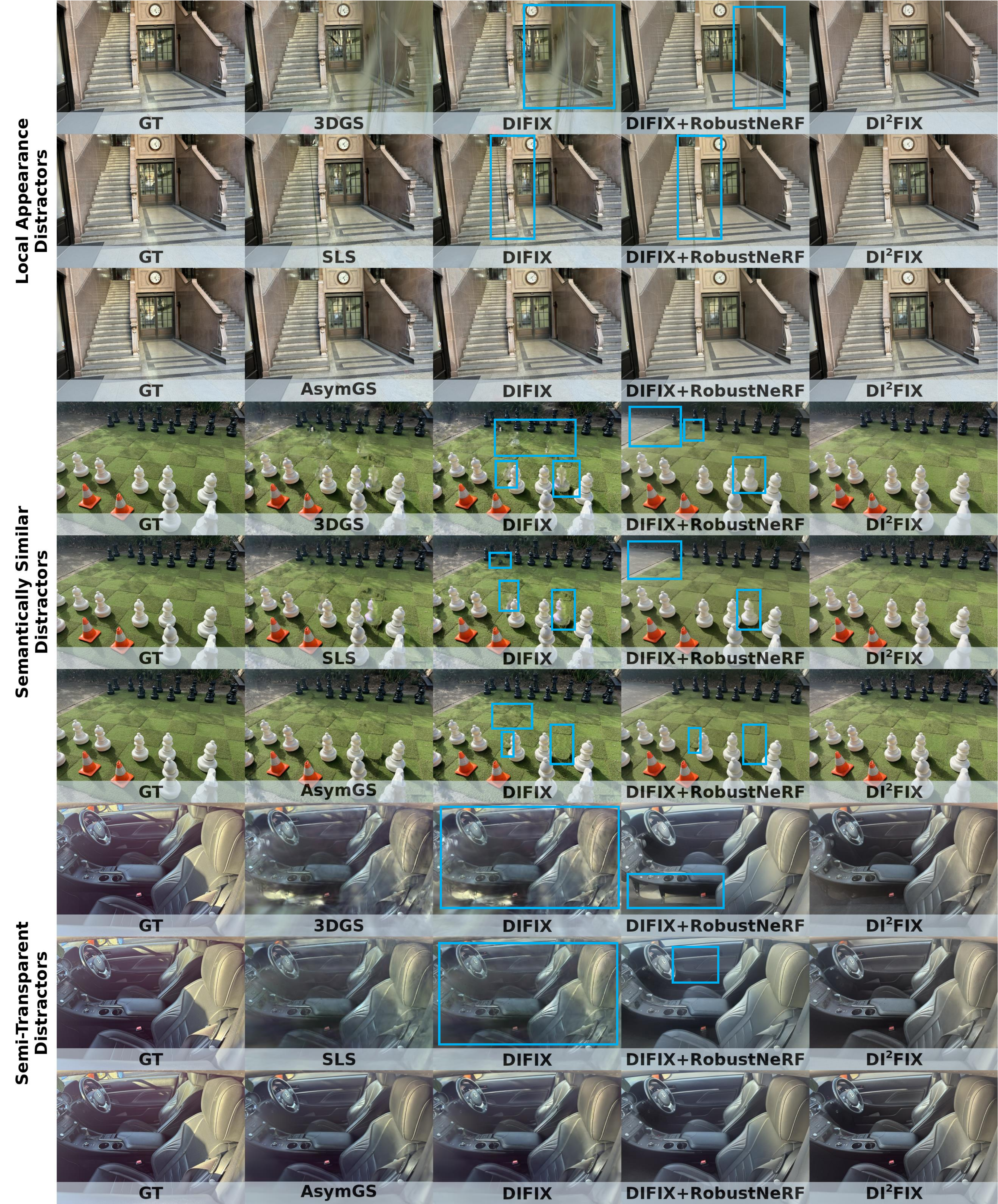}
  \caption{
\textbf{Qualitative comparison of enhancers on radiance-field outputs under local appearance, semantically similar, and semi-transparent distractor scenarios.} 
Although DI$^2$FIX cannot always restore severely degraded regions, it shows promising results in mitigating distractor artifacts and improving visual quality across different methods.}
  \label{fig:fix_41_3}
\end{figure}

\begin{figure}[tbh!]
  \centering
  \includegraphics[width=0.99\linewidth]{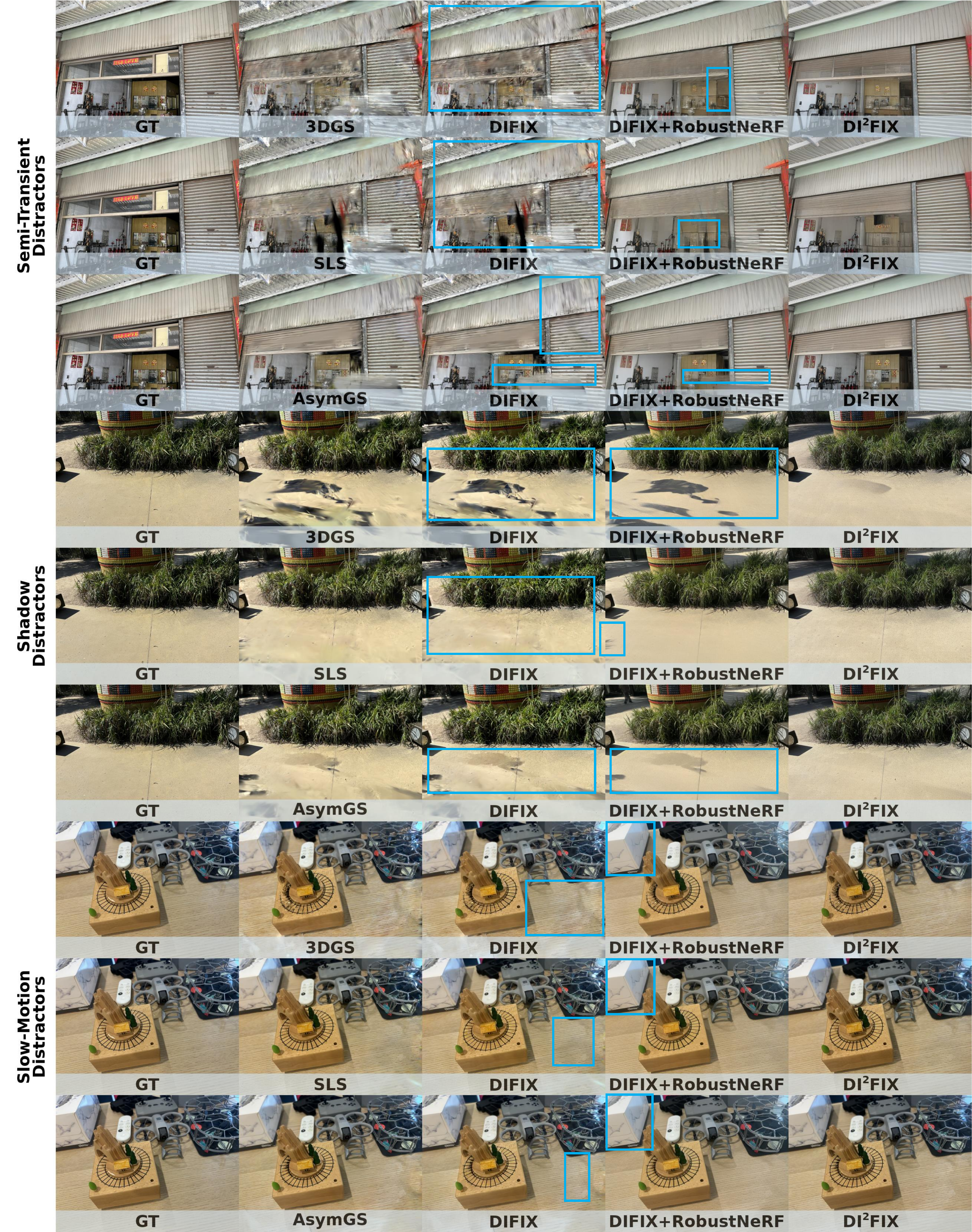}
  \caption{
\textbf{Qualitative comparison of enhancers on radiance-field outputs under semi-transient, shadow, and slow-motion distractor scenarios.} 
Although DI$^2$FIX cannot always restore severely degraded regions, it shows promising results in mitigating distractor artifacts and improving visual quality across different methods.}
  \label{fig:fix_41_4}
\end{figure}

\begin{figure}[tbh!]
  \centering
  \includegraphics[width=0.99\linewidth]{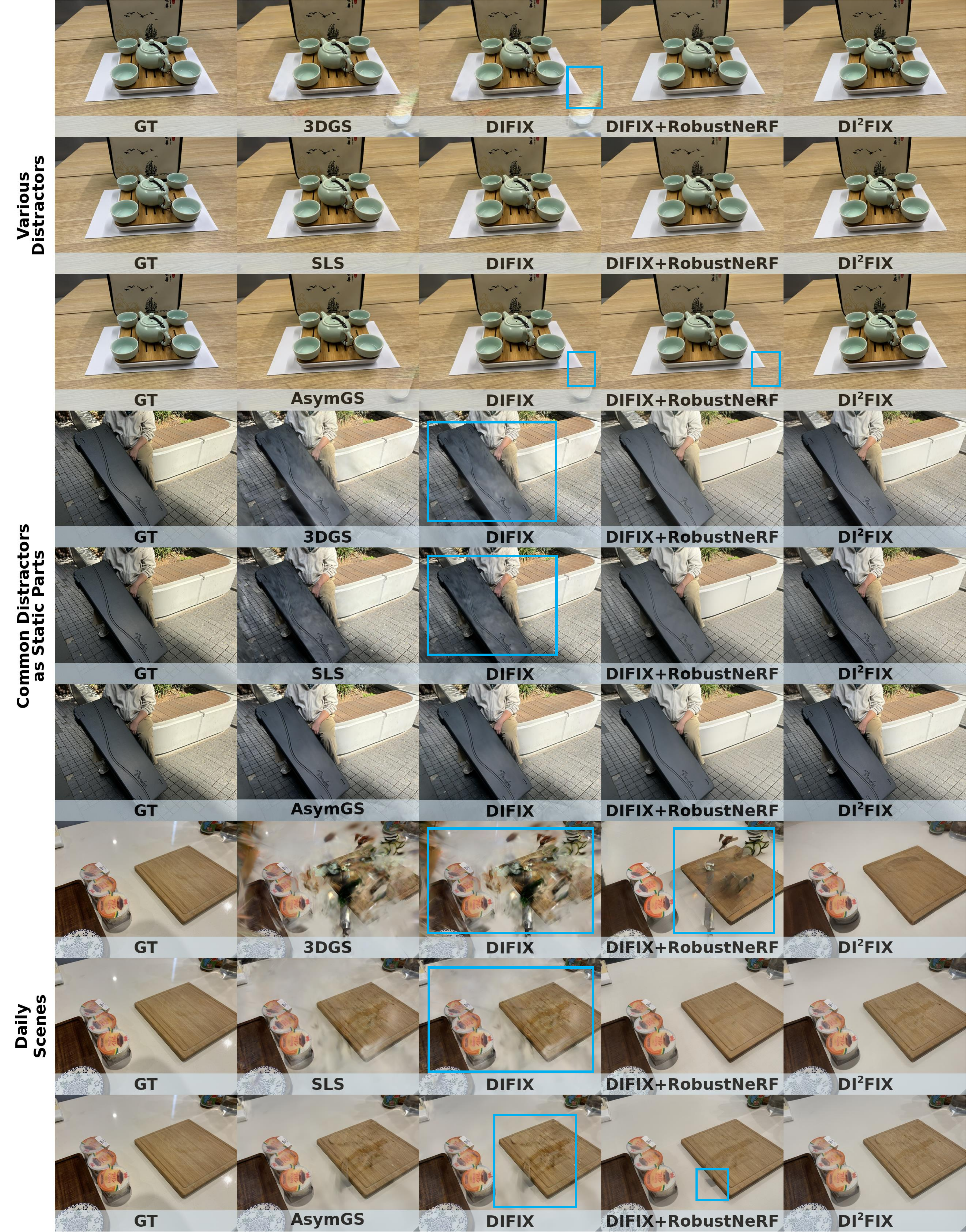}
  \caption{
\textbf{Qualitative comparison of enhancers on radiance-field outputs across various distractor scenarios, as well as common distractors treated as static parts and daily scene scenarios.}
Although DI$^2$FIX cannot always restore severely degraded regions, it shows promising results in mitigating distractor artifacts and improving visual quality across different methods.}
  \label{fig:fix_41_5}
\end{figure}

\begin{figure}[tbh!]
  \centering
  \includegraphics[width=0.99\linewidth]{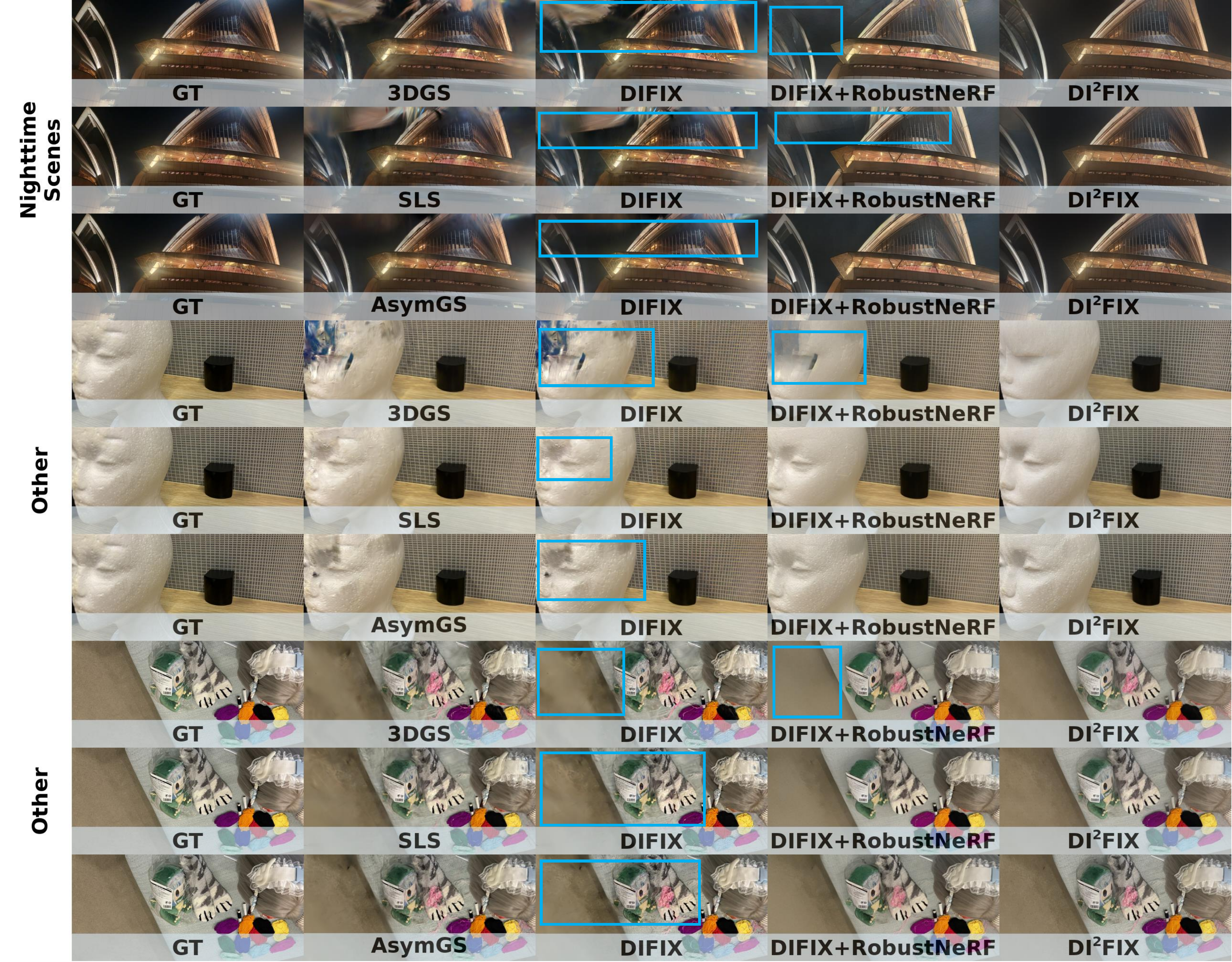}
  \caption{
\textbf{Qualitative comparison of enhancers on radiance-field outputs under nighttime and other scene scenarios.}
Although DI$^2$FIX cannot always restore severely degraded regions, it shows promising results in mitigating distractor artifacts and improving visual quality across different methods.}
  \label{fig:fix_41_6}
\end{figure}

We present enhanced results for all methods except SLS~\cite{sabour2025spotlesssplats} and AsymGS~\cite{li2025asymgs} in the main paper for qualitative comparisons.
Therefore, for additional qualitative evaluation, we select SLS~\cite{sabour2025spotlesssplats} and AsymGS~\cite{li2025asymgs}, together with 3DGS~\cite{kerbl20233d}, as target methods for enhancement.
\cref{fig:fix_41_1,fig:fix_41_2,fig:fix_41_3,fig:fix_41_4,fig:fix_41_5,fig:fix_41_6} show qualitative comparisons among DIFIX~\cite{wu2025difix3d+}, DIFIX+RobustNeRF, and DI$^2$FIX when applied to renderings produced by 3DGS~\cite{kerbl20233d}, SLS~\cite{sabour2025spotlesssplats}, and AsymGS~\cite{li2025asymgs}.
From a visual perspective, DI$^2$FIX produces more effective corrections than DIFIX and DIFIX+RobustNeRF, successfully removing distractor artifacts and inpainting occluded regions in most scenes.
In comparison, DIFIX tends to preserve the original rendering with limited corrections, while DIFIX+RobustNeRF often oversmooths fine details.
%\textcolor{blue}{talk about the figure...}
Specifically,~\cref{fig:fix_41_1} shows that DI$^2$FIX identifies and corrects color-similar distractor artifacts. 
For fluid distractors, DI$^2$FIX largely mitigates blending artifacts while preserving better background details, as illustrated by the AsymGS~\cite{li2025asymgs} row. 
In the frontal distractor scenario, DI$^2$FIX also preserves the black hole on the middle wall, which is often lost in DIFIX+RobustNeRF.
In~\cref{fig:fix_41_2}, DI$^2$FIX more effectively removes CD artifacts in the highly reflective distractor scenario and eliminates floaters in front of the building in the large-scale distractor scenario. 
It also successfully recovers the ceiling soffit in the 3DGS~\cite{kerbl20233d} and SLS~\cite{sabour2025spotlesssplats} rows in the local air distractor scenario.
~\cref{fig:fix_41_3} further shows that floaters in the local appearance distractor scenario are removed and the background is recovered. 
In the semantically similar distractor scenario, DI$^2$FIX more accurately identifies the dynamic chess pieces.
In~\cref{fig:fix_41_4}, although all methods fail to correct the rolling shutter artifact in front of the transparent window, DI$^2$FIX better removes the remaining fragments in the semi-transient distractor scenario and more effectively corrects shadow artifacts in the shadow distractor scenario. 
In the slow-motion scenario, DIFIX+RobustNeRF smooths the pattern on the tissue box.
Finally,~\cref{fig:fix_41_5} shows that DI$^2$FIX removes distractor artifacts across various distractor scenarios that remain in DIFIX+RobustNeRF results in the AsymGS~\cite{li2025asymgs} row, and also corrects vegetable artifacts in daily scene scenarios.
\\
\\
\textbf{Our-of-distribution Test.}
To assess the generalization capability of DI$^2$FIX to previously unseen radiance-field methods, we perform a leave-one-method-out cross-method training and evaluation protocol.
For each of the ten distractor-free radiance field methods~\cite{kerbl20233d,2025RobustSplat,ling2025ocsplats,markin2024t,kulhanek2024wildgaussians,wang2025desplat,wang2025degauss,li2025asymgs}, we exclude one method at a time and construct training pairs using renderings generated by the remaining nine methods. 
DIFIX is then fine-tuned on each reduced training set to obtain a corresponding DI$^2$FIX model, resulting in ten models in total, each trained without exposure to renderings from one specific method.
During evaluation, each model is tested on all methods, and performance differences are reported relative to the DI$^2$FIX model trained using renderings from all ten methods.

As shown in~\cref{tab:difix_cross_method_abl}, the performance variations remain small across all settings, with worst-case changes bounded within 0.1 PSNR, a 0.005 decrease in SSIM, and a 0.009 increase in LPIPS.
These results indicate stable out-of-distribution behavior and demonstrate that DI$^2$FIX generalizes effectively to radiance-field methods not observed during training.
Interestingly, excluding training data from AsymGS~\cite{li2025asymgs} occasionally yields slightly higher PSNR and SSIM, though the differences remain within the expected variance range.
\\
\\
\textbf{Limitations of DI$^2$FIX.}
As illustrated in~\cref{fig:fix_fail}, DI$^2$FIX may fail under extreme degradation or in cases affected by confirmation bias.
When the input views are heavily corrupted, the model lacks sufficient reliable visual evidence and therefore relies primarily on learned diffusion priors to reconstruct occluded regions, which can lead to reduced reconstruction fidelity.
In addition, when multiple views consistently contain a strong and visually unambiguous distractor, the model may incorrectly interpret the artifact as a valid scene component, resulting in confirmation bias during enhancement.
A potential solution is to incorporate multiple reference views to provide stronger supervision.
Exploring such multi-reference frameworks or view-selection approaches~\cite{lu2026hestia} is left for future work, as the primary goal of this work is to demonstrate the value of DF3DV-1K.

\begin{table*}[t]
\caption{
\textbf{Effect of enhancers on each radiance field method.}
Vanilla is the rendering results of the methods~\cite{kerbl20233d, li2025asymgs, wang2025degauss, wang2025desplat, ling2025ocsplats, 2025RobustSplat, sabour2025spotlesssplats, markin2024t, kulhanek2024wildgaussians}.
DIFIX+RobustNeRF and DI$^2$FIX are DIFIX~\cite{wu2025difix3d+} fine-tuned on RobustNeRF~\cite{sabour2023robustnerf} and DF3DV-1K*, respectively.
Values report the mean performance change on On-the-go~\cite{Ren2024NeRF} and DF3DV-41 relative to the Vanilla.
Positive values indicate improvement for PSNR and SSIM, while negative values indicate improvement for LPIPS.
}
\label{tab:difix_all_metrics}
\centering
\begin{adjustbox}{width=\textwidth}
\begin{tabular}{cccccccccc}
\toprule
\makecell{\textbf{3DGS}\\\cite{kerbl20233d}} &
\makecell{\textbf{T-3DGS}\\\cite{markin2024t}} &
\makecell{\textbf{T-3DGS-TMR}\\\cite{markin2024t}} &
\makecell{\textbf{WildGaussian}\\\cite{kulhanek2024wildgaussians}} &
\makecell{\textbf{SLS}\\\cite{sabour2025spotlesssplats}} &
\makecell{\textbf{DeSplat}\\\cite{wang2025desplat}} &
\makecell{\textbf{DeGauss}\\\cite{wang2025degauss}} &
\makecell{\textbf{OCSplats}\\\cite{ling2025ocsplats}} &
\makecell{\textbf{RobustSplat}\\\cite{2025RobustSplat}} &
\makecell{\textbf{AsymGS}\\\cite{li2025asymgs}} \\
\midrule

% ===================== PSNR =====================
%\rowcolor{gray!20}
\multicolumn{10}{c}{\textbf{PSNR$\uparrow$}}\\
\rowcolor{gray!15}
\multicolumn{10}{l}{Vanilla}\\
$18.71$ & $20.90$ & $19.12$ & $20.93$ & $21.05$ & $21.02$ & $21.77$ & $21.41$ & $21.52$ & $21.75$ \\

\rowcolor{gray!15}
\multicolumn{10}{l}{DIFIX}\\
$\bad{-0.19}$ & $\bad{-0.62}$ & $\bad{-0.27}$ & $\bad{-0.64}$ & $\bad{-0.66}$ &
$\bad{-0.80}$ & $\bad{-0.94}$ & $\bad{-0.80}$ & $\bad{-0.83}$ & $\bad{-0.85}$ \\

\rowcolor{gray!15}
\multicolumn{10}{l}{DIFIX+RobustNeRF}\\
$\good{+0.41}$ & $\bad{-0.12}$ & $\good{+0.53}$ & $\bad{-0.37}$ & $\bad{-0.50}$ &
$\bad{-0.29}$ & $\bad{-0.71}$ & $\bad{-0.46}$ & $\bad{-0.56}$ & $\bad{-0.65}$ \\

\rowcolor{gray!15}
\multicolumn{10}{l}{DI$^2$FIX}\\
$\good{\bm{+1.83}}$ & $\good{\bm{+1.26}}$ & $\good{\bm{+2.02}}$ & $\good{\bm{+0.77}}$ & $\good{\bm{+0.67}}$ &
$\good{\bm{+0.86}}$ & $\good{\bm{+0.47}}$ & $\good{\bm{+0.63}}$ & $\good{\bm{+0.69}}$ & $\good{\bm{+0.39}}$ \\

\midrule

% ===================== SSIM =====================
%\rowcolor{gray!20}
\multicolumn{10}{c}{\textbf{SSIM$\uparrow$}}\\
\rowcolor{gray!15}
\multicolumn{10}{l}{Vanilla}\\
$0.641$ & $0.745$ & $0.695$ & $0.732$ & $0.721$ & $0.733$ & $0.763$ & $0.753$ & $0.760$ & $0.767$ \\

\rowcolor{gray!15}
\multicolumn{10}{l}{DIFIX}\\
$\bad{-0.047}$ & $\bad{-0.068}$ & $\bad{-0.061}$ & $\bad{-0.065}$ & $\bad{-0.060}$ &
$\bad{-0.072}$ & $\bad{-0.075}$ & $\bad{-0.076}$ & $\bad{-0.076}$ & $\bad{-0.075}$ \\

\rowcolor{gray!15}
\multicolumn{10}{l}{DIFIX+RobustNeRF}\\
$\bad{-0.014}$ & $\bad{-0.033}$ & $\bad{-0.019}$ & $\bad{-0.039}$ & $\bad{-0.035}$ &
$\bad{-0.034}$ & $\bad{-0.047}$ & $\bad{-0.042}$ & $\bad{-0.045}$ & $\bad{-0.050}$ \\

\rowcolor{gray!15}
\multicolumn{10}{l}{DI$^2$FIX}\\
$\good{\bm{+0.023}}$ & $\good{\bm{+0.011}}$ & $\good{\bm{+0.026}}$ & $\good{\bm{+0.001}}$ & $\good{\bm{+0.002}}$ &
$\good{\bm{+0.008}}$ & $\bad{\bm{-0.003}}$ & $\good{\bm{+0.000}}$ & $\good{\bm{+0.000}}$ & $\bad{\bm{-0.005}}$ \\

\midrule

% ===================== LPIPS =====================
%\rowcolor{gray!20}
\multicolumn{10}{c}{\textbf{LPIPS$\downarrow$}}\\
\rowcolor{gray!15}
\multicolumn{10}{l}{Vanilla}\\
$0.319$ & $0.193$ & $0.255$ & $0.220$ & $0.220$ & $0.184$ & $0.172$ & $0.178$ & $0.172$ & $0.192$ \\

\rowcolor{gray!15}
\multicolumn{10}{l}{DIFIX}\\
$\good{-0.001}$ & $\bad{+0.012}$ & $\bad{+0.001}$ & $\bad{+0.009}$ & $\bad{+0.011}$ &
$\bad{+0.019}$ & $\bad{+0.020}$ & $\bad{+0.018}$ & $\bad{+0.018}$ & $\bad{+0.014}$ \\

\rowcolor{gray!15}
\multicolumn{10}{l}{DIFIX+RobustNeRF}\\
$\good{-0.047}$ & $\good{-0.005}$ & $\good{-0.032}$ & $\good{-0.007}$ & $\good{-0.009}$ &
$\good{-0.000}$ & $\bad{+0.008}$ & $\bad{+0.003}$ & $\bad{+0.006}$ & $\bad{+0.001}$ \\

\rowcolor{gray!15}
\multicolumn{10}{l}{DI$^2$FIX}\\
$\good{\bm{-0.111}}$ & $\good{\bm{-0.050}}$ & $\good{\bm{-0.085}}$ & $\good{\bm{-0.058}}$ & $\good{\bm{-0.059}}$ &
$\good{\bm{-0.041}}$ & $\good{\bm{-0.036}}$ & $\good{\bm{-0.038}}$ & $\good{\bm{-0.036}}$ & $\good{\bm{-0.049}}$ \\

\bottomrule
\end{tabular}
\end{adjustbox}
\end{table*}

\begin{figure}[tbh!]
  \centering
  \includegraphics[width=0.99\linewidth]{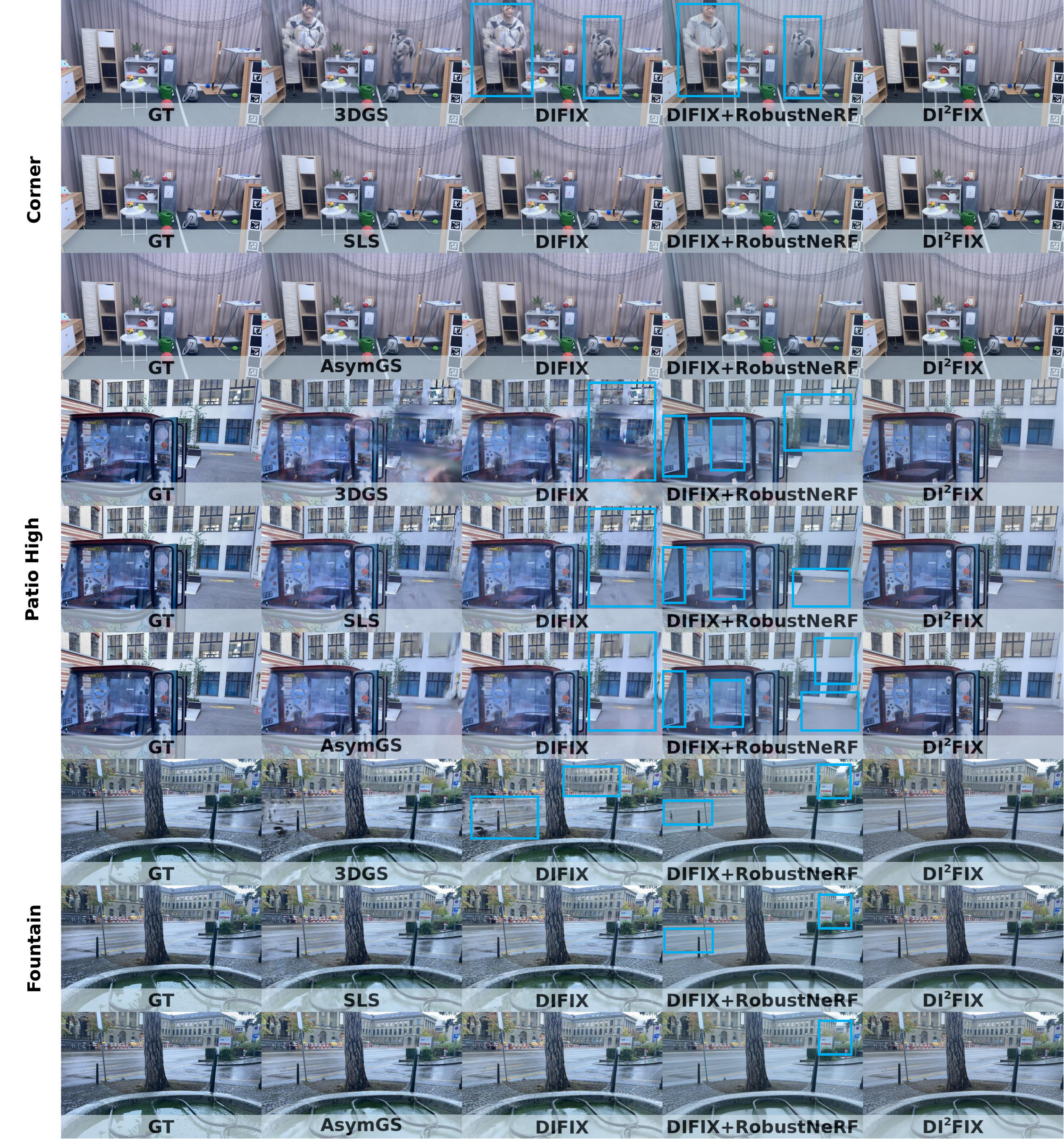}
  \caption{
\textbf{Qualitative comparison of enhancers on radiance-field outputs for the Corner, Patio High, and Fountain scenes from On-the-go~\cite{Ren2024NeRF}.}
DI$^2$FIX significantly enhances the results of 3DGS~\cite{kerbl20233d}. 
Although recent state-of-the-art methods already perform well on the On-the-go~\cite{Ren2024NeRF} dataset, DI$^2$FIX can further improve rendering quality when artifacts are present (e.g., the window in the background of the Patio High scene).
}
\label{fig:fix_onthego_1}
\end{figure}

\begin{table}[t]
\caption{
\textbf{Cross-method ablation of DI$^2$FIX.}
Each “Without” row denotes a leave-one-method-out setting, where DI$^2$FIX is trained without data from that method and evaluated on all methods. 
Values report the mean performance change relative to DI$^2$FIX trained on all ten methods. 
Positive values indicate improvement for PSNR/SSIM, while negative values indicate improvement for LPIPS. 
Across all metrics, performance drops remain small, with maximum decreases of 0.1 dB in PSNR, 0.005 in SSIM, and 0.009 in LPIPS, demonstrating strong cross-method generalization. 
We do not observe reduced generalization to the held-out method, as indicated by the diagonal entries.
Although the differences fall within the error bars, it is interesting to observe that excluding AsymGS~\cite{li2025asymgs} results in slightly better PSNR and SSIM.
}
\label{tab:difix_cross_method_abl}
\centering
\setlength{\tabcolsep}{3.5pt}
\begin{adjustbox}{width=\textwidth}
\begin{tabular}{l|cccccccccc}
\toprule
  \textbf{Without} &
  \makecell{\textbf{3DGS}\\\cite{kerbl20233d}} &
  \makecell{\textbf{T-3DGS}\\\cite{markin2024t}} &
  \makecell{\textbf{T-3DGS-TMR}\\\cite{markin2024t}} &
  \makecell{\textbf{WildGaussian}\\\cite{kulhanek2024wildgaussians}} &
  \makecell{\textbf{SLS}\\\cite{sabour2025spotlesssplats}} &
  \makecell{\textbf{DeSplat}\\\cite{wang2025desplat}} &
  \makecell{\textbf{DeGauss}\\\cite{wang2025degauss}} &
  \makecell{\textbf{OCSplats}\\\cite{ling2025ocsplats}} &
  \makecell{\textbf{RobustSplat}\\\cite{2025RobustSplat}} &
  \makecell{\textbf{AsymGS}\\\cite{li2025asymgs}} \\
\midrule

% ======================= PSNR =======================
%\rowcolor{gray!15}
\multicolumn{11}{c}{\textbf{PSNR}$\uparrow$} \\
\rowcolor{gray!15}
DI$^2$FIX &
20.55 & 22.16 & 21.15 & 21.71 & 21.73 &
21.89 & 22.25 & 22.05 & 22.21 & 22.14 \\
\midrule
3DGS~\cite{kerbl20233d} & \bad{$-0.10$} & \bad{$-0.02$} & \bad{$-0.04$} & \bad{$-0.07$} & \bad{$-0.08$} &
\bad{$-0.01$} & \bad{$-0.04$} & \bad{$-0.04$} & \bad{$-0.03$} & \bad{$-0.04$} \\
T-3DGS~\cite{markin2024t} & \good{$+0.01$} & \good{$+0.05$} & \good{$+0.03$} & \good{$+0.03$} & \good{$+0.03$} &
\good{$+0.07$} & \good{$+0.02$} & \good{$+0.05$} & \good{$+0.04$} & \good{$+0.04$} \\
T-3DGS-TMR~\cite{markin2024t} & \good{$+0.00$} & \good{$+0.00$} & \good{$+0.01$} & \good{$+0.01$} & \bad{$-0.03$} &
\good{$+0.00$} & \bad{$-0.04$} & \good{$+0.00$} & \good{$+0.00$} & \good{$+0.02$} \\
WildGaussian~\cite{kulhanek2024wildgaussians} & \good{$+0.07$} & \good{$+0.06$} & \good{$+0.10$} & \good{$+0.05$} & \good{$+0.05$} &
\good{$+0.04$} & \good{$+0.03$} & \good{$+0.05$} & \good{$+0.05$} & \good{$+0.06$} \\
SLS~\cite{sabour2025spotlesssplats} & \bad{$-0.08$} & \bad{$-0.01$} & \good{$+0.00$} & \bad{$-0.05$} & \bad{$-0.05$} &
\bad{$-0.02$} & \bad{$-0.05$} & \good{$+0.00$} & \bad{$-0.03$} & \good{$+0.00$} \\
DeSplat~\cite{wang2025desplat} & \bad{$-0.10$} & \good{$+0.00$} & \bad{$-0.02$} & \bad{$-0.02$} & \bad{$-0.04$} &
\bad{$-0.08$} & \bad{$-0.07$} & \good{$+0.01$} & \bad{$-0.01$} & \bad{$-0.02$} \\
DeGauss~\cite{wang2025degauss} & \good{$+0.02$} & \bad{$-0.03$} & \good{$+0.04$} & \bad{$-0.04$} & \bad{$-0.08$} &
\good{$+0.00$} & \bad{$-0.08$} & \bad{$-0.01$} & \bad{$-0.01$} & \bad{$-0.02$} \\
OCSplats~\cite{ling2025ocsplats} & \bad{$-0.05$} & \bad{$-0.08$} & \bad{$-0.04$} & \bad{$-0.07$} & \bad{$-0.04$} &
\bad{$-0.09$} & \bad{$-0.09$} & \bad{$-0.06$} & \bad{$-0.05$} & \bad{$-0.02$} \\
RobustSplat~\cite{2025RobustSplat} & \bad{$-0.03$} & \bad{$-0.05$} & \good{$+0.00$} & \bad{$-0.03$} & \bad{$-0.05$} &
\bad{$-0.02$} & \bad{$-0.04$} & \good{$+0.00$} & \good{$+0.00$} & \good{$+0.00$} \\
AsymGS~\cite{li2025asymgs} & \good{$+0.07$} & \good{$+0.05$} & \good{$+0.10$} & \good{$+0.05$} & \good{$+0.01$} &
\good{$+0.10$} & \good{$+0.03$} & \good{$+0.05$} & \good{$+0.04$} & \good{$+0.07$} \\

% ======================= SSIM =======================
\midrule
%\rowcolor{gray!15}
\multicolumn{11}{c}{\textbf{SSIM}$\uparrow$} \\
\rowcolor{gray!15}
DI$^2$FIX &
0.664 & 0.756 & 0.722 & 0.733 & 0.724 &
0.742 & 0.759 & 0.753 & 0.760 & 0.761 \\
\midrule
3DGS~\cite{kerbl20233d} & \bad{$-0.002$} & \good{$+0.000$} & \good{$+0.000$} & \bad{$-0.002$} & \bad{$-0.002$} &
\good{$+0.000$} & \bad{$-0.001$} & \bad{$-0.001$} & \bad{$-0.001$} & \bad{$-0.002$} \\
T-3DGS~\cite{markin2024t} & \good{$+0.000$} & \bad{$-0.001$} & \good{$+0.000$} & \bad{$-0.002$} & \bad{$-0.001$} &
\good{$+0.000$} & \bad{$-0.002$} & \bad{$-0.001$} & \bad{$-0.002$} & \bad{$-0.002$} \\
T-3DGS-TMR~\cite{markin2024t} & \bad{$-0.002$} & \bad{$-0.002$} & \bad{$-0.002$} & \bad{$-0.002$} & \bad{$-0.002$} &
\bad{$-0.001$} & \bad{$-0.003$} & \bad{$-0.002$} & \bad{$-0.002$} & \bad{$-0.003$} \\
WildGaussian~\cite{kulhanek2024wildgaussians}& \good{$+0.001$} & \good{$+0.000$} & \good{$+0.000$} & \good{$+0.000$} & \good{$+0.000$} &
\good{$+0.000$} & \good{$+0.000$} & \good{$+0.000$} & \good{$+0.000$} & \good{$+0.000$} \\
SLS~\cite{sabour2025spotlesssplats} & \bad{$-0.001$} & \bad{$-0.003$} & \bad{$-0.002$} & \bad{$-0.003$} & \bad{$-0.003$} &
\bad{$-0.003$} & \bad{$-0.004$} & \bad{$-0.003$} & \bad{$-0.004$} & \bad{$-0.004$} \\
DeSplat~\cite{wang2025desplat} & \bad{$-0.005$} & \bad{$-0.003$} & \bad{$-0.004$} & \bad{$-0.004$} & \bad{$-0.005$} &
\bad{$-0.004$} & \bad{$-0.005$} & \bad{$-0.004$} & \bad{$-0.004$} & \bad{$-0.005$} \\
DeGauss~\cite{wang2025degauss} & \bad{$-0.001$} & \bad{$-0.001$} & \good{$+0.000$} & \bad{$-0.002$} & \bad{$-0.003$} &
\bad{$-0.001$} & \bad{$-0.003$} & \bad{$-0.002$} & \bad{$-0.003$} & \bad{$-0.003$} \\
OCSplats~\cite{ling2025ocsplats} & \bad{$-0.002$} & \bad{$-0.002$} & \bad{$-0.002$} & \bad{$-0.003$} & \bad{$-0.003$} &
\bad{$-0.002$} & \bad{$-0.003$} & \bad{$-0.002$} & \bad{$-0.003$} & \bad{$-0.003$} \\
RobustSplat~\cite{2025RobustSplat} & \bad{$-0.002$} & \bad{$-0.002$} & \good{$+0.000$} & \bad{$-0.002$} & \bad{$-0.003$} &
\bad{$-0.001$} & \bad{$-0.003$} & \bad{$-0.002$} & \bad{$-0.002$} & \bad{$-0.002$} \\
AsymGS~\cite{li2025asymgs} & \good{$+0.000$} & \good{$+0.000$} & \good{$+0.001$} & \good{$+0.000$} & \good{$+0.000$} &
\good{$+0.001$} & \good{$+0.000$} & \good{$+0.000$} & \good{$+0.000$} & \good{$+0.000$} \\

% ======================= LPIPS =======================
\midrule
%\rowcolor{gray!15}
\multicolumn{11}{c}{\textbf{LPIPS}$\downarrow$} \\
\rowcolor{gray!15}
DI$^2$FIX &
0.208 & 0.142 & 0.169 & 0.161 & 0.161 &
0.143 & 0.135 & 0.140 & 0.136 & 0.142 \\
\midrule
3DGS~\cite{kerbl20233d} & \bad{$+0.007$} & \bad{$+0.003$} & \bad{$+0.004$} & \bad{$+0.006$} & \bad{$+0.005$} &
\bad{$+0.005$} & \bad{$+0.004$} & \bad{$+0.004$} & \bad{$+0.004$} & \bad{$+0.004$} \\
T-3DGS~\cite{markin2024t} & \bad{$+0.006$} & \bad{$+0.004$} & \bad{$+0.004$} & \bad{$+0.006$} & \bad{$+0.005$} &
\bad{$+0.005$} & \bad{$+0.005$} & \bad{$+0.005$} & \bad{$+0.005$} & \bad{$+0.006$} \\
T-3DGS-TMR~\cite{markin2024t} & \bad{$+0.003$} & \bad{$+0.002$} & \bad{$+0.002$} & \bad{$+0.002$} & \bad{$+0.003$} &
\bad{$+0.003$} & \bad{$+0.002$} & \bad{$+0.003$} & \bad{$+0.002$} & \bad{$+0.002$} \\
WildGaussian~\cite{kulhanek2024wildgaussians} & \bad{$+0.001$} & \good{$-0.000$} & \good{$-0.000$} & \bad{$+0.005$} & \good{$-0.000$} &
\bad{$+0.001$} & \bad{$+0.001$} & \good{$-0.000$} & \bad{$+0.001$} & \bad{$+0.003$} \\
SLS~\cite{sabour2025spotlesssplats} & \bad{$+0.007$} & \bad{$+0.002$} & \bad{$+0.001$} & \bad{$+0.004$} & \bad{$+0.006$} &
\bad{$+0.003$} & \bad{$+0.004$} & \bad{$+0.004$} & \bad{$+0.004$} & \bad{$+0.004$} \\
DeSplat~\cite{wang2025desplat} & \bad{$+0.006$} & \bad{$+0.005$} & \bad{$+0.004$} & \bad{$+0.008$} & \bad{$+0.006$} &
\bad{$+0.006$} & \bad{$+0.006$} & \bad{$+0.006$} & \bad{$+0.006$} & \bad{$+0.007$} \\
DeGauss~\cite{wang2025degauss} & \bad{$+0.005$} & \bad{$+0.004$} & \bad{$+0.003$} & \bad{$+0.008$} & \bad{$+0.007$} &
\bad{$+0.006$} & \bad{$+0.007$} & \bad{$+0.006$} & \bad{$+0.006$} & \bad{$+0.007$} \\
OCSplats~\cite{ling2025ocsplats} & \bad{$+0.009$} & \bad{$+0.005$} & \bad{$+0.005$} & \bad{$+0.009$} & \bad{$+0.008$} &
\bad{$+0.007$} & \bad{$+0.007$} & \bad{$+0.007$} & \bad{$+0.007$} & \bad{$+0.007$} \\
RobustSplat~\cite{2025RobustSplat} & \bad{$+0.006$} & \bad{$+0.004$} & \bad{$+0.003$} & \bad{$+0.005$} & \bad{$+0.006$} &
\bad{$+0.005$} & \bad{$+0.005$} & \bad{$+0.005$} & \bad{$+0.005$} & \bad{$+0.004$} \\
AsymGS~\cite{li2025asymgs} & \bad{$+0.004$} & \bad{$+0.001$} & \good{$-0.000$} & \bad{$+0.004$} & \bad{$+0.004$} &
\bad{$+0.004$} & \bad{$+0.003$} & \bad{$+0.004$} & \bad{$+0.004$} & \bad{$+0.005$} \\

\bottomrule
\end{tabular}
\end{adjustbox}
\end{table}

\begin{figure}[tbh!]
\centering
\includegraphics[width=0.99\linewidth]{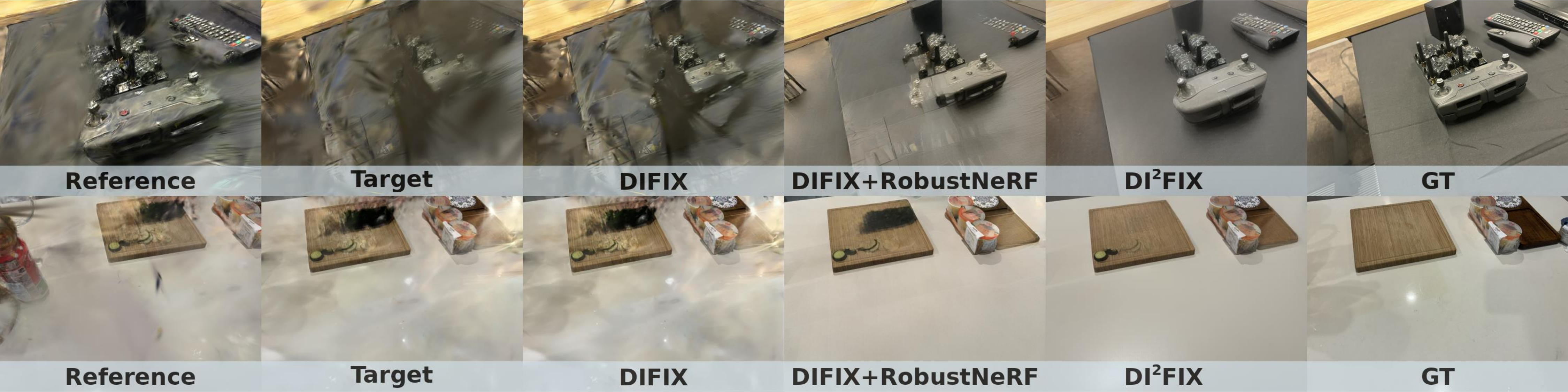}
\caption{
\textbf{Failure cases of DI$^2$FIX.}
Each row shows the reference view, the target view to be fixed, the results of DIFIX~\cite{wu2025difix3d+}, DIFIX+RobustNeRF, and DI$^2$FIX, and the ground truth.
While DI$^2$FIX improves visual quality in many regions, it can struggle under extremely severe degradations affecting the reference or target views (e.g., large floaters in both the target and reference views in row~1), or under cross-view confirmation bias (e.g., the vegetable-like distractor appearing in both the target and reference views in row~2), where artifacts may persist or scene structures cannot be fully recovered.
}
\label{fig:fix_fail}
\end{figure}

\section{DF3DV-1K and DF3DV-41}
\label{sec:df3dv}

\begin{figure}[tb]
  \centering
  \includegraphics[width=0.99\linewidth]{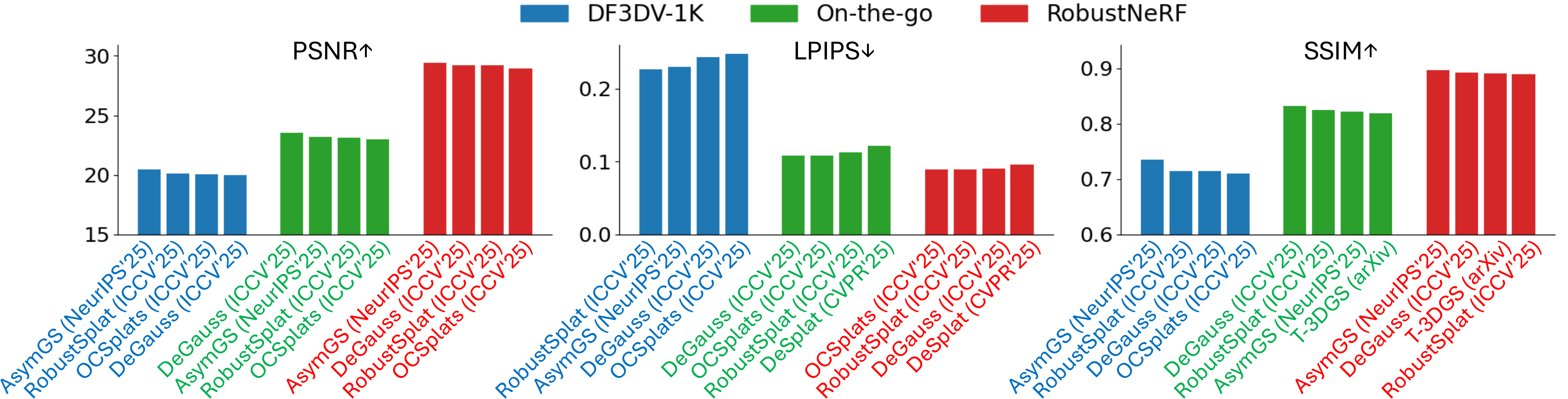}
  \caption{
\textbf{Top-4 performing methods on each dataset.}
Results on DF3DV-1K identify AsymGS~\cite{li2025asymgs} and RobustSplat~\cite{2025RobustSplat} as the most robust methods, followed by OCSplats~\cite{ling2025ocsplats} and DeGauss~\cite{wang2025degauss}.
This ranking differs from those observed on On-the-go~\cite{Ren2024NeRF} and RobustNeRF~\cite{sabour2023robustnerf}, highlighting the importance of large-scale evaluation.
Interestingly, performance trends on DF3DV-1K largely follow publication chronology, with AsymGS~\cite{li2025asymgs} (NeurIPS 2025) outperforming RobustSplat~\cite{2025RobustSplat}, OCSplats~\cite{ling2025ocsplats}, and DeGauss~\cite{wang2025degauss} (ICCV 2025), while the three ICCV 2025 methods exhibit similar performance in PSNR and SSIM.
Such trends are less evident on existing datasets.
}
\label{fig:rank}
\end{figure}

\begin{figure}[tb]
  \centering
  \includegraphics[width=0.99\linewidth]{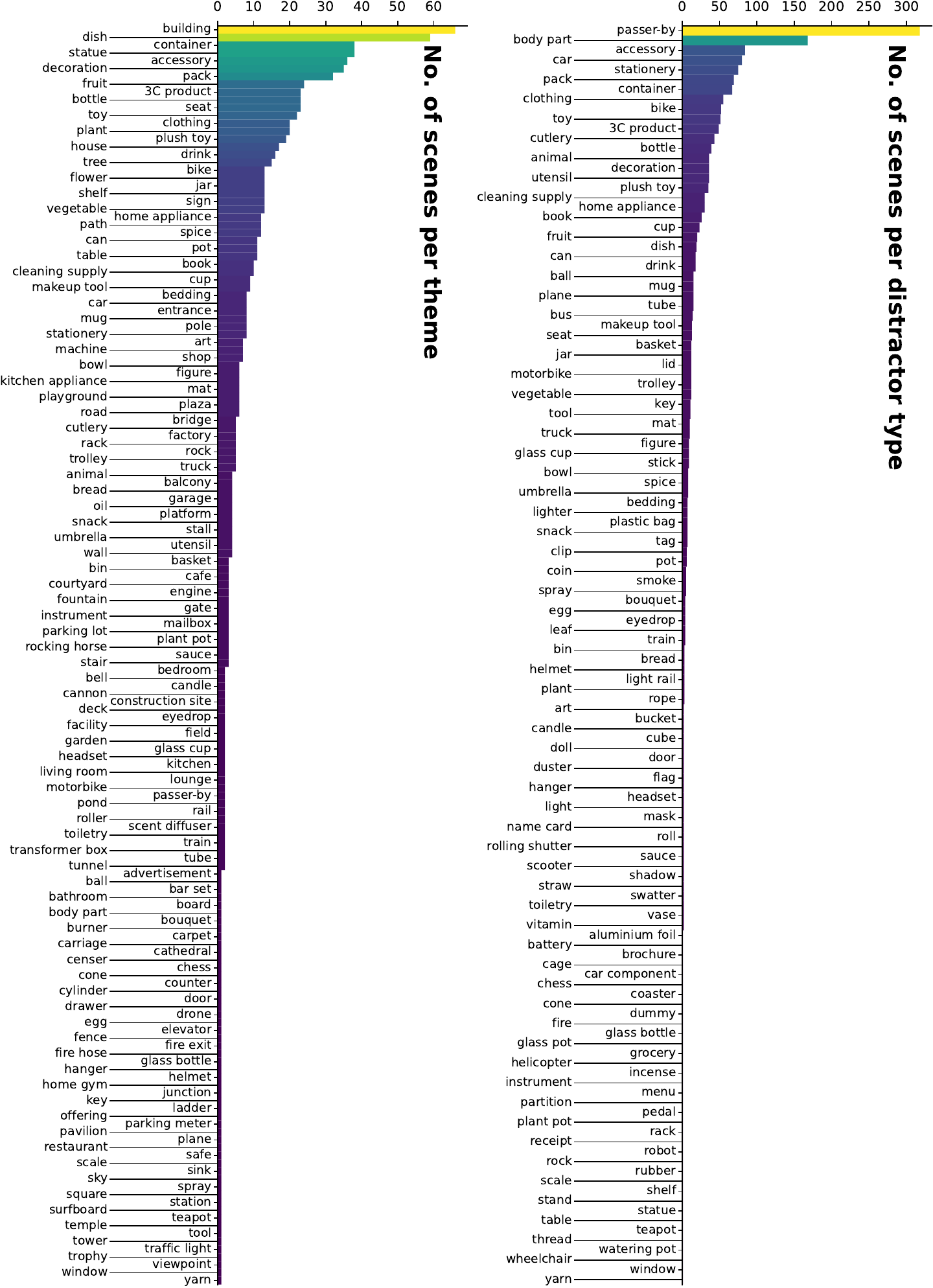}
  \caption{
\textbf{Distribution of scenes by theme and distractor type.}
Number of scenes per theme (left) and per distractor type (right). 
DF3DV-1K includes 161 themes and 128 distractor types.}
  \label{fig:scene_and_theme}
\end{figure}

\begin{figure}[tb]
  \centering
  \includegraphics[width=\linewidth]{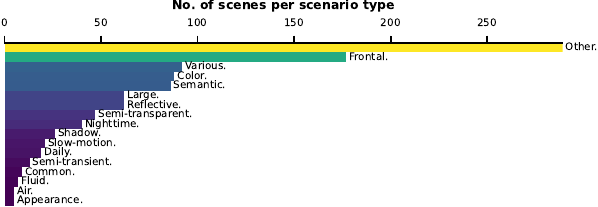}
   \caption{\textbf{Distribution of scenes by scenario.} 
   In addition to DF3DV-41, which provides a more uniform distribution of scenes across scenarios, we further provide the scenario label for each scene in the full dataset.}
   \label{fig:scenario_dis}
\end{figure}

We introduce \textbf{DF3DV-1K (\underline{D}istractor-\underline{F}ree \underline{3D} \underline{V}ision \underline{1K})}, a newly collected dataset designed for distractor-free novel view synthesis.
The dataset provides clean and cluttered images for each scene, enabling systematic evaluation under diverse distractor conditions.
In total, DF3DV-1K contains \textbf{1,048} indoor and outdoor scenes covering \textbf{128} distractor types and \textbf{161} scene themes (see~\cref{fig:scene_and_theme}), comprising \textbf{89,924} images overall.
To better reflect practical usage scenarios, the data are captured under real-world conditions using consumer cameras, emphasizing casually captured imagery.
Within DF3DV-1K, we further introduce DF3DV-41, a curated subset consisting of systematically designed challenging scenarios intended to systematically evaluate distractor-free radiance field methods.
Details of the scene design and scenario categories are provided in~\cref{fig:bench_41_all_1,fig:bench_41_all_2}.
DF3DV-1K* follows a real-world long-tail distribution dominated by common scenes/objects (e.g., buildings, cars), and DF3DV-41 is more uniformly distributed by design. 
In addition, scenes can have multiple distractor types, so category counts are not mutually exclusive.
We believe the long-tail distribution better reflects real-world conditions, whereas DF3DV-41 supports systematic evaluation.
We also provide scenario annotations for each scene in DF3DV-1K to support future investigations (see~\cref{fig:scenario_dis}).
Through comprehensive benchmarking, we identify AsymGS~\cite{li2025asymgs} and RobustSplat~\cite{2025RobustSplat} as the most robust approaches, followed by OCSplats~\cite{ling2025ocsplats} and DeGauss~\cite{wang2025degauss}.
As illustrated in~\cref{fig:rank}, the ranking trends observed on DF3DV-1K broadly align with the chronological progression of recent research developments, suggesting that performance improvements in distractor-free radiance field methods have steadily advanced alongside methodological innovations.
This observation indicates that DF3DV-1K provides a realistic and challenging benchmark capable of reflecting progress in the field.
We also identify the weaknesses of recent distractor-free radiance field methods, such as their reliance on semantic features for distractor identification, which can lead to ambiguous decisions when distractors share similar semantics or appearance with static scene elements.
Finally, we present a generalizable model demonstrating that DF3DV-1K facilitates the transition from scene-specific methods toward more generalizable solutions through DI$^2$FIX, a distractor-free enhancement module for radiance fields.
DI$^2$FIX shows consistent improvements across ten radiance field methods, achieving average gains of 0.96 dB in PSNR and a 0.057 reduction in LPIPS.
\\
\\
\textbf{Annotation Logic.}
Although each data instance in DF3DV-1K is scene-level, theme annotations follow two configuration types. 
The first is scene-based, where the spatial environment is clearly identifiable and well defined (e.g., factory or construction site), similar to the naming logic of On-the-go~\cite{Ren2024NeRF}. 
The second is object-based, applied when the environment is less distinguishable (e.g., a restaurant dining scene), where themes are named according to the dominant object (e.g., a dish), similar to the naming strategy used in RobustNeRF~\cite{sabour2023robustnerf}. 
For most themes, we adopt high-level categories to ensure clarity and reproducibility (e.g., forks, knives, and spoons are grouped as cutlery, while cooking tools are categorized as utensils). 
To reduce semantic ambiguity, objects with multiple possible functions (e.g., a bowl) in our dataset are labeled using neutral object names rather than context-dependent categories.

Regarding distractor annotation logic, we also control semantic ambiguity, and distractors are distinguished based on perceptual specificity rather than broad functional class membership. 
For example, a general label is used (e.g., toy) when an object does not possess defining visual features that warrant further specification, while a more specific label is used when morphology, material, or texture clearly indicates a subtype (e.g., plush toy). 
For transportation-related categories, following common computer vision conventions, semantically distinct categories (e.g., car, truck, and bus) are preserved rather than merged into a single label.
For future studies, clustering-based approaches~\cite{yang2024toward,yang2026multi} represent a promising direction for automatically annotating the dataset with feature-based scenarios and reducing manual annotation effort.

\begin{figure}[tb]
  \centering
  \includegraphics[width=\linewidth]{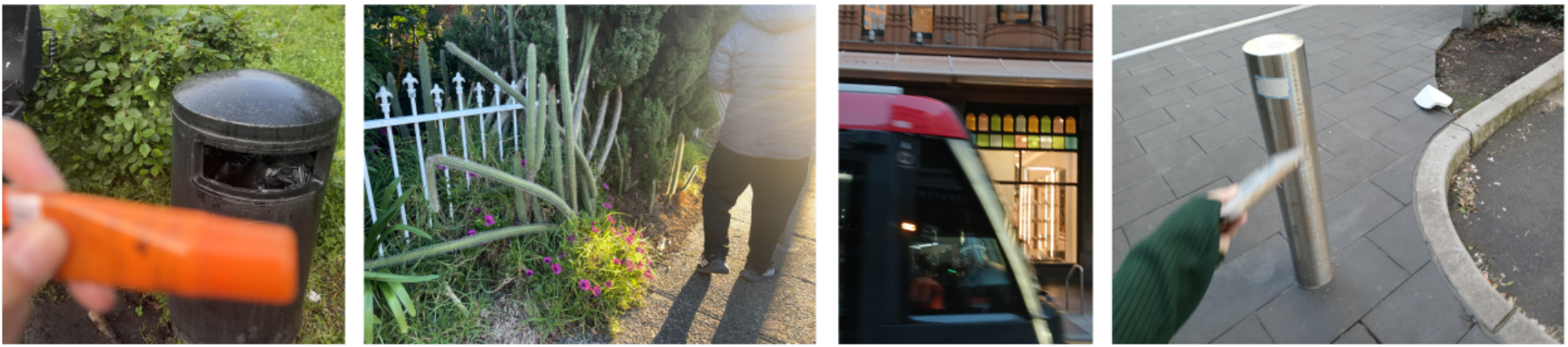}
   \caption{\textbf{Low-quality distractor examples.} 
   DF3DV-1K includes casually captured distractors with motion blur, partial visibility, defocus, and fast-moving objects, reflecting realistic capture conditions.}
   \label{fig:low_q}
\end{figure}

\end{document}